\documentclass[10pt,twocolumn,letterpaper]{article}

\usepackage{cvpr}
\usepackage{times}
\usepackage{epsfig}
\usepackage{graphicx}
\usepackage{amsmath}
\usepackage{amssymb}
\usepackage{bm}
\usepackage{cite}
\usepackage[percent]{overpic}
\usepackage{multicol}
\usepackage{multirow}
\usepackage[table]{xcolor}
\usepackage{color}

\def\bal#1\eal{\begin{align}#1\end{align}} 
\def\suml{\sum\limits}

\newcommand{\br}[1]{\left[#1\right]} 
\newcommand{\pr}[1]{\left(#1\right)} 
\newcommand{\cbr}[1]{\left\{#1\right\}} 
\DeclareMathOperator*{\argmin}{arg\,min} 
\DeclareMathOperator*{\sign}{sign}
\def\transp{\mathsf{T}} 
\def\m{\mathbf}
\def\mc{\mathcal}
\def\R{\mathbb{R}}
\def\ast{*}

\newcommand{\grad}{\ensuremath{\nabla}} 
\newcommand{\norm}[2]{\ensuremath{\left\|#1\right\|_{#2}}}
\newcommand{\ip}[2]{\ensuremath{\langle#1\,,\,#2\rangle}} 
\newcommand{\abs}[1]{\ensuremath{\lvert #1\rvert}}
\newcommand {\bbmtx}{\begin{bmatrix}} 
\newcommand {\ebmtx}{\end{bmatrix}} 
\newcommand{\mean}[1]{\bar{#1}}

\usepackage[pagebackref=true,breaklinks=true,letterpaper=true,colorlinks,bookmarks=false]{hyperref}

 \cvprfinalcopy 

\def\cvprPaperID{****} 

\ifcvprfinal\fi
\begin{document}

\title{Universal Denoising Networks : A Novel CNN Architecture for Image Denoising
}

\author{Stamatios Lefkimmiatis\\
Skolkovo Institute of Science and Technology (Skoltech), Moscow, Russia\\
{\tt\small s.lefkimmiatis@skoltech.ru}
}

\maketitle

\begin{abstract}
We design a novel network architecture for learning discriminative image models that are employed to efficiently
tackle the problem of grayscale and color image denoising. Based on the proposed architecture, we introduce two different 
variants. The first network involves convolutional layers as a core component, while the second one 
relies instead on non-local filtering layers and thus it is able to exploit the inherent non-local self-similarity property of 
natural images. As opposed to most of the existing deep network approaches, which require the training of a specific model for each 
considered noise level, the proposed models are able to handle a wide range of noise levels using a single set of learned parameters, while they are very robust when the noise degrading the latent image does not match the statistics of the noise used during training. The latter argument is supported by results that we report on publicly available images corrupted by unknown noise and which we compare against solutions obtained by competing methods.  At the same time the introduced networks achieve excellent results under additive white Gaussian noise (AWGN), which are comparable to those of the current state-of-the-art network, while they depend on a more shallow architecture with the number of trained parameters being one order of magnitude smaller. These properties make the proposed networks  ideal candidates to serve as sub-solvers on restoration methods that deal with general inverse imaging problems such as deblurring, demosaicking, superresolution, \etc.
\end{abstract}\vspace{-0.5cm}

\section{Introduction}
Image denoising is among the basic low-level computer-vision problems and has received significant attention in both academic
research as well as in practical digital imaging applications~\cite{NeatImage,DxO}. However, during the past decade there was little progress 
in improving the state-of-the art denoising performance and it has been suggested that denoising algorithms have reached 
optimality and cannot be further improved~\cite{Levin2011}. Despite these beliefs, very recently and thanks to the advent of deep learning techniques, several powerful image denoising algorithms that managed to significantly improve the state-of-the-art performance have been introduced~\cite{Schmidt2014,Chen2016,Vemulapalli2016,Lefkimmiatis2017,Zhang2017,Zhang2017b}. Nevertheless, their wide applicability in real-world applications is currently hindered mainly because the majority of them involves the training of a specific model for each considered noise level. Such requirement is rather impractical since it implies that a huge number of network parameters, proportional to the number of noise levels that the models are trained for, needs to be stored. This directly excludes the application of such methods on devices with limited memory storage. Another important limitation of such deep-learning methods is that their denoising performance deteriorates very fast when the noise level distorting the input images deviates from the one that the model was originally trained for. 

In this work, motivated by the recent advances in deep learning and relying on the rich body of algorithmic ideas developed in the past for  image restoration problems, we introduce a novel network architecture specifically tailored to image denoising, which can overcome the aforementioned limitations. Specifically, our derived networks allow the training of image models that can handle a wide range of noise levels and they come in two variants. The first network involves convolutional layers as a core component and behaves similarly to local variational methods, while the second one relies on non-local filtering layers that allow us to exploit the inherent non-local self-similarity property of natural images. Both networks lead to very competitive results, which are directly comparable to the state-of-the art, while their advantage is that they involve considerably less parameters than the current best-performing network. Furthermore, they are robust and perform very well for inputs distorted by noise whose statistics differ from the ones of the noise model used during training.

\section{Image Restoration}
To restore a latent grayscale or color image $\m X$ from a 
corrupted observation $\m Y$, we rely on the linear model 
\bal
\m y = \m H\m x + \m n.
\label{eq:ForwardModel}
\eal
In this setting, $\m y$, $\m x\in\R^{N\cdot C}$ are the vectorized versions of the observed and latent images, $\m Y$ and $\m X$ 
respectively, $N$ is the number of pixel in each image channel, $C$ is the number of channels, $\m H$ is a linear operator 
that corresponds to the response of the imaging device,  and $\m n$ is the measurement noise that accounts for all possible 
errors during image acquisition, including stochastic noise and the possible mismatch between the observation model and the physical
image acquisition process. 
For image denoising, which is the focus of this work, the linear operator $\m H$ reduces to the 
identity matrix $\m I$, since it is assumed that the imaging device does not introduce any other distortions to the latent
signal. Regarding the term $\m n$, the most common assumption in the literature, which we also adopt in this work, is that it is 
zero-mean i.i.d Gaussian noise with variance $\sigma^2$. 

While the additive white Gaussian noise (AWGN) assumption is not 
frequently met in practice, an efficient solution of this problem is extremely valuable for two main reasons. The first one is that even in cases 
where the noise is signal dependent, there are several techniques available in the literature, such as variance stabilization transforms (VST)~\cite{Anscombe1948, Fisz1955, Makitalo2014},  
which are able to transform the input data in a different domain so that the noise follows a Gaussian distribution with a fixed variance. Therefore, the solution can be obtained by first performing Gaussian denoising in the transform domain and then mapping
the solution back to the original domain using the inverse VST. The second reason is that such a solution, in the context of convex optimization, can be interpreted as a proximal map~\cite{Parikh2013} of a regularization function. Such proximal maps typically serve as building blocks of several powerful optimization schemes that have been proposed in the literature, including Majorization-Minimization~\cite{Hunter2004,Figueiredo2007} and variable-splitting strategies~\cite{Boyd2011}. These optimization strategies can address more general image restoration problems such as image deblurring, superresolution, demosaicking, inpainting, \etc. 
\vspace{-.05cm}
\subsection{Image Priors}
\label{sec:imagePriors}

While Eq.~\eqref{eq:ForwardModel} corresponds to a linear problem, the presence of the noise, whose exact 
realization is unknown, combined with the fact that typically the operator $\m H$ is singular, makes it an 
ill-posed problem~\cite{Bertero1998,Vogel2002}. This implies that a unique solution does not exist and therefore we cannot rely solely 
on the image evidence but we need to further exploit \emph{a priori} information. In this case, the utilization of suitable prior models of image or scene properties plays an instrumental role in the success of image restoration methods.

Several strategies for imposing prior knowledge on the solution are available and among the most popular ones
is the variational approach. In this framework, image recovery is cast as a minimization problem of an objective 
function of the form
\bal
f\pr{\m x} = d\pr{\m x; \m H, \m y} + \lambda\, r\pr{\m x},
\label{eq:objectiveFun}
\eal
where the minimizer corresponds to the recovered latent image. 
The role of the objective function is to quantify the quality of the solution and typically consists of two terms as shown in
Eq.~\eqref{eq:objectiveFun}. The first term is the \emph{data fidelity},
which measures the proximity of the solution to the observation, and the second one is the \emph{regularizer}. The role of 
the regularizer is crucial since it encodes our prior knowledge by penalizing solutions that do not feature the desired properties. 
The parameter $\lambda \ge 0$, is used to combine the two terms and to adjust their contribution on the final result. 

Interestingly, the variational approach has direct links to Bayesian estimation methods and the derived solutions can be interpreted either as penalized maximum likelihood or as maximum a posteriori (MAP) estimates~\cite{Bertero1998,Figueiredo2007}.

As emphasized previously, a good choice for the regularizer is instrumental to the success of any variational-based image restoration method. 
A generic formulation that can be used to describe the majority of the most successful regularizers in the literature, is provided below
\bal
r\pr{\m x} = \suml_{k=1}^K \phi\pr{\m L_k \m x},
\label{eq:analysisReg}
\eal
where $\m L: \R^N \mapsto \R^{K \times D}$ corresponds to the regularization operator ($\m L_k\m x\in \R^D$ denotes the D-dimensional $k$-th entry of the result obtained by applying $\m L$ to the image $\m x$) and $\phi : \R^D \mapsto \R$ is a potential function. 
Indeed, by varying the regularization operator $\m L$ and the potential function $\phi$ we can derive several existing regularization functionals. Typical choices for the operator $\m L$ are first or higher-order differential operators such as the gradient~\cite{Rudin1992,Bredies2010}, the structure tensor~\cite{Lefkimmiatis2015J}, the Laplacian and the Hessian~\cite{Lefkimmiatis2012J,Lefkimmiatis2013J}. For the potential function there is also a wide variety of possible choices with the most popular ones being the $\ell_p$ vector norms and the Schatten matrix norms. The main reason for this is that their combinations with linear operators leads to convex regularizers which are amenable to efficient optimization and provide certain convergence guarantees.

Besides the local regularization methods mentioned above\footnote{These methods are considered local in the sense that the regularization operator is localized and its influence is restricted in a small area around the pixel of interest.}, the definition in Eq.~\eqref{eq:analysisReg} can also be used to describe non-local regularization functionals such as those in~\cite{Zhou2005,Kindermann2005,Elmoataz2008,Gilboa2008,Lefkimmiatis2015bJ}. In this case, $\m L$ is designed so that it allows interactions between distant points in the image domain. This way it is possible to capture long-range dependencies between image points and thus model the so called \emph{non-local self similarity} (NLSS) property that natural images exhibit. This property implies that images typically consist of localized patterns that are repeated in different and possibly distant locations in the image domain. NLSS is an important property and if properly exploited it can effectively distinguish the image content from noise and other types of distortions. This has been succesfuly demonstrated for several image restoration problems~\cite{Dabov2007,Gilboa2008,Lefkimmiatis2015bJ}.

\subsection{Constrained Optimization}\label{sec:constrOptim}
In the variational framework the choice of the regularizer has an important effect on the quality of the restored image. Equally important is our ability to efficiently compute the minimizer of the overall objective function. Image denoising under AWGN, amounts to solving an unconstrained optimization problem of the form :\vspace{-0.2cm}
\bal
\m x^{\ast}&=\argmin_{\m x}\frac{1}{2}\norm{\m y - \m x}{2}^2 + \lambda \suml_{k=1}^{K}\phi\pr{\m L_k \m x},
\label{eq:energyMin}
\eal
where for the regularizer we use the generic description of Eq.~\eqref{eq:analysisReg}, while for the fidelity term we use
a quadratic cost, in accordance with the Gaussian noise assumption. 

As mentioned earlier, $\lambda$ is a `free' parameter that needs to be tuned by the user and different values lead to different restoration results of varying image quality. Therefore, among others one of the main challenges is to choose the value for the regularization parameter $\lambda$, that will lead to the optimum result under some image quality criterion. Unfortunately, there is not a direct way to \emph{a priori} relate the strength of $\lambda$ with the quality of the result. Therefore, in practice, $\lambda$ is either tuned empirically or heuristic techniques such as the L-curve method~\cite{Hansen2000} are employed, which involve solving Problem~\eqref{eq:energyMin} for several values of $\lambda$. 

One way to circumvent this difficulty, is to consider the following equivalent formulation\vspace{-0.2cm}
\bal
\m x^{\ast}&=\argmin_{\substack{ \\ \ \norm{\m y - \m x}{2}\le \varepsilon}}\,\sum\limits_{k=1}^{K}\phi\pr{\m L_k \m x},
\label{eq:constrEnergyMin}
\eal
which transforms the original problem to a constrained optimization form. Problems~\eqref{eq:energyMin} and \eqref{eq:constrEnergyMin} are equivalent in the sense that : for any $\varepsilon > 0$ such that Problem~\eqref{eq:constrEnergyMin} is feasible, a solution of \eqref{eq:constrEnergyMin} is either the null vector or else it is a solution of Problem~\eqref{eq:energyMin} for some $\lambda > 0$~\cite{Rockafellar1970}. To highlight what is the gain by pursuing such a reformulation, we note that while in Eq.~\eqref{eq:constrEnergyMin} there is still a free parameter $\varepsilon$ that needs to be tuned, this parameter is directly related to the noise level distorting the latent image $\m x$. In particular, it holds that $\norm{\m y - \m x}{2} = \norm{\m n}{2} \propto \sigma$. Given that there are several methods available for estimating the standard deviation of the noise from the noisy input~\cite{Foi2009,Liu2013}, we now have a good indication about the range of values that the parameter $\varepsilon$ should lie in, as opposed to $\lambda$ in the previous formulation.

\subsection{Minimization Strategy}\label{sec:PGM}

To attack the minimization problem of Eq.~\eqref{eq:constrEnergyMin} we can rely on a splitting variable technique such as the Alternating Direction Method of Multipliers~\cite{Boyd2011}. Here, however, we opt for a simpler approach that utilizes a gradient descent algorithm. To do so, we first rewrite Eq.~\eqref{eq:constrEnergyMin} as 
\bal
\m x^{\ast}&=\argmin_{\m x}\,\sum\limits_{k=1}^{K}\phi\pr{\m L_k \m x} + \iota_{\mc{C}\pr{\m y,\varepsilon}}\pr{\m x},
\label{eq:constrEnergyMinEquiv}
\eal
where \vspace{-0.3cm}
\bal
\iota_{\mc{C}\pr{\m y,\varepsilon}}\pr{\m x} = 
\begin{cases}
0, &  \mbox{if }  \norm{\m y - \m x}{2}\le \varepsilon \\
\infty,  & \mbox{otherwise}
\end{cases},
\eal
is the indicator function of the convex set $\mc{C}$.

Next, we assume that the potential function $\phi$ is smooth and thus we can compute its partial derivatives. Since this is 
not the case for the indicator function, instead of the gradient descent algorithm we employ the proximal gradient method (PGM)~\cite{Parikh2013}. This is a gradient descent variant that can deal with functions consisting of both smooth and non-smooth terms. According to PGM the function $f\pr{\m x}$ to be minimized is split into two terms, a smooth and a non-smooth one. In our case we naturally have $f\pr{\m x} = r\pr{\m x} + \iota_{\mc{C}}\pr{\m x}$, where based on the smoothness assumption for the potential function $\phi$, the regularizer $r\pr{\m x}$ corresponds to the smooth term. Then, the solution is computed in an iterative fashion, using the update rule
\bal
\m x^{t} & =\mbox{prox}_{\gamma^t\iota_C}\pr{\m x^{t-1} - \gamma^t \grad_{\m x} r\pr{\m x^{t-1}}},
\label{eq:update}
\eal
where $\gamma^t$ is a step-size and $\mbox{prox}_{\gamma^t\iota_C}$ is the proximal operator~\cite{Parikh2013} related to the indicator function $\iota_{\mc{C}}$. 

The proximal map of the indicator function $\iota_{\mc{C}}$ in Eq.~\eqref{eq:update} corresponds to an orthogonal projection of the input onto the set $\mc{C}$. This can be computed in closed form as 
\bal
\Pi_{\mc{C}}\pr{\bm v} = \m y + \varepsilon \frac{\bm v - \m y}{\max\pr{\norm{\bm v -\m y}{2},\varepsilon}}.
\label{eq:projection}
\eal
Given that the gradient of the regularizer is computed as 
\bal
\grad_{\m x} r\pr{\m x} = \sum\limits_{k=1}^{K}\m L_k^{\transp}\psi\pr{\m L_k \m x} \equiv h\pr{\m x},
\label{eq:regGrad}
\eal
with  $\psi\pr{\m z} = \grad_{\m z}\phi\pr{\m z}$, $\m z\in \R^D$ and using Eq.~\eqref{eq:projection}, we re-write Eq.~\eqref{eq:update} as 
\bal\hspace{-.18cm}
\m x^{t} & = \Pi_{\mc{C}}\pr{\m x^{t-1} - h^{t}\pr{\m x^{t-1}}} \mbox{with } h^{t}\pr{\m x}=\gamma^t\, h\pr{\m x}.
\label{eq:update_}
\eal

A careful inspection of Eqs.~\eqref{eq:projection} and \eqref{eq:update_} leads us to the useful observation that under this approach the solution is obtained by recursively subtracting from the input refined estimates of the noise realization that distorts it. In particular, for the first iteration and given that $\m x^0 = \m y$ we have 
\bal
\m x^1 = \m y -  \varepsilon \frac{h^{1}\pr{\m y}}{\max\pr{\norm{h^{1}\pr{\m y}}{2},\varepsilon}} = \m x + (\m n - \m n^1).
\eal
Here $h^1\pr{\m y}$ can be interpreted as a noise estimator, which infers from the input the noise realization that distorts it.
The noise realization estimate is further normalized to ensure that it has the correct variance and then it is subtracted from the noisy input. This leads to an output which consists of the latent image plus some residual noise, $\m n - \m n^1$. The subsequent updates refine the noise estimate and remove it from the original input as follows 
\bal
\m x^k = \m y -\m n^{k} = \m x + \pr{\m n - \m n^k}, \quad k > 1
\eal
where $\m n^k = \varepsilon\,\frac{\pr{\m n^{k-1} + h^k\pr{\m x^{k-1}}}}{\max\pr{\norm{\m n^{k-1} + h^k\pr{\m x^{k-1}}}{2},\varepsilon}} .$

\begin{figure}[t]
\centering
   \includegraphics[width=0.8\linewidth]{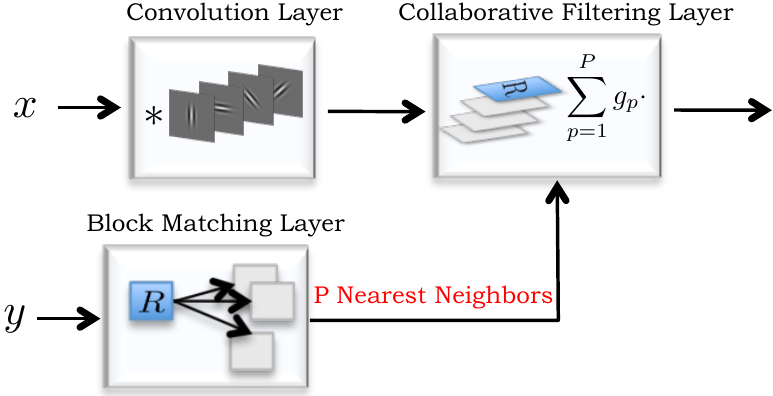}
 \caption{\small The architecture of the non-local filtering layer. The input of the layer is $x$ while $y$ is the network input based on which the block matching is computed.}
\label{fig:NLLayer}
\vspace{-0.35cm}
\end{figure}
\vspace{-0.1cm}
\section{Proposed Network}\label{sec:Net}
From the previous analysis it is clear that the success of the iterative denoising scheme that we described in Section~\ref{sec:PGM} depends exclusively on how well the function $h$, defined in Eq.~\eqref{eq:regGrad}, can estimate the realization of the noise. Designing such a function amounts to specifying the operator $\m L$ and the gradient of the potential function $\phi$. Manually selecting proper values for these parameters is a cumbersome task. For this reason, we pursue a machine learning approach and design a neural network that has the capacity to learn these parameters in a discriminative fashion from training data. Towards this end, we consider each PGM update as a composition of network layers and construct our network as a cascade of them. We emphasize that as opposed to previous networks~\cite{Schmidt2014,Chen2016,Lefkimmiatis2017} that followed a similar unrolling strategy, our network architecture is based on a constrained minimization formulation~\eqref{eq:constrEnergyMin} rather than on an unconstrained one~\eqref{eq:energyMin}. This is an important difference and the key for deriving networks that can handle a wide range of noise levels using a single set of learned parameters. The main reason is that under our adopted formulation the parameter $\varepsilon$ does not need to be learned for a specific noise level as is the case for the parameter $\lambda$ in~\eqref{eq:energyMin}. We further note that this is accomplished without any sacrifice in reconstruction quality or computational complexity. 

The remaining issue to be addressed is the parameterization of the operator $\m L$ and function $\psi$ in a way that will facilitate the learning of the network's parameters in an efficient and computationally tractable way. 
\begin{figure}[t]
\hspace{-0.6cm}
   \includegraphics[width=1.15\linewidth]{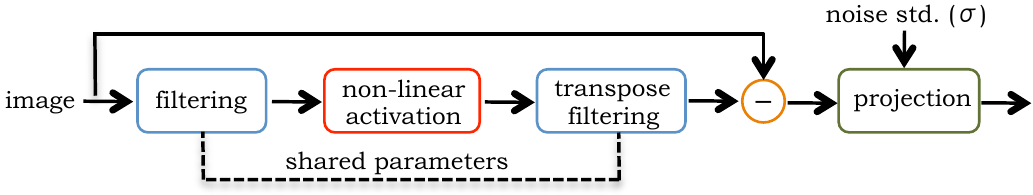}
 \caption{\small Schematic representation of the composite layer that serves as the core component of the proposed network architecture. Depending on the parametrization of the regularization operator, the filtering (transpose filtering) layer corresponds to either a convolutional or a non-local filtering layer.}
\vspace{-0.35cm}
\label{fig:CompositeLayer}
\end{figure}
\subsection{Local and Non-local Operators}\label{sec:OperatorModeling}
As mentioned in Section~\ref{sec:imagePriors} common choices for the regularization operator $\m L$ are local differential operators.
In the discrete setting, image derivatives are typically computed as convolutions of the image with a filterbank consisting of several high-pass kernels. Naturally, this leads us to parametrize $\m L$ as a convolutional layer, which is a widely used component in modern deep neural networks. One point however that requires our attention is that in order to learn a valid regularization operator, the filters utilized for its parametrization need to be zero-mean~\cite{Chen2014}. Moreover, since the function $\psi$ will also be learned, if we inspect equation~\eqref{eq:regGrad} we will notice that without further imposing a fixed scale to the operators $\m L_k$, it is possible that two different sets of parameters $\pr{\m L_k,\psi}$ and $\pr{\hat{\m L}_k,\hat{\psi}}$ lead to the same result. Such a situation can occur if we choose $\hat{\m L}_k = \beta \m L_k$ and $\hat{\psi}\pr{\beta \m z} = \frac{1}{\beta} \psi\pr{\m z}$, which gives $\m L_k^{\transp}\psi\pr{\m L_k\m x} = \hat{\m L}_k^{\transp}\hat{\psi}\pr{\hat{\m L}_k \m x}$. To ensure that the learned operators will satisfy both the zero-mean and the fixed-scale constraints, we parametrize the weights $\bm w\in \R^L$ of each filter in the convolutional layer as 
\bal
\bm w = s \pr{\m v - \mean{\m v}}/\norm{\m v - \mean{\m v}}{}, 
\label{eq:weightNorm}
\eal
where $s$ is a scalar trainable parameter. Training the convolutional layers with this new parametrization can be done as usual using a stochastic gradient descent method. The gradient of the new parameters $s$ and $\m v$ w.r.t the loss function $\mc{L}$ can be computed as 
\bal
\grad_{\m s}\mc{L} =\ip{\bm w / s}{\grad_{\bm w}\mc{L}}\,\mbox{ and }\, \grad_{\m v}\mc{L} =\m M_{\m v}\grad_{\bm w}\mc{L},  
\eal
where $\m M_{\m v} = \frac{s}{\norm{\m v - \mean{\m v}}{}}\pr{\m I - \frac{\m 1 \m 1^\transp}{L}}\pr{\m I - \frac{\pr{\m v - \mean{\m v}}\pr{\m v - \mean{\m v}}^\transp}{\norm{\m v - \mean{\m v}}{}^2}}$, $\mean{\m v}$ denotes the mean value of $\m v$,  and $\m 1$ is a column vector of ones.

Interestingly, the same parametrization of Eq.~\eqref{eq:weightNorm} but without the mean subtraction has been recently proposed in~\cite{Salimans2016} as an alternative to batch-normalization~\cite{Ioffe2015}. However, in~\cite{Salimans2016} the motivation is different and the goal is to make the training of deep networks more robust. 

Besides the parametrization of $\m L$ as a local operator, we further explore another option where we model $\m L$ as a non-local operator. This leads to a second variant of the proposed network architecture. Our motivation for employing a non-local regularization operator is that the resulting network can take advantage of the NLSS property that we referred to in detail in the introduction. To this end, we adopt the parametrization that was proposed in~\cite{Lefkimmiatis2017}. Specifically, the non-local operator can be expressed as the composition of three layers : \textbf{1}) a convolution layer, where we use the parametrization that we adopted earlier, which applies a linear transformation to every patch extracted from the image\footnote{Passing an image through a convolution layer of $F$ filters whose support are $H\times W$, corresponds to applying a linear mapping $\R^{H\times W}\mapsto \R^F$ to every image patch of size $H\times W$. In addition, the stride of the convolutional layer determines the overlap between consecutive  image patches.}, \textbf{2}) a block-matching layer which forms a 3-D group for every valid patch and it consists of the $P$ most similar patches to the reference patch (including the reference patch itself) and \textbf{3}) a collaborative-filtering layer that filters the grouped patches along the dimension of the group in a similar fashion as the Non-Local Means filter (NLM)~\cite{Buades2010}. In this last layer each 3-D group is projected to a single patch. A schematic representation of the described parametrization is provided in Fig.~\ref{fig:NLLayer}. This combination of layers leads to a non-local filtering approach similar to the one initially proposed in~\cite{Dabov2007} but with two main differences. The first one is that the authors in~\cite{Dabov2007} use fixed transforms while here the transforms are learned. The second one is that in the third step of the non-local operation an 1-D transformation along the group dimension is applied instead of the weighted sum that we use in this work. Therefore, in our case the output of the adopted non-local filtering layer leads to an output of the same size as the input, while in~\cite{Dabov2007} the output is augmented in the third dimension by a factor of $P$. An additional remark relevant to the implementation of the described non-local filtering layer is that we constrain the weights of the third sub-layer to sum to one. This is consistent to the way that the NLM filter is defined. To impose such constraint we parametrize the weights $\bm g\in R^P$ as 
$\bm g = \nu^{-1}\bm u$ with $\nu = \ip{\m 1}{\bm u}$. In this case the gradient of the new parameters $\bm u$ w.r.t the loss function $\mc{L}$ are computed as 
\bal 
\grad_{\bm u}\mc{L} = \nu^{-1}\pr{\m I - \nu^{-1}\bm u^\transp}\grad_{\bm g}\mc{L}.
\eal

\subsection{Parametrization of the Potential Function}\label{sec:RBF}
Having defined the parametrization of the operator $\m L$, we further need to model the function $\psi$ (see Eq.~\eqref{eq:regGrad}), which corresponds to the gradient of the potential function $\phi$. 
To do so, first we assume that the potential function $\phi$ is separable, that is it can be expressed in the form
\vspace{-0.2cm}
\begin{equation}
\phi\pr{\m z} = \suml_{d=1}^{D} \phi_d\pr{\m z_d},
\vspace{-0.15cm}
\end{equation}
and thus, $\psi\pr{\m z} \!\!=\!\! \bbmtx \psi_1\pr{\m{z}_1}\!\!&\!\! \psi_2\pr{\m{z}_2}\!\!&\!\! \hdots \!\!&\!\! \psi_D\pr{\m{z}_D}\ebmtx^{\transp}\!\equiv\!\grad_{\m z}\phi\pr{\m z}$, with $\psi_i\pr{\m z_i}  = \partial \phi\pr{\m z}/\partial{\m z_i}$. Next, we
parametrize the partial derivatives $\psi_i$ as a linear combination of Radial Basis Functions (RBFs),  \ie 
\vspace{-0.2cm}
\begin{equation}
\psi_i\pr{x} = \suml_{j=1}^M \pi_{ij}\rho_j\pr{\abs{x-\mu_j}},
\vspace{-0.15cm}
\end{equation}
where $\pi_{ij}$ are the expansion coefficients and $\mu_j$ are the centers of the basis functions $\rho_j$~\cite{Hu2001}. For our networks we use Gaussian RBFs, $\rho_j\pr{r} = \exp\pr{-a_j r^2}$, and we employ $M=51$ Gaussian kernels whose centers are distributed equidistantly in the range [-100, 100] and they all share the same precision parameter $a$. To make sure that the input $x$ lies in the specified range, a clipping layer is preceding the RBF-mixture layer.
The representation of $\psi_i$ using mixtures of RBFs is very powerful and allow us to approximate with high accuracy arbitrary non-linear functions. Details about the computation of the gradient of the parameters $\pi_{ij}$ and of the input $\m z$ w.r.t to the loss function $\mc{L}$ can be found in~\cite{Lefkimmiatis2017}.

\subsection{Trainable Projection Layer}\label{sec:Projection}
The final component for the construction of the proposed network architecture is the projection layer, which is defined in Eq.~\eqref{eq:projection}. We parametrize the threshold $\varepsilon$ as $\varepsilon = e^{\alpha}\sigma \sqrt{N_t -1}$, where $\sigma$ is the standard deviation of the noise distorting the network input, $N_t$ is the total number of pixel in the image, and $\alpha$ is a trainable parameter. We note, that in our work we learn a single common $\alpha$ for various values of $\sigma$, where $\sigma$ as shown in Fig.~\ref{fig:CompositeLayer} is provided as an additional input to our network.

Based on this parametrization and using the identity $\max\pr{x,y} = 0.5\pr{\abs{x-y} + x + y}$, we compute the gradient of the input $\bm v$ w.r.t to the loss function $\mc{L}$ as 
\bal
\grad_{\bm v}\mc{L} = \varepsilon\gamma\pr{\m I -\beta^+\gamma^2\pr{\bm v -\m y}\pr{\bm v -\m y}^\transp}\grad_{\bm q}\mc{L},
\label{eq:grad_input}
\eal
where $\bm q = \Pi_\mc{C}\pr{\bm v}$, $\beta^+ = \pr{1+\sign\pr{\norm{\bm v -\m y}{2}-\varepsilon}}/2$ and $\gamma = 1/\max\pr{\norm{\bm v -\m y}{2},\varepsilon}$. Additionally, the gradient of the  parameter $\alpha$ w.r.t the loss function $\mc{L}$ is computed as
\bal
\grad_{\alpha}\mc{L} = \mu\pr{\bm v -\m y}^\transp\grad_{\bm q}\mc{L},
\label{eq:grad_param}
\eal
where $\beta^- = \pr{1-\sign\pr{\norm{\bm v -\m y}{2}-\varepsilon}}/2$ and $\mu = \varepsilon\gamma\pr{1-\varepsilon\gamma\beta^{-}}$. Note that for all the formulas above we are using the convention that $\sign\pr{0} = -1$. 


\begin{figure*}[ht]
\centering
\begin{tabular}{@{} c @{} c @{} c @{} c @{} c @{}}
\hspace{-0.6cm}
 \begin{overpic}[width=.215\linewidth]{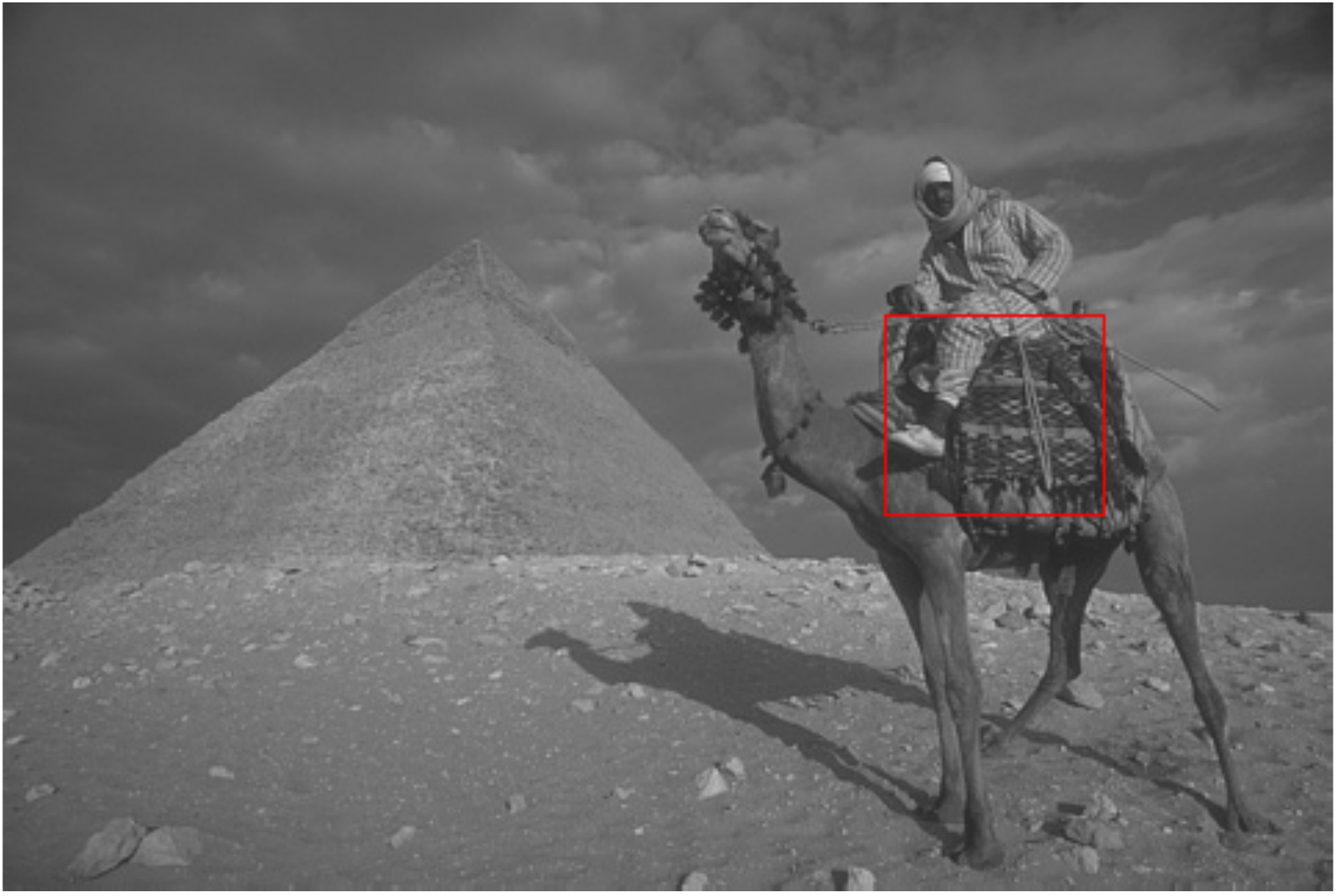}
  \put(79.5,48.5){\textcolor{red}{\fboxrule=0.5pt\fboxsep=0pt\fbox{\includegraphics[scale=.25,trim=1 2 2 1,clip=true]{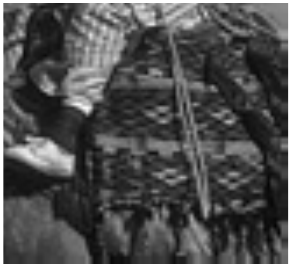}}}}
  \end{overpic}&
 \begin{overpic}[width=.215\linewidth]{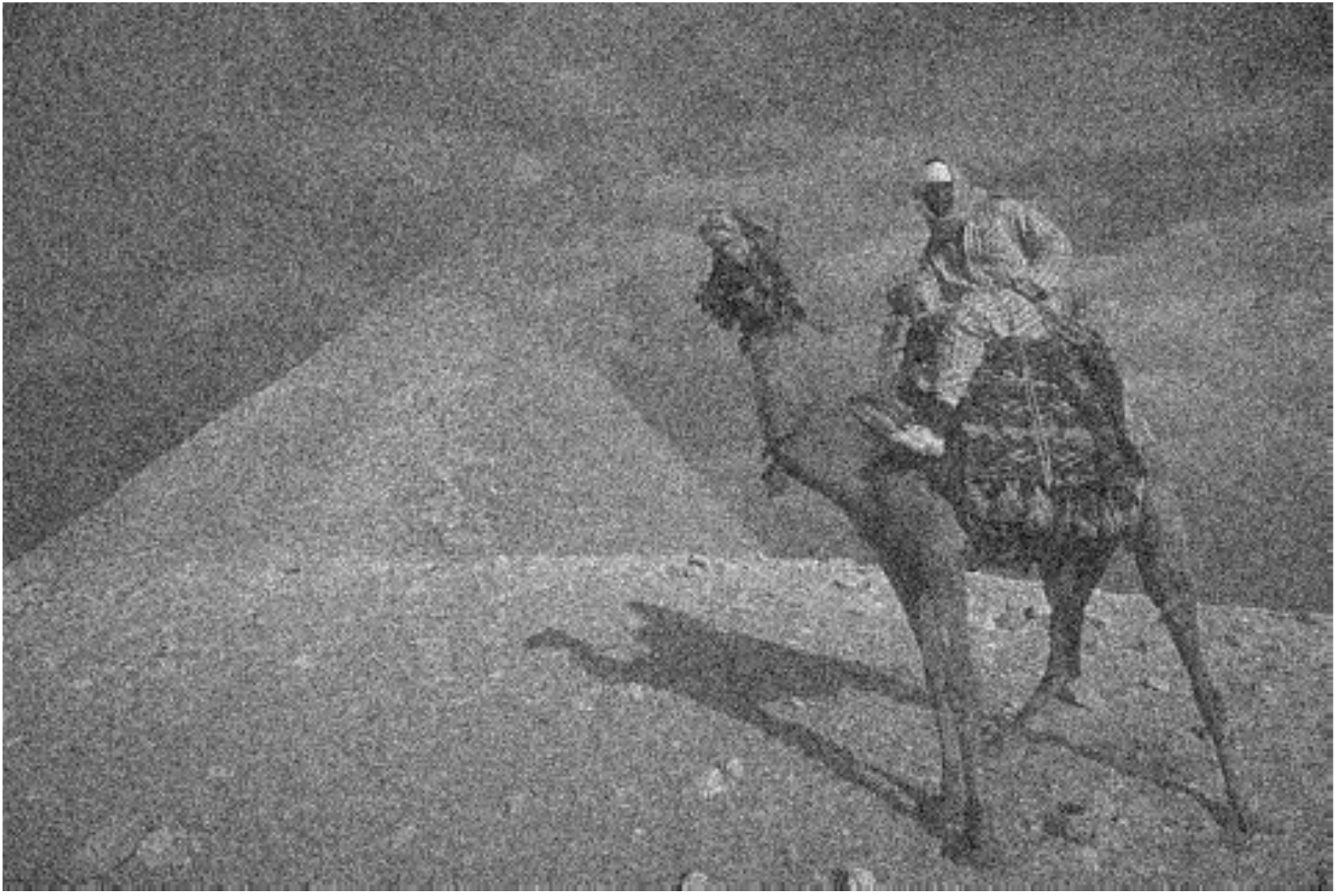}
  \put(79.5,48.5){\textcolor{red}{\fboxrule=0.5pt\fboxsep=0pt\fbox{\includegraphics[scale=.25,trim=1 2 2 1,clip=true]{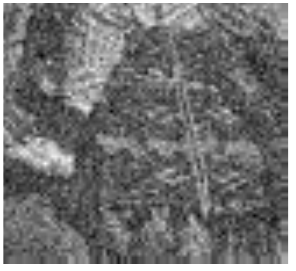}}}}
  \end{overpic}&  
 \begin{overpic}[width=.215\linewidth]{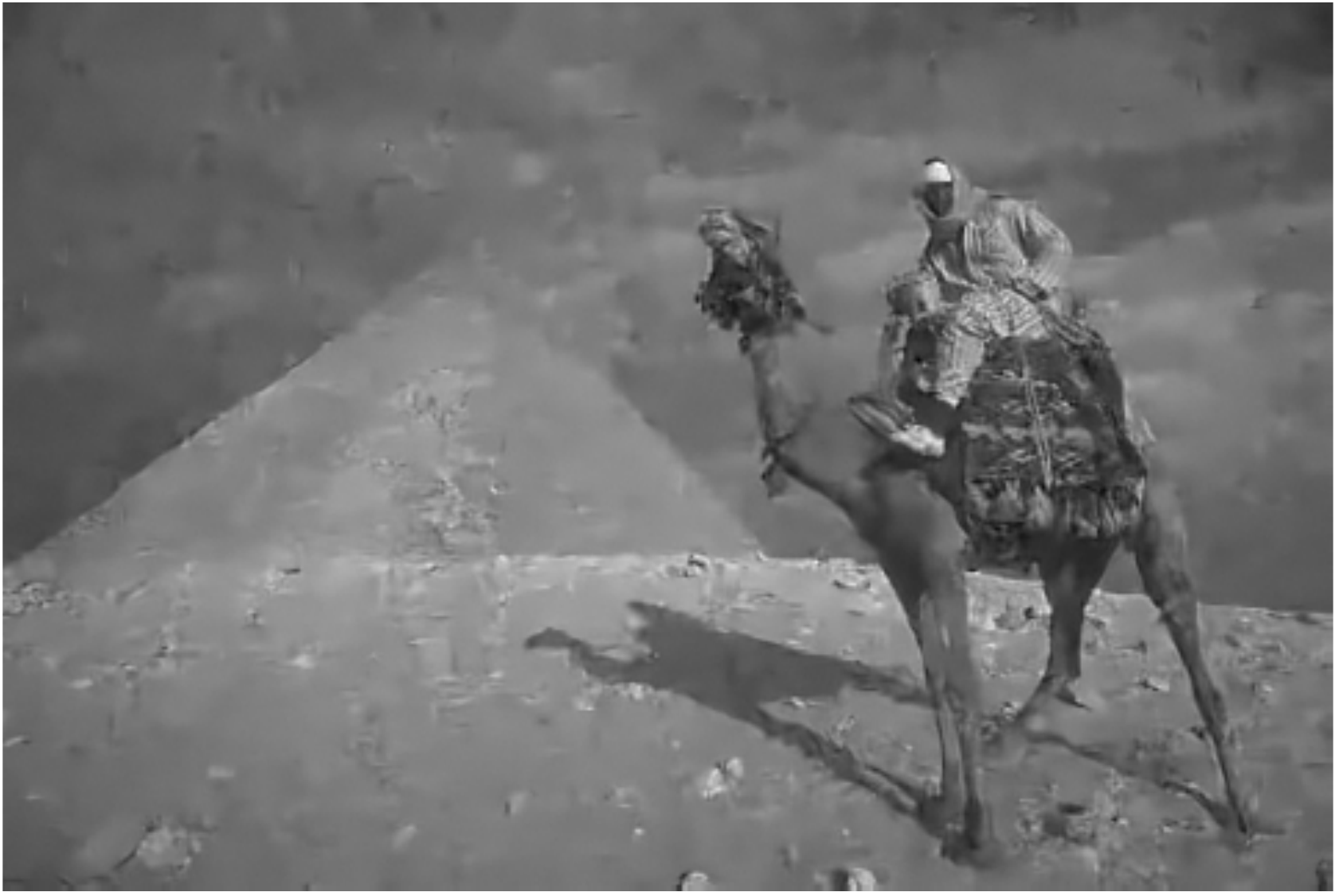}
  \put(79.5,48.5){\textcolor{red}{\fboxrule=0.5pt\fboxsep=0pt\fbox{\includegraphics[scale=.25,trim=1 2 2 1,clip=true]{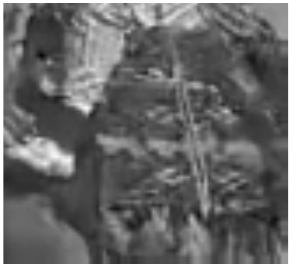}}}}
  \end{overpic}&    
 \begin{overpic}[width=.215\linewidth]{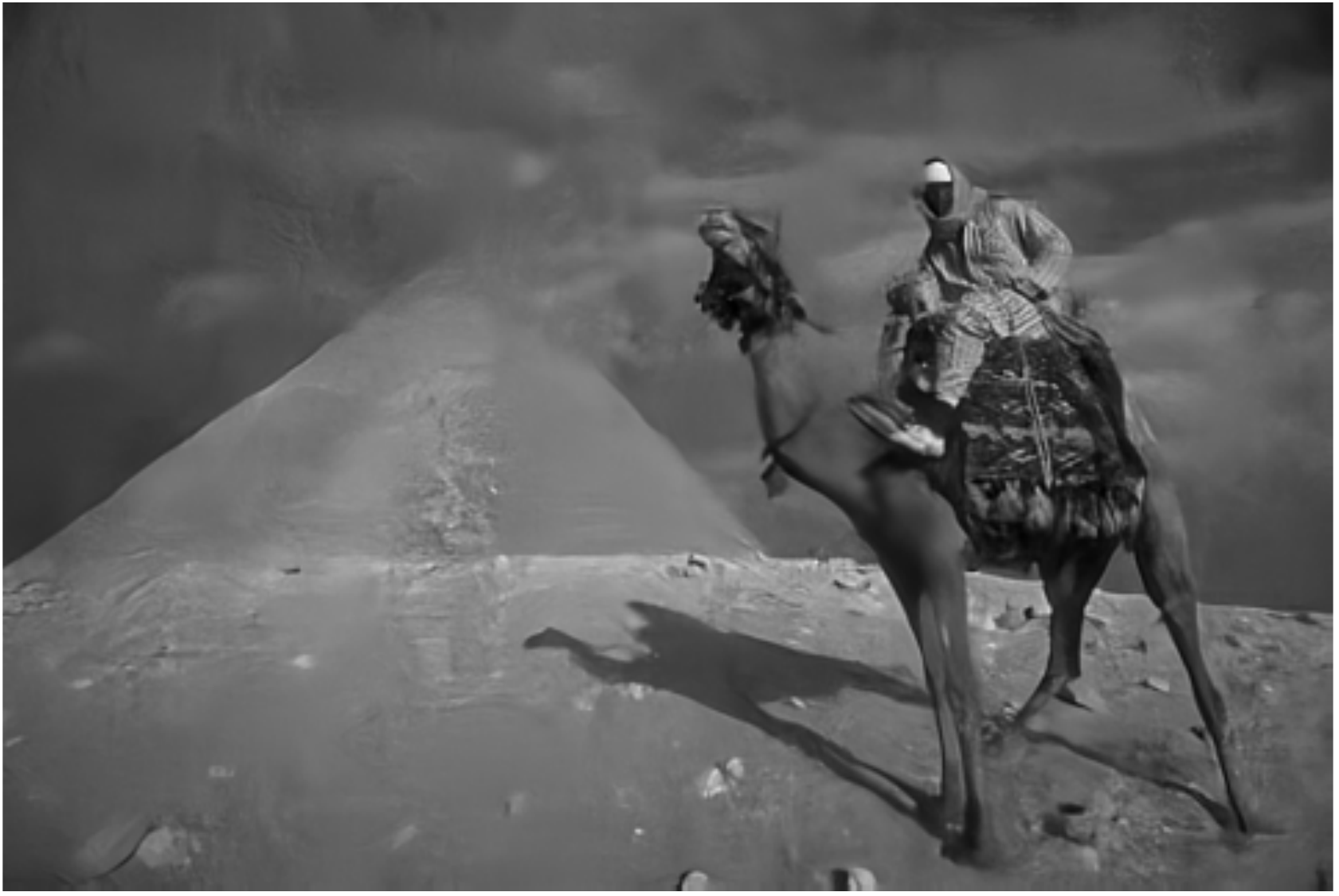}
  \put(79.5,48.5){\textcolor{red}{\fboxrule=0.5pt\fboxsep=0pt\fbox{\includegraphics[scale=.25,trim=1 2 2 1,clip=true]{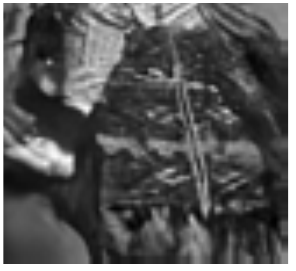}}}}
  \end{overpic}&    
 \begin{overpic}[width=.215\linewidth]{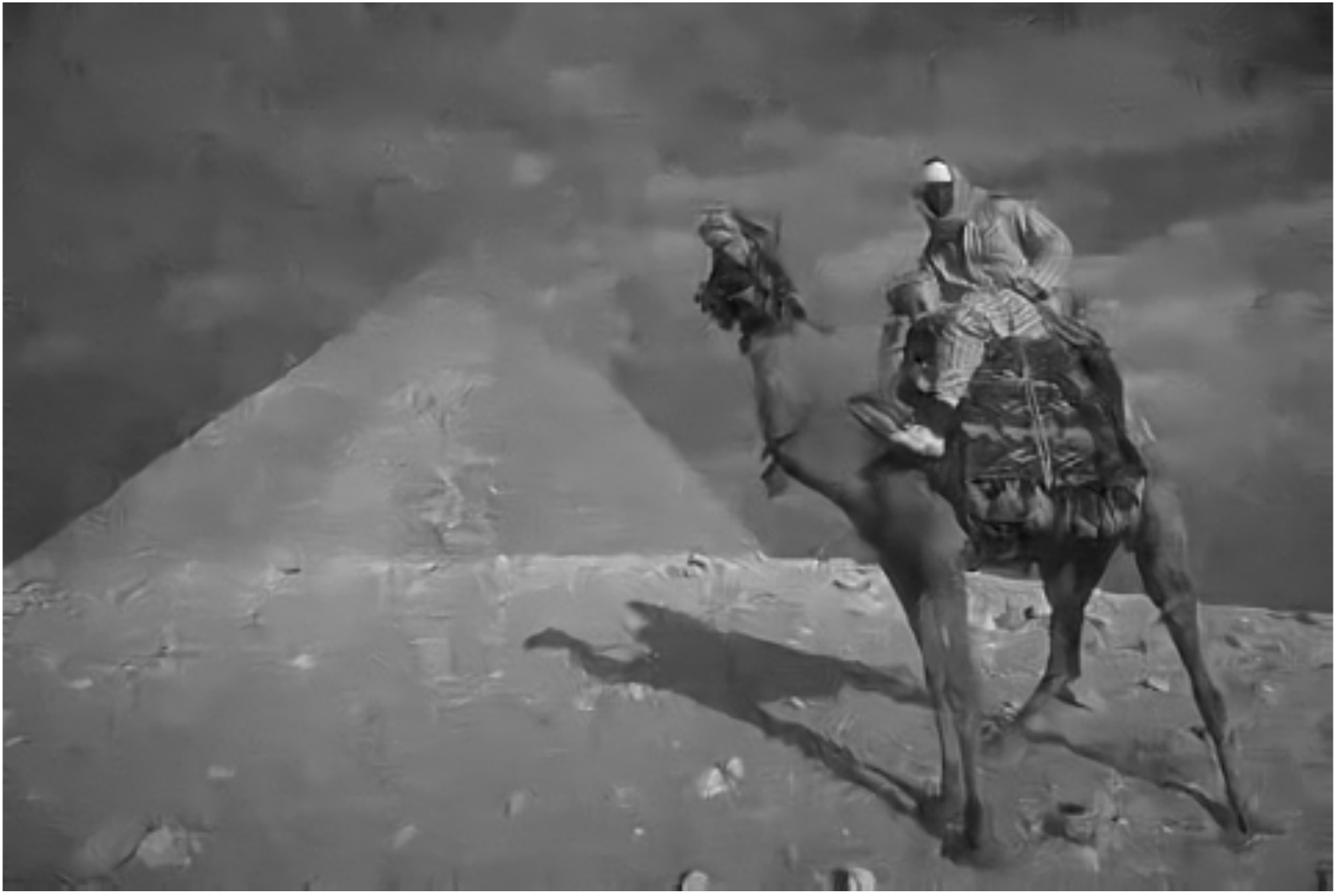}
  \put(79.5,48.5){\textcolor{red}{\fboxrule=0.5pt\fboxsep=0pt\fbox{\includegraphics[scale=.25,trim=1 2 2 1,clip=true]{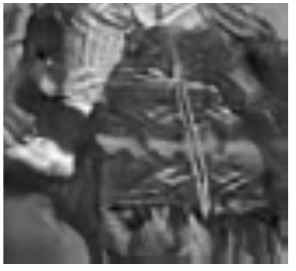}}}}
  \end{overpic}  \\   
  {\footnotesize (a)} & {\footnotesize (b)} &  {\footnotesize (e)} &  {\footnotesize (d)} &  {\footnotesize (e)}
\end{tabular}
   \caption{\footnotesize Grayscale image denoising. (a) Original image, (b) Noisy image (AWGN with $\sigma = 20$) ; $\operatorname{PSNR} = 22.10 \text{ dB}$. (c) Denoised image using $\operatorname{EPLL}$~\cite{Zoran2011} ; $\operatorname{PSNR} = 31.54 \text{ dB}$. (d) Denoised image using $\operatorname{DnCNN}$~\cite{Zhang2017} ; $\operatorname{PSNR} = 31.83 \text{ dB}$. (e) Denoised image using  $\operatorname{UNet}_5$; $\operatorname{PSNR} = 31.71 \text{ dB}$.\vspace{-0.2cm}}
   \label{fig:GrayComp}
   \vspace{-0.2cm}
\end{figure*}
\begin{figure*}[ht]
\centering
\begin{tabular}{@{} c @{} c @{} c @{} c @{} c @{}}
\hspace{-0.6cm}
 \begin{overpic}[width=.215\linewidth]{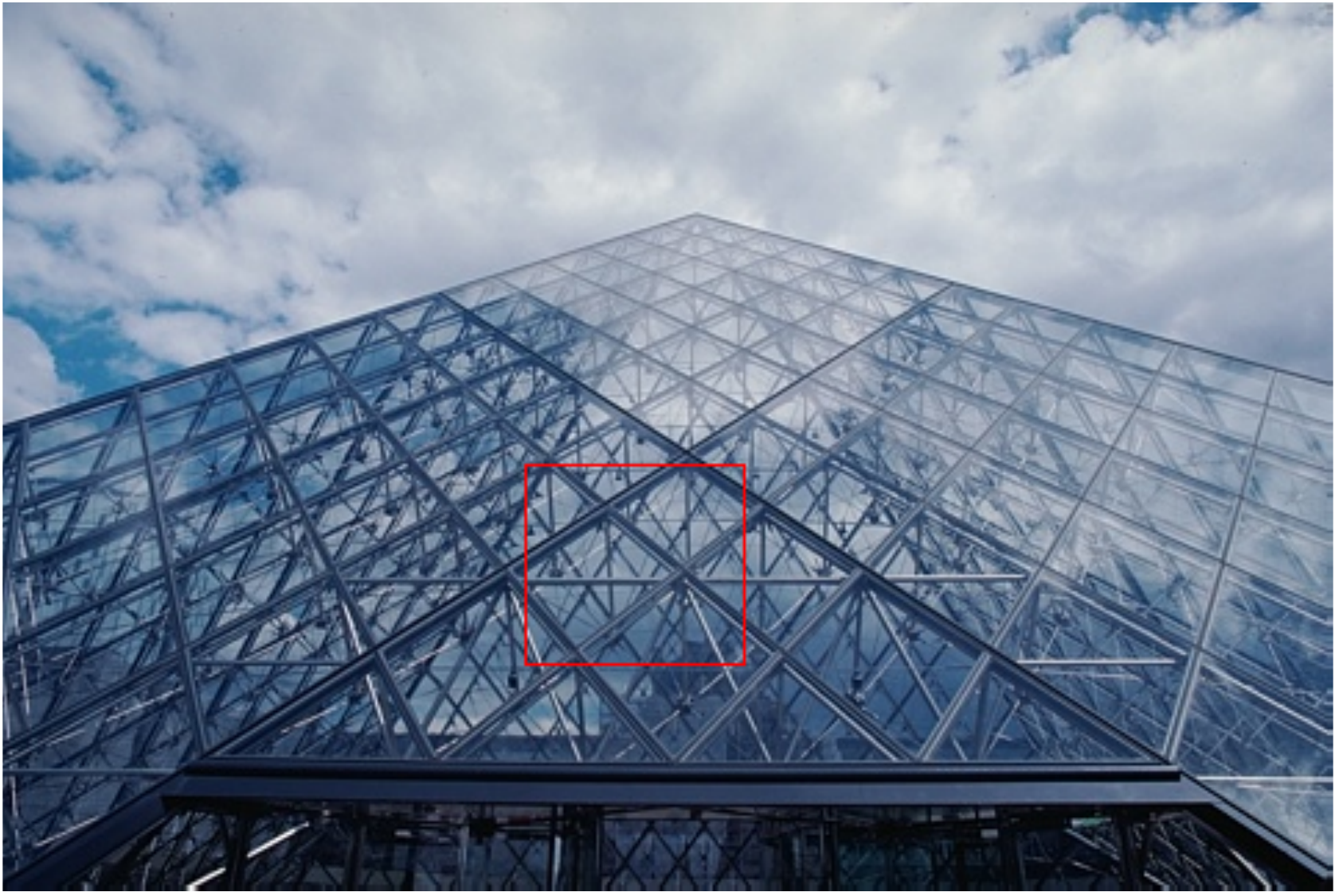}
  \put(79.5,48.5){\textcolor{red}{\fboxrule=0.5pt\fboxsep=0pt\fbox{\includegraphics[scale=.25,trim=1 2 2 1,clip=true]{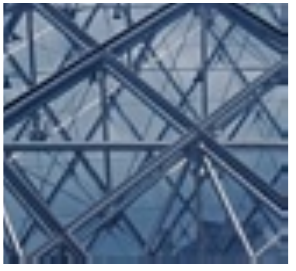}}}}
  \end{overpic}&
 \begin{overpic}[width=.215\linewidth]{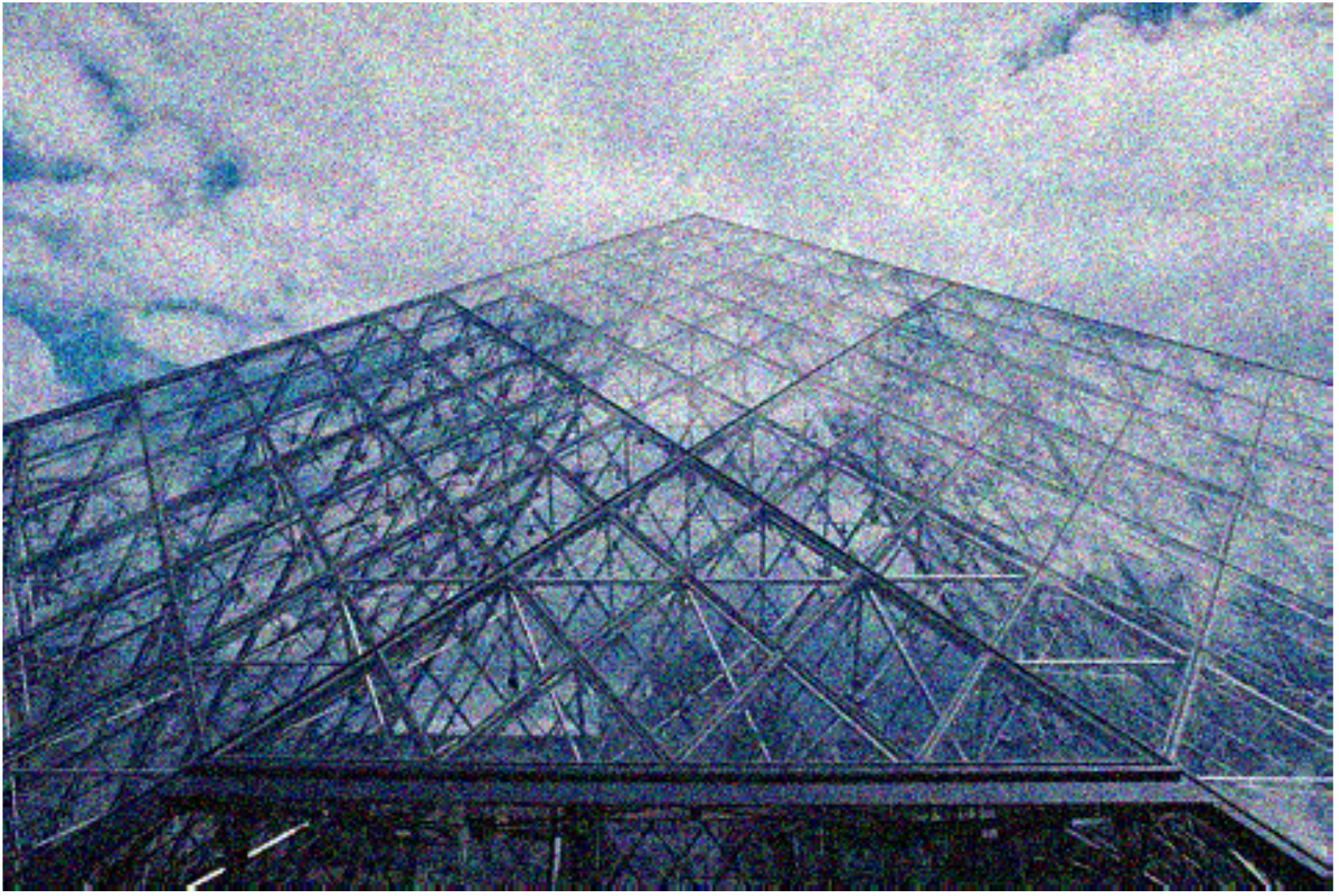}
  \put(79.5,48.5){\textcolor{red}{\fboxrule=0.5pt\fboxsep=0pt\fbox{\includegraphics[scale=.25,trim=1 2 2 1,clip=true]{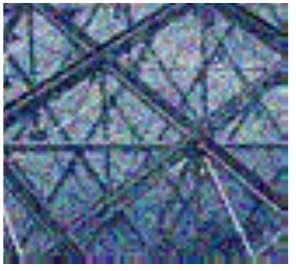}}}}
  \end{overpic}&  
 \begin{overpic}[width=.215\linewidth]{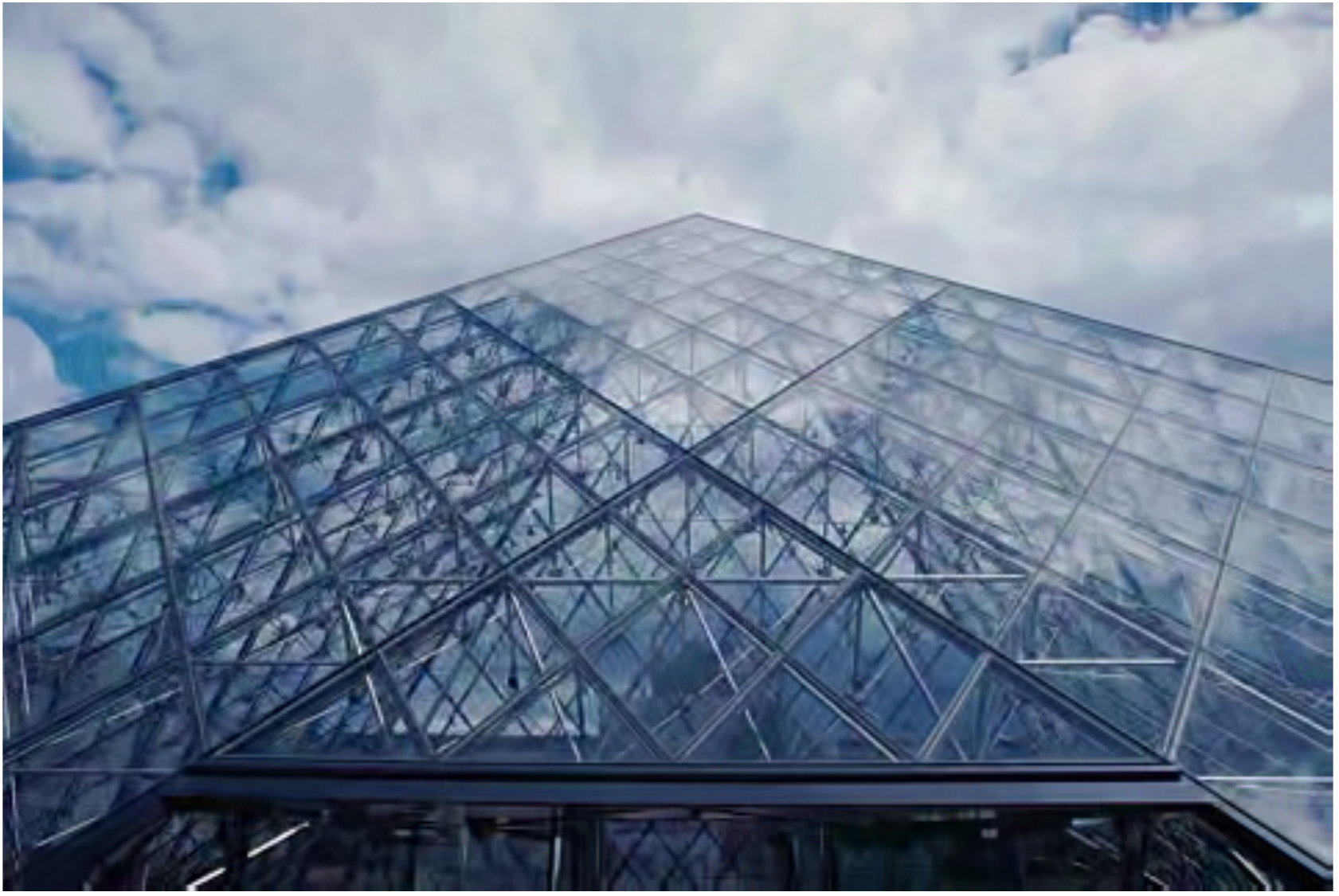}
  \put(79.5,48.5){\textcolor{red}{\fboxrule=0.5pt\fboxsep=0pt\fbox{\includegraphics[scale=.25,trim=1 2 2 1,clip=true]{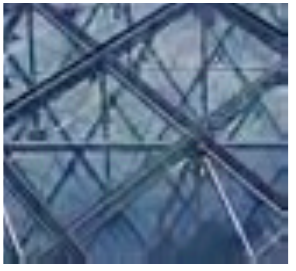}}}}
  \end{overpic}&    
 \begin{overpic}[width=.215\linewidth]{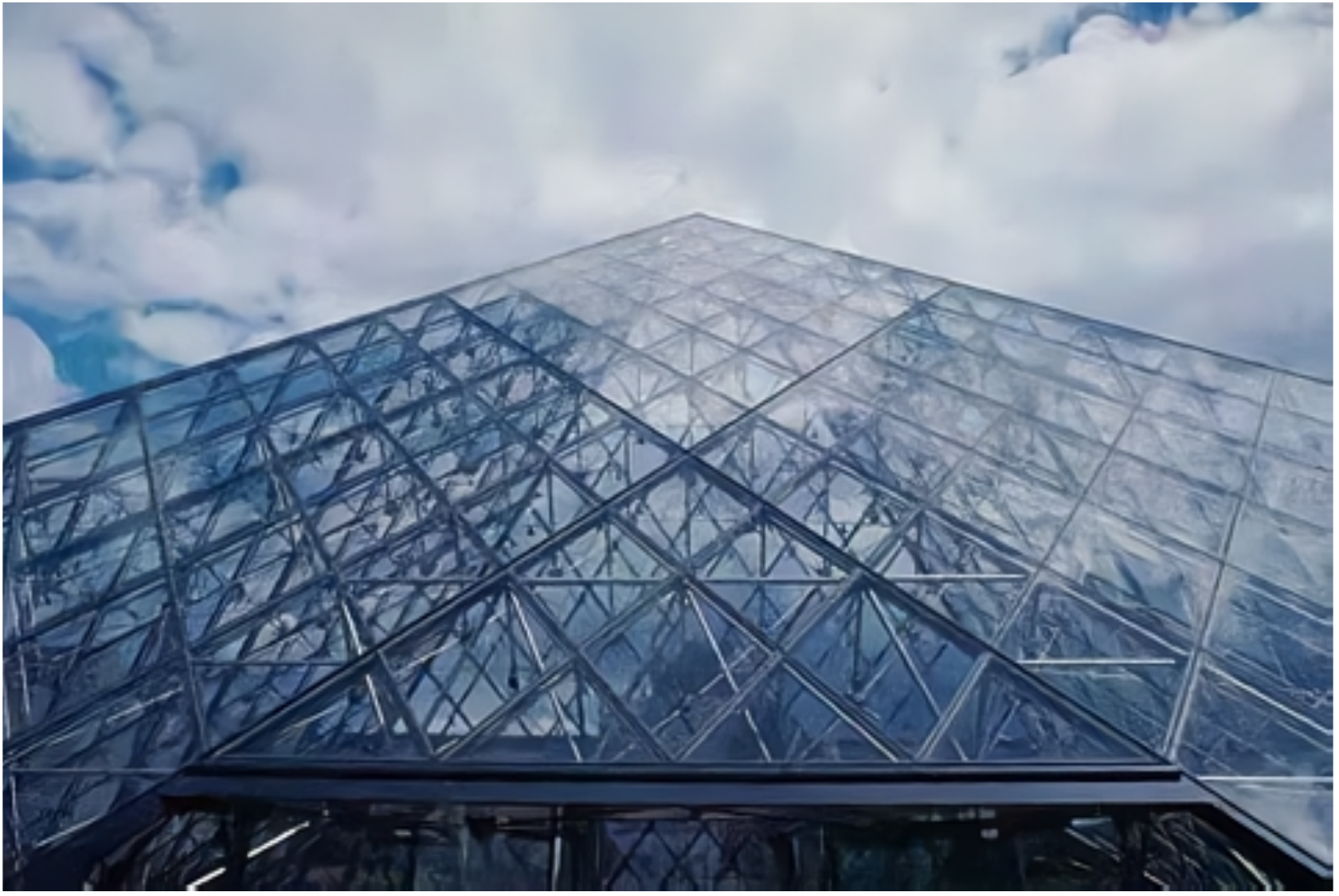}
  \put(79.5,48.5){\textcolor{red}{\fboxrule=0.5pt\fboxsep=0pt\fbox{\includegraphics[scale=.25,trim=1 2 2 1,clip=true]{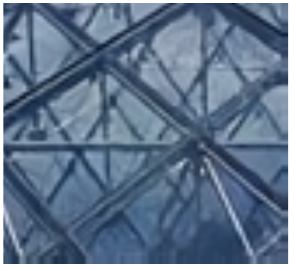}}}}
  \end{overpic}&    
 \begin{overpic}[width=.215\linewidth]{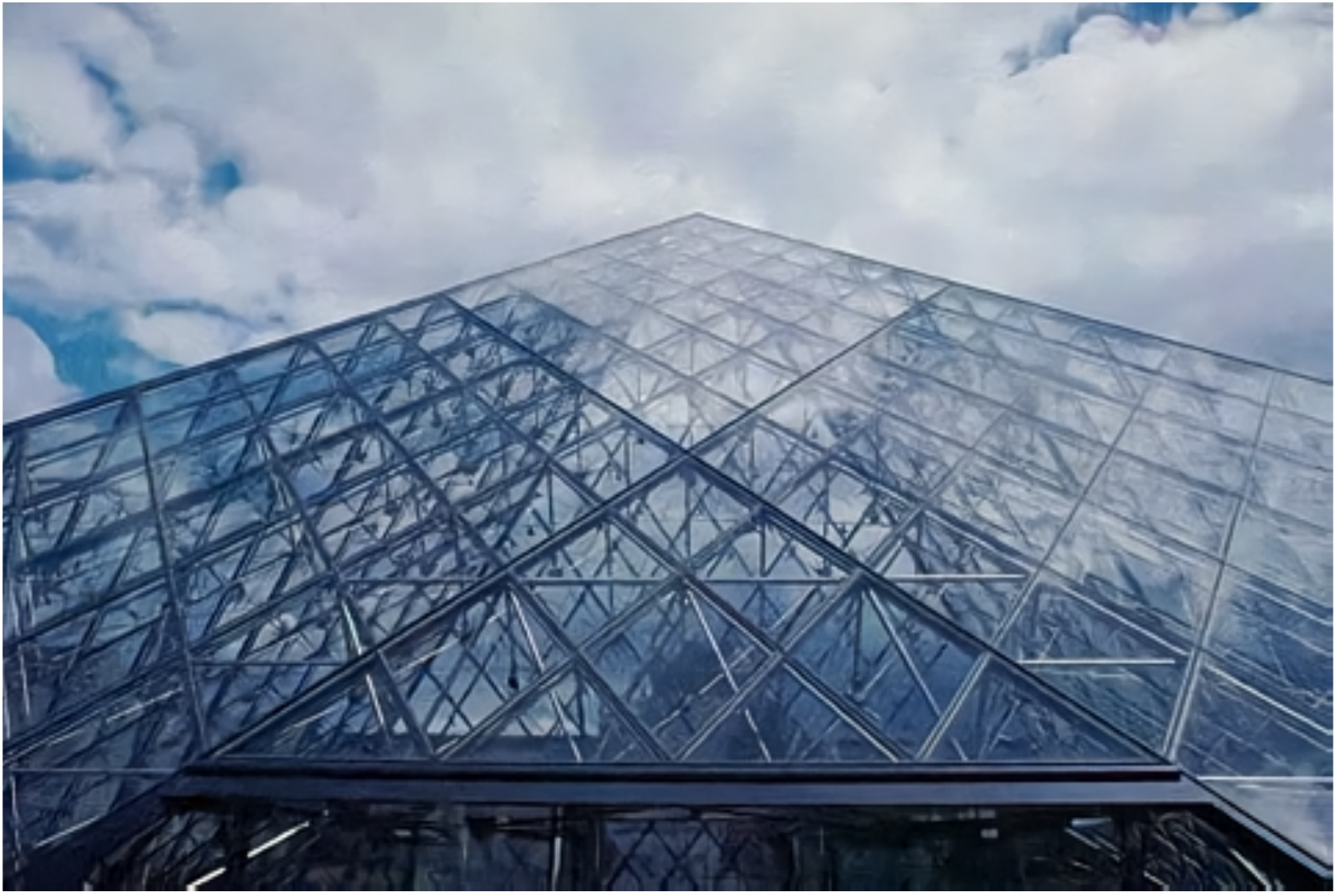}
  \put(79.5,48.5){\textcolor{red}{\fboxrule=0.5pt\fboxsep=0pt\fbox{\includegraphics[scale=.25,trim=1 2 2 1,clip=true]{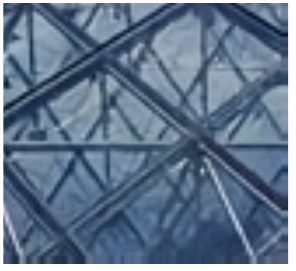}}}}
  \end{overpic}  \\   
      {\footnotesize (a)} & {\footnotesize (b)} &  {\footnotesize (e)} &  {\footnotesize (d)} &  {\footnotesize (e)}
\end{tabular}
   \caption{\footnotesize Color image denoising. (a) Original image, (b) Noisy image (AWGN with $\sigma = 30$) ; $\operatorname{PSNR} = 18.57 \text{ dB}$. (c) Denoised image using $\operatorname{CBM3D}$~\cite{Dabov2007} ; $\operatorname{PSNR} = 28.55 \text{ dB}$. (d) Denoised image using $\operatorname{CDnCNN}$~\cite{Zhang2017} ; $\operatorname{PSNR} = 29.08 \text{ dB}$. (e) Denoised image using  $\operatorname{CUNLNet}_5$; $\operatorname{PSNR} = 29.13 \text{ dB}$.}
   \label{fig:ColorComp}
\end{figure*}

\section{Network Training}
\label{sec:NetTrain}
We train our networks for grayscale and color image denoising under i.i.d Gaussian noise. Each network consists of a cascade of $S$ composite layers, as the one shown in Fig.~\ref{fig:CompositeLayer}, plus an additional clipping layer placed just before the output of the network. This last layer incorporates our prior knowledge about the valid range of image intensities and forces the pixel values of the restored image to lie in the range [0,\,255]. The network parameters $\bm{\Theta} = \br{\bm{\Theta}^1, \hdots, \bm{\Theta}^S}$, where $\bm{\Theta}^t = \cbr{s^t,\m{v}^t, \bm{g}^t, \bm{\pi}^t, \alpha^t}$\footnote{For the local variants of the proposed network, the parameters $\bm{g}^t$ are not present in the parameter set $\bm{\Theta}^t$.} denotes the set of parameters for the $t$-th layer, are learned using a loss-minimization strategy given $Q$ pairs of training data $\cbr{\m y_{q}, \m x_{q}}_{q=1}^{Q}$. Here $\m y_{q}$ is a noisy input and $\m x_{q}$ is the corresponding ground-truth image. To achieve an increased capacity for the network, we learn different parameters for each composite layer. However, the convolutional and non-local filtering layers (for the local and non-local version of the network, respectively) in each composite layer share the same parameters $\cbr{s^t,\m{v}^t,\bm{g}^t }$ with their transpose layers.  

Since the objective function to be minimized is non-convex, to avoid getting stuck in poor local-minima we initialize our networks with the parameters that are learned following a greedy-training strategy. The same approach has been adopted in~\cite{Schmidt2014,Chen2016,Lefkimmiatis2017} and it amounts to learning the parameters of each composite layer by keeping all the preceding layers of the network fixed and minimizing the cost 
\vspace{-.35cm}
\begin{equation}
\mc{L}\pr{\bm{\Theta}^t} = \suml_{q=1}^Q \ell\pr{\hat{\m{x}}_{q}^{t},\m x_{q}}.
\label{eq:Objective}
\vspace{-.25cm}
\end{equation}
In Eq.~\eqref{eq:Objective}, $\hat{\m{x}}_{q}^t$ is the output of the $t$-th composite layer and the loss function $\ell$ corresponds to the negative peak signal-to-noise-ratio (PSNR). This is computed as
$\ell\pr{\m y,\m x} =  -20\log_{10}\pr{p/\norm{\m y - \m x}{2}}$, where $p = 255\sqrt{N_t}$.
While these learned parameters are sub-optimal, we have experimentally observed that they serve as a good initialization for the joint optimization training that follows.

To minimize the objective function in Eq.~\eqref{eq:Objective} w.r.t the parameters $\bm \Theta^t$  we employ the Adam algorithm~\cite{Kingma2014}, which is a variant of the stochastic gradient descent (SGD) that involves adaptive normalization of the learning rate. Each layer is trained for 100 epochs using an initial learning rate 1e-2 (1e-3 for grayscale images), while the configuration parameters for Adam are chosen as {\small\verb!beta_1! = 0.9, \verb!beta_2! = 0.999} and {\small\verb!eps! = 1e-4}. 

The final parameters of our network are obtained by using the previous learned parameters as initial values and by jointly minimizing the objective function 
\vspace{-.3cm}
\begin{equation}
\mc{L}\pr{\bm{\Theta}} = \suml_{q=1}^Q \ell\pr{\hat{\m{x}}_{q}^{S},\m x_{q}},
\vspace{-.2cm}
\end{equation}
w.r.t to all network parameters $\bm{\Theta}$. This cost function does not take into account anymore the intermediate results (outputs of each composite layer) but only depends on the final output of the network $\hat{\m{x}}_{q}^{S}$. In this case the training is performed by running 100 epochs using Adam optimization with the same configuration parameters as before.

\vspace{-.2cm}\section{Experiments and Results}\vspace{-.1cm}
To train our local and non-local models we generated the training data using the Berkeley segmentation dataset (BSDS)~\cite{Martin2001}, which consists of 500 images. We split these images in two sets, a training set which consists of 400 images and the validation set which consists of the remaining 100 images. All the images were randomly cropped so that their size is $180 \times 180$ pixel.  We note that the 68 BSDS images of~\cite{Roth2009} that are used for the comparisons reported in Tables~\ref{tab:GrayComp} and ~\ref{tab:ColorComp} are strictly excluded from the training set and only cropped versions of them are used in the validation set. The proposed models were trained on a NVIDIA 1080 Ti GPU and the software we used for training and testing\footnote{The code implementing the proposed networks is available from the author's website.} was built on top of MatConvnet~\cite{Vedaldi2015}.
\renewcommand{\tabcolsep}{.1cm}
\begin{table*}[t]
\centering
 \begin{tabular}{ccccccccccccc}\hline\hline
\multicolumn{1}{c|}{}& \multicolumn{11}{c}{Noise level - $\sigma$ (std.)} \\

\multicolumn{1}{c|}{Methods} &
\multicolumn{1}{c|}{5} &
\multicolumn{1}{c|}{10} &
\multicolumn{1}{c|}{15} &
\multicolumn{1}{c|}{20} &
\multicolumn{1}{c|}{25} &
\multicolumn{1}{c|}{30} &
\multicolumn{1}{c|}{35} &
\multicolumn{1}{c|}{40} &
\multicolumn{1}{c|}{45} &
\multicolumn{1}{c|}{50} &
\multicolumn{1}{c||}{55} &
\multicolumn{1}{c}{avg.} \\
\cline{1-13}
\multicolumn{1}{c|}{\small BM3D~\cite{Dabov2007}} &
\multicolumn{1}{c|}{\small 37.57}&
\multicolumn{1}{c|}{\small 33.30}&
\multicolumn{1}{c|}{\small 31.06}&
\multicolumn{1}{c|}{\small 29.60}&
\multicolumn{1}{c|}{\small 28.55}&
\multicolumn{1}{c|}{\small 27.74}&
\multicolumn{1}{c|}{\small 27.07}&
\multicolumn{1}{c|}{\small 26.45}&
\multicolumn{1}{c|}{\small 25.99}&
\multicolumn{1}{c|}{\small 25.60}&
\multicolumn{1}{c||}{\small 25.26}&
\multicolumn{1}{c}{\small 28.93}\\

\multicolumn{1}{c|}{\small EPLL~\cite{Zoran2011}} &
\multicolumn{1}{c|}{\small 37.55 }&
\multicolumn{1}{c|}{\small 33.36}&
\multicolumn{1}{c|}{\small 31.18}&
\multicolumn{1}{c|}{\small 29.73}&
\multicolumn{1}{c|}{\small 28.67}&
\multicolumn{1}{c|}{\small 27.84}&
\multicolumn{1}{c|}{\small 27.16}&
\multicolumn{1}{c|}{\small 26.58}&
\multicolumn{1}{c|}{\small 26.09}&
\multicolumn{1}{c|}{\small 25.71}&
\multicolumn{1}{c||}{\small 25.34}&
\multicolumn{1}{c}{\small 29.02}\\

\multicolumn{1}{c|}{\small WNMM~\cite{Gu2014}} &
\multicolumn{1}{c|}{\small 37.76}&
\multicolumn{1}{c|}{\small 33.55}&
\multicolumn{1}{c|}{\small 31.31 }&
\multicolumn{1}{c|}{\small 29.83}&
\multicolumn{1}{c|}{\small 28.73}&
\multicolumn{1}{c|}{\small 27.94}&
\multicolumn{1}{c|}{\small 27.28}&
\multicolumn{1}{c|}{\small 26.72}&
\multicolumn{1}{c|}{\small 26.26}&
\multicolumn{1}{c|}{\small 25.85}&
\multicolumn{1}{c||}{\small 25.49}&
\multicolumn{1}{c}{\small 29.16}\\


\multicolumn{1}{c|}{\small DnCNN~\cite{Zhang2017}} &
\multicolumn{1}{c|}{\small 37.68}&
\multicolumn{1}{c|}{\small 33.72}&
\multicolumn{1}{c|}{\small 31.60}&
\multicolumn{1}{c|}{\small 30.19}&
\multicolumn{1}{c|}{\small 29.15}&
\multicolumn{1}{c|}{\small 28.33}&
\multicolumn{1}{c|}{\small 27.66}&
\multicolumn{1}{c|}{\small 27.10}&
\multicolumn{1}{c|}{\small 26.62}&
\multicolumn{1}{c|}{\small 26.21}&
\multicolumn{1}{c||}{\small 25.80}&
\multicolumn{1}{c}{\small 29.46}\\

\multicolumn{1}{c|}{\small $\mathrm{UNet}_5$}&
\multicolumn{1}{c|}{\small 37.59}&
\multicolumn{1}{c|}{\small 33.54}&
\multicolumn{1}{c|}{\small 31.38}&
\multicolumn{1}{c|}{\small 29.93}&
\multicolumn{1}{c|}{\small 28.84}&
\multicolumn{1}{c|}{\small 28.01}&
\multicolumn{1}{c|}{\small 27.38}&
\multicolumn{1}{c|}{\small 26.85}&
\multicolumn{1}{c|}{\small 26.38}&
\multicolumn{1}{c|}{\small 25.95}&
\multicolumn{1}{c||}{\small 25.53}&
\multicolumn{1}{c}{\small 29.22}\\

\multicolumn{1}{c|}{\small $\mathrm{UNLNet}_5$}&
\multicolumn{1}{c|}{\small 37.62}&
\multicolumn{1}{c|}{\small 33.62}&
\multicolumn{1}{c|}{\small 31.47}&
\multicolumn{1}{c|}{\small 30.04}&
\multicolumn{1}{c|}{\small 28.96}&
\multicolumn{1}{c|}{\small 28.13}&
\multicolumn{1}{c|}{\small 27.50}&
\multicolumn{1}{c|}{\small 26.96}&
\multicolumn{1}{c|}{\small 26.48}&
\multicolumn{1}{c|}{\small 26.04}&
\multicolumn{1}{c||}{\small 25.64}&
\multicolumn{1}{c}{\small 29.32}\\

\multicolumn{1}{c|}{\small $\mathrm{UNLNet}_5^{\mathrm{orc}}$}&
\multicolumn{1}{c|}{\small \color{red}{37.79}}&
\multicolumn{1}{c|}{\small \color{red}{33.97}}&
\multicolumn{1}{c|}{\small \color{red}{31.95}}&
\multicolumn{1}{c|}{\small \color{red}{30.59}}&
\multicolumn{1}{c|}{\small \color{red}{29.51}}&
\multicolumn{1}{c|}{\small \color{red}{28.54}}&
\multicolumn{1}{c|}{\small \color{red}{27.97}}&
\multicolumn{1}{c|}{\small \color{red}{27.47}}&
\multicolumn{1}{c|}{\small \color{red}{26.97}}&
\multicolumn{1}{c|}{\small \color{red}{26.41}}&
\multicolumn{1}{c||}{\small \color{red}{25.80}}&
\multicolumn{1}{c}{\small \color{red}{29.72}}\\

\hline\hline
\end{tabular}
\\
\vspace{.08cm}
\caption{Grayscale denoising comparisons for different noise levels over the standard set of 68~\cite{Roth2009} Berkeley images. Performance is measured in terms of average PSNR (in dB). The highlighted results refer to those of the non-local model with oracle grouping.}
\label{tab:GrayComp}
\vspace{-.25cm}
\end{table*}

\renewcommand{\tabcolsep}{.05cm}
\begin{table}[t]
\centering
 \begin{tabular}{cccccccc}\hline\hline
\multicolumn{2}{c|}{Noise}& \multicolumn{5}{c}{Methods} \\
\multicolumn{2}{c|}{$\sigma$ (std.)}&
\multicolumn{1}{c|}{\footnotesize CBM3D~\cite{Dabov2007}}&
\multicolumn{1}{c|}{\footnotesize CDnCNN~\cite{Zhang2017}} &
\multicolumn{1}{c|}{\footnotesize $\mathrm{CUNet}_5$}&
\multicolumn{1}{c|}{\footnotesize $\mathrm{CUNLNet}_5$}&
\multicolumn{1}{c}{\footnotesize $\mathrm{CUNLNet}_5^{\mathrm{orc}}$}\\
\hline

\multicolumn{2}{c|}{\small 5}
&\multicolumn{1}{c|}{\small 40.24}
&\multicolumn{1}{c|}{\small 40.11}
&\multicolumn{1}{c|}{\small 40.31}
&\multicolumn{1}{c|}{\small 40.39}
&\multicolumn{1}{c}{\small \color{red}{40.54}}\\

\multicolumn{2}{c|}{\small 10}
&\multicolumn{1}{c|}{\small 35.88}
&\multicolumn{1}{c|}{\small 36.11}
&\multicolumn{1}{c|}{\small 36.08}
&\multicolumn{1}{c|}{\small 36.20}
&\multicolumn{1}{c}{\small \color{red}{36.70}}\\

\multicolumn{2}{c|}{\small 15}
&\multicolumn{1}{c|}{\small 33.49}
&\multicolumn{1}{c|}{\small 33.88}
&\multicolumn{1}{c|}{\small 33.78}
&\multicolumn{1}{c|}{\small 33.90}
&\multicolumn{1}{c}{\small \color{red}{34.58}}\\

\multicolumn{2}{c|}{\small 20}
&\multicolumn{1}{c|}{\small 31.88}
&\multicolumn{1}{c|}{\small 32.36}
&\multicolumn{1}{c|}{\small 32.21}
&\multicolumn{1}{c|}{\small 32.34}
&\multicolumn{1}{c}{\small \color{red}{33.11}}\\

\multicolumn{2}{c|}{\small 25}
&\multicolumn{1}{c|}{\small 30.68}
&\multicolumn{1}{c|}{\small 31.22}
&\multicolumn{1}{c|}{\small 31.03}
&\multicolumn{1}{c|}{\small 31.17}
&\multicolumn{1}{c}{\small \color{red}{31.95}}\\

\multicolumn{2}{c|}{\small 30}
&\multicolumn{1}{c|}{\small 29.71}
&\multicolumn{1}{c|}{\small 30.31}
&\multicolumn{1}{c|}{\small 30.06}
&\multicolumn{1}{c|}{\small 30.24}
&\multicolumn{1}{c}{\small \color{red}{31.06}}\\

\multicolumn{2}{c|}{\small 35}
&\multicolumn{1}{c|}{\small 28.86}
&\multicolumn{1}{c|}{\small 29.57}
&\multicolumn{1}{c|}{\small 29.37}
&\multicolumn{1}{c|}{\small 29.53}
&\multicolumn{1}{c}{\small \color{red}{30.37}}\\

\multicolumn{2}{c|}{\small 40}
&\multicolumn{1}{c|}{\small 28.06}
&\multicolumn{1}{c|}{\small 28.94}
&\multicolumn{1}{c|}{\small 28.77}
&\multicolumn{1}{c|}{\small 28.91}
&\multicolumn{1}{c}{\small \color{red}{29.75}}\\

\multicolumn{2}{c|}{\small 45}
&\multicolumn{1}{c|}{\small 27.82}
&\multicolumn{1}{c|}{\small 28.39}
&\multicolumn{1}{c|}{\small 28.23}
&\multicolumn{1}{c|}{\small 28.37}
&\multicolumn{1}{c}{\small \color{red}{29.19}}\\

\multicolumn{2}{c|}{\small 50}
&\multicolumn{1}{c|}{\small 27.36}
&\multicolumn{1}{c|}{\small 27.91}
&\multicolumn{1}{c|}{\small 27.74}
&\multicolumn{1}{c|}{\small 27.89}
&\multicolumn{1}{c}{\small \color{red}{28.65}}\\

\multicolumn{2}{c|}{\small 55}
&\multicolumn{1}{c|}{\small 26.95}
&\multicolumn{1}{c|}{\small 27.45}
&\multicolumn{1}{c|}{\small 27.27}
&\multicolumn{1}{c|}{\small 27.44}
&\multicolumn{1}{c}{\small \color{red}{28.10}}\\
\hline
\multicolumn{2}{c|}{\small avg.}
&\multicolumn{1}{c|}{\small 30.99}
&\multicolumn{1}{c|}{\small 31.48}
&\multicolumn{1}{c|}{\small 31.35}
&\multicolumn{1}{c|}{\small 31.49}
&\multicolumn{1}{c}{\small \color{red}{32.18}}\\
\hline\hline
\end{tabular}
\\
\vspace{.08cm}
\caption{Color denoising comparisons for different noise levels over the standard set of 68~\cite{Roth2009} Berkeley images. Performance is measured in terms of average PSNR (in dB). The highlighted results refer to those of the non-local model with oracle grouping.}
\label{tab:ColorComp}
\vspace{-.5cm}
\end{table}
\vspace{-.5cm}
\paragraph{Grayscale denoising} Following the strategy described in Section~\ref{sec:NetTrain}, we have trained two variants of our proposed network. In the first network, we parametrize the regularization operator $\m L$ as a local operator and in the second one as a non-local operator. Both networks consist of $S=5$ composite layers each and we will refer to them as ${\operatorname{\footnotesize UNet}_5}$ and $\operatorname{UNLNet}_5$, respectively. 
For the local model, in order to parametrize the operator $\m L$, in each composite layer we employ a convolution layer of 48 filters, which are zero-mean and have a $7\times 7$ support. For the non-local model, instead of the convolution layer we utilize the non-local filtering layer as described in Section~\ref{sec:OperatorModeling}. In this case, similar to the local network, we utilize 48 filters of $7\times 7$ support. As explained in Section~\ref{sec:OperatorModeling} this corresponds to applying a non-redundant linear transformation, $\mc{T}: \R^{7\times 7} \mapsto \R^{48}$ on every image patch of size $7\times 7$ extracted from the input image, excluding the DC component (the low-pass content) of the transform. 
Finally, in order to form the group of similar patches as required by the second step of the non-local filtering layer, we use the $P=8$ closest neighbors (including the reference patch) while the similar patches are searched on the noisy input of the network in a window of $31 \times 31$ centered around each pixel. The same group indices are then used for all the composite layers of the network. 

We trained two $\operatorname{\footnotesize UNet}_5$ and $\operatorname{\footnotesize UNLNet}_5$ networks, one for low input noise levels ($\sigma < 30$) and one for high input noise levels ($ 30 \le \sigma < 55$). For the low-noise network training, the training data were distorted with AWGN of standard deviation that varies from $\sigma = 5$ to $\sigma = 29$ with increments of 4
and for the high-noise network with AWGN of standard deviation that varies from $\sigma = 30$ to $\sigma = 55$ using the same increments.
To evaluate the restoration performance of our proposed networks, in Table~\ref{tab:GrayComp} we report comparisons with recent state-of-the-art denoising methods on the standard evaluation dataset of 68 images~\cite{Roth2009} for eleven different noise levels, where the standard deviation of the noise varies from $\sigma = 5$ to $\sigma = 55$ with increments of 5.

\begin{figure*}[t]
\centering
\begin{tabular}{@{} c @{} c @{} c @{} c @{} c @{} }
   \includegraphics[width=.2\linewidth]{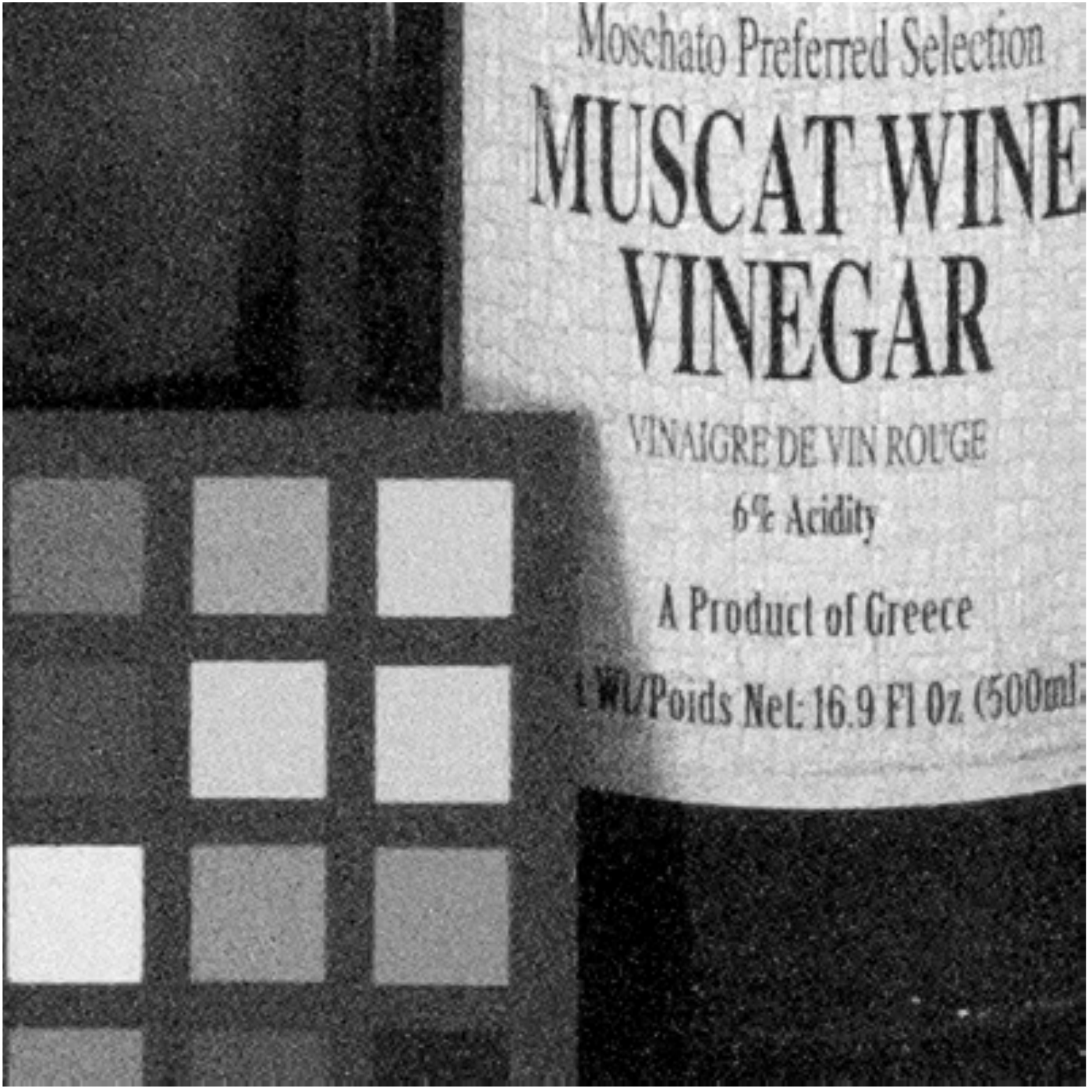}&
   \includegraphics[width=.2\linewidth]{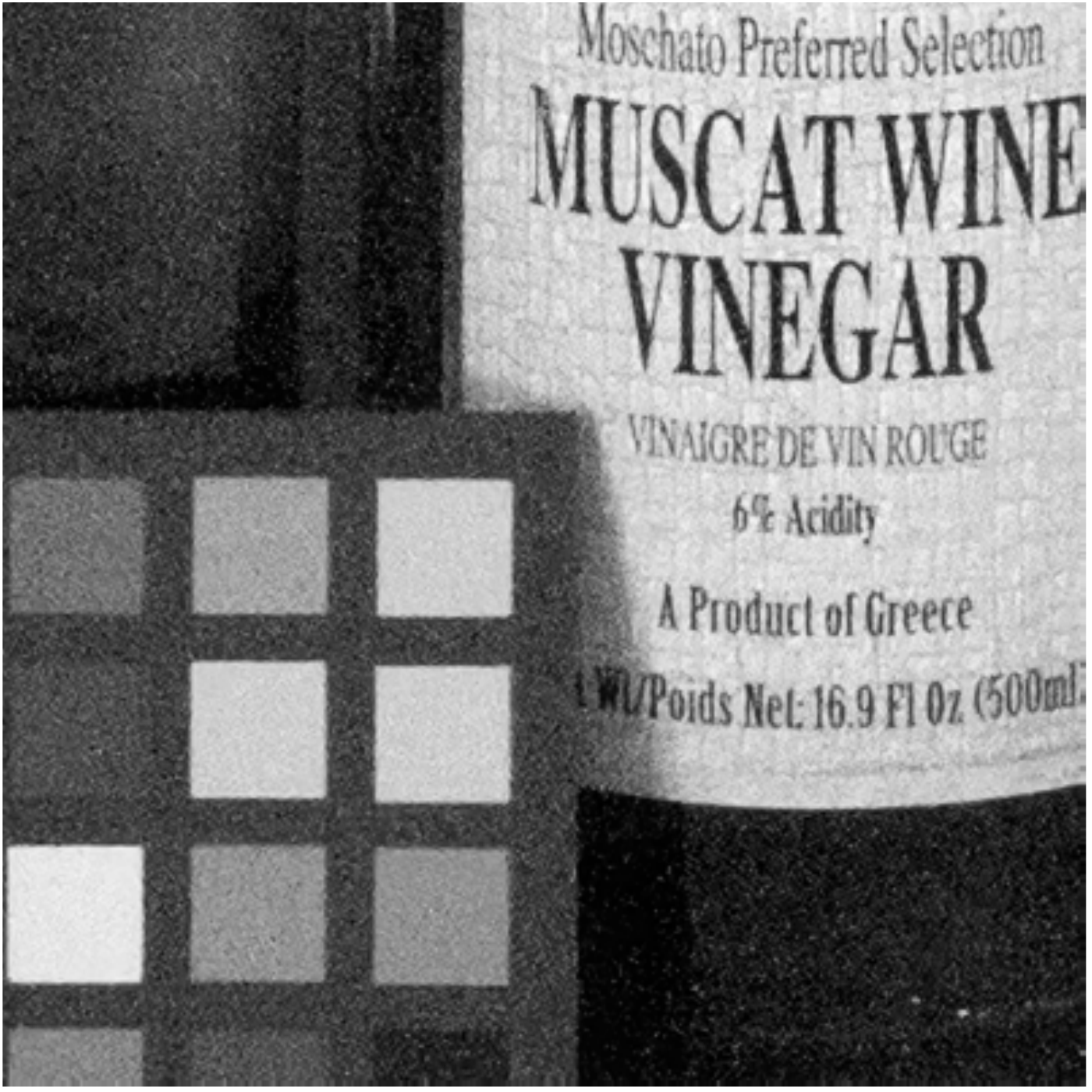}&   
   \includegraphics[width=.2\linewidth]{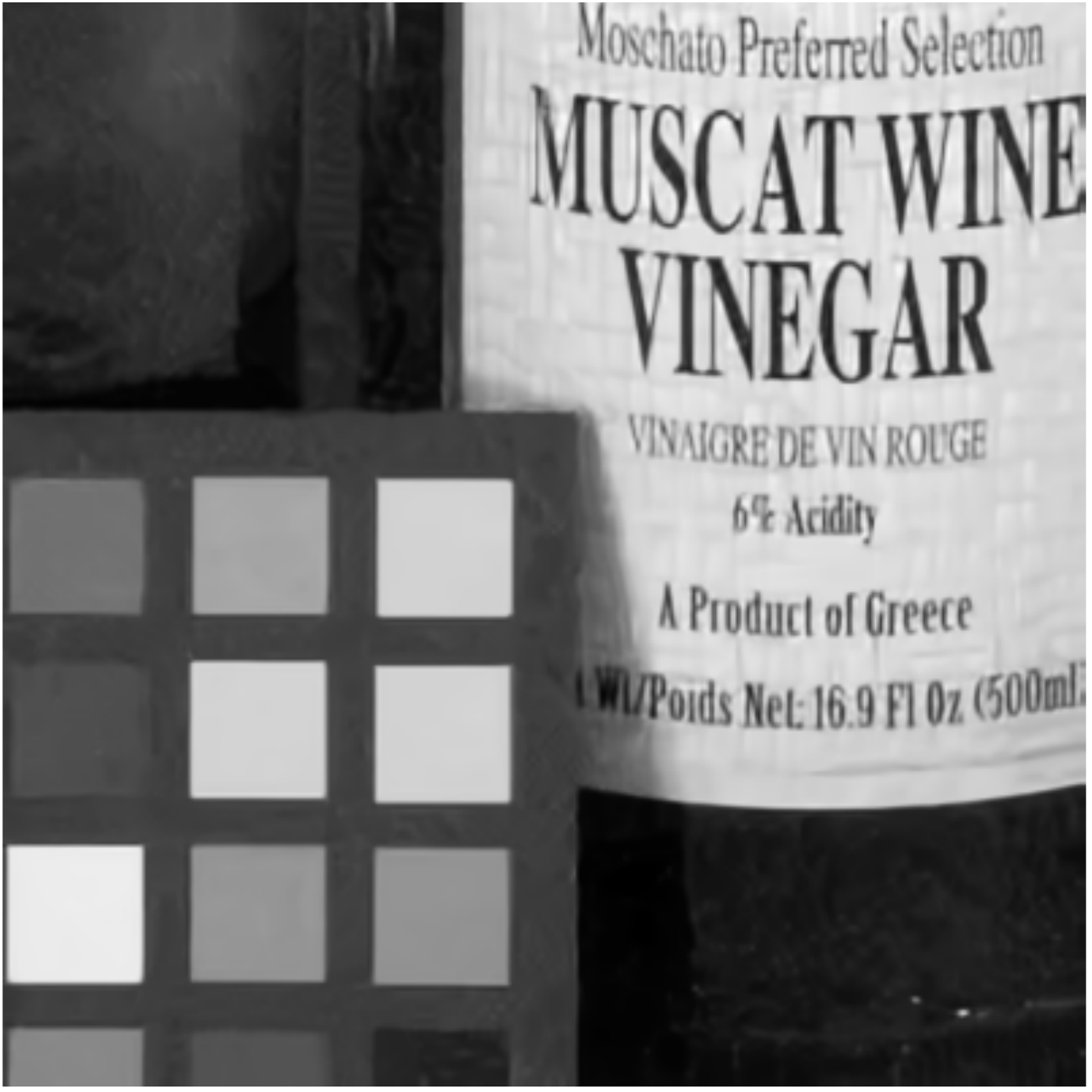}&
   \includegraphics[width=.2\linewidth]{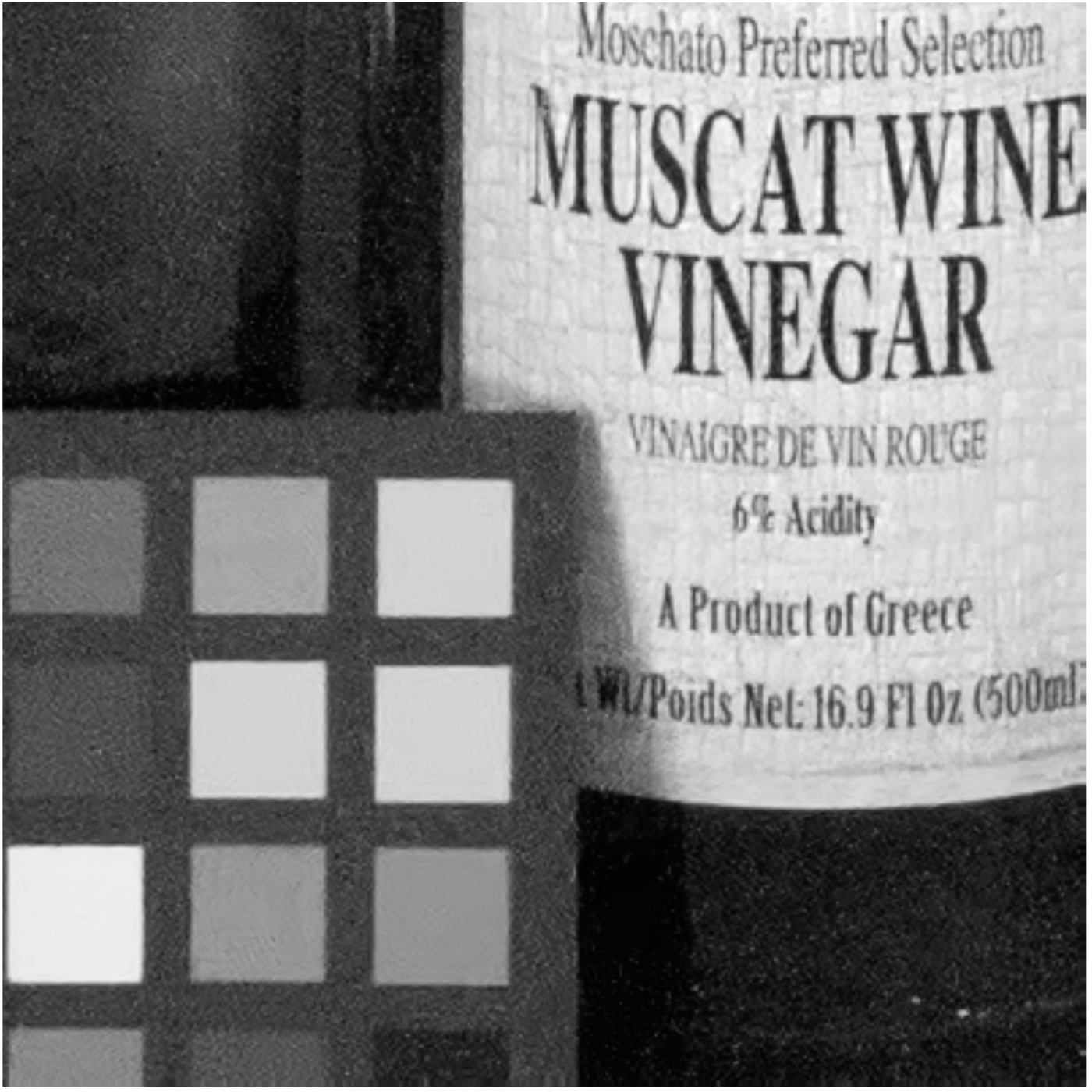}&
   \includegraphics[width=.2\linewidth]{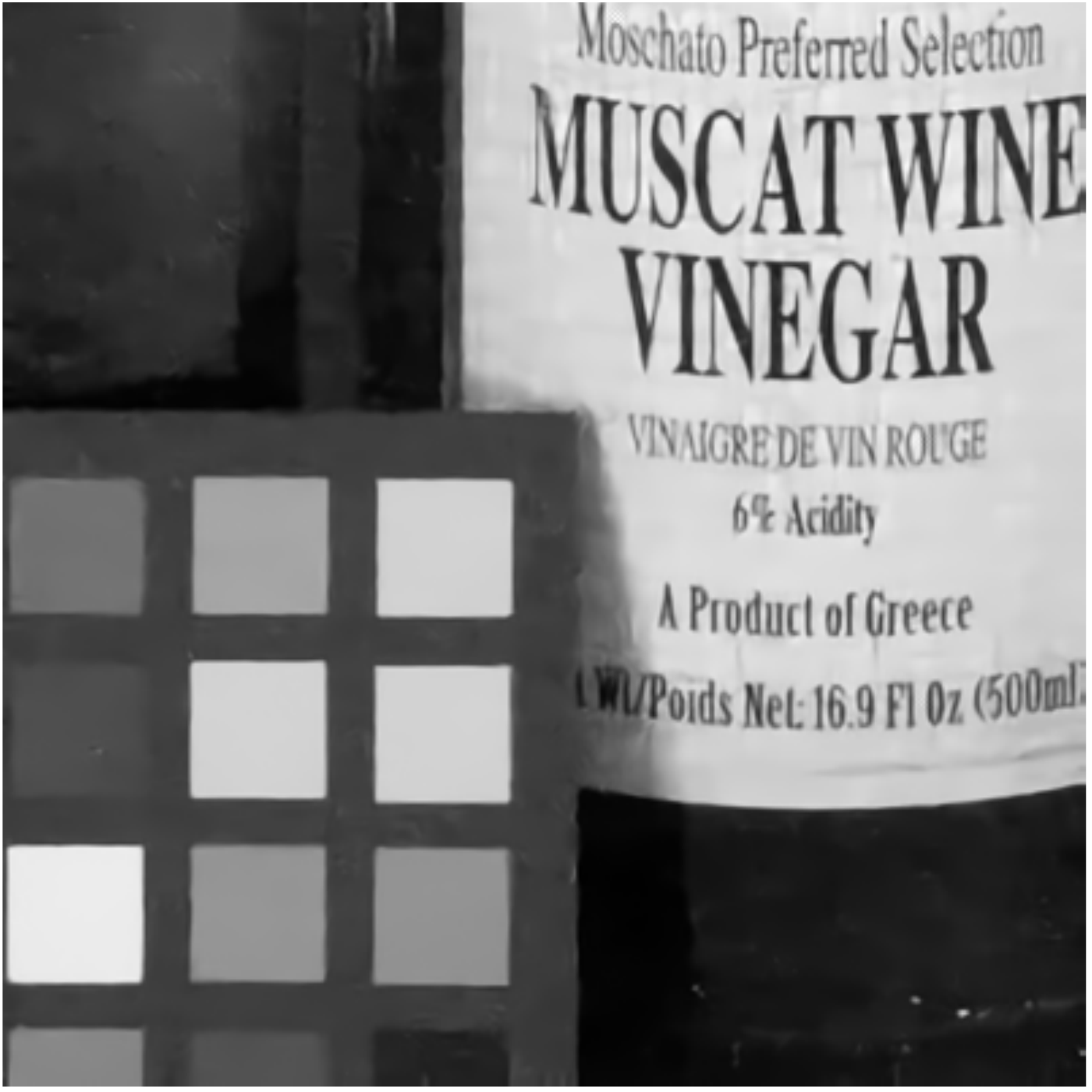}\vspace{-0.15cm}\\
  {\footnotesize (a) Noisy image ($\sigma = 15$)} & \footnotesize{(b) DnCNN~\cite{Zhang2017} }& \footnotesize{(c) BM3D~\cite{Dabov2007}} & \footnotesize{(d) Noise Clinic~\cite{Lebrun2015}} & \footnotesize{(e) $\operatorname{UNet}_5$} \\ 
   \includegraphics[width=.2\linewidth]{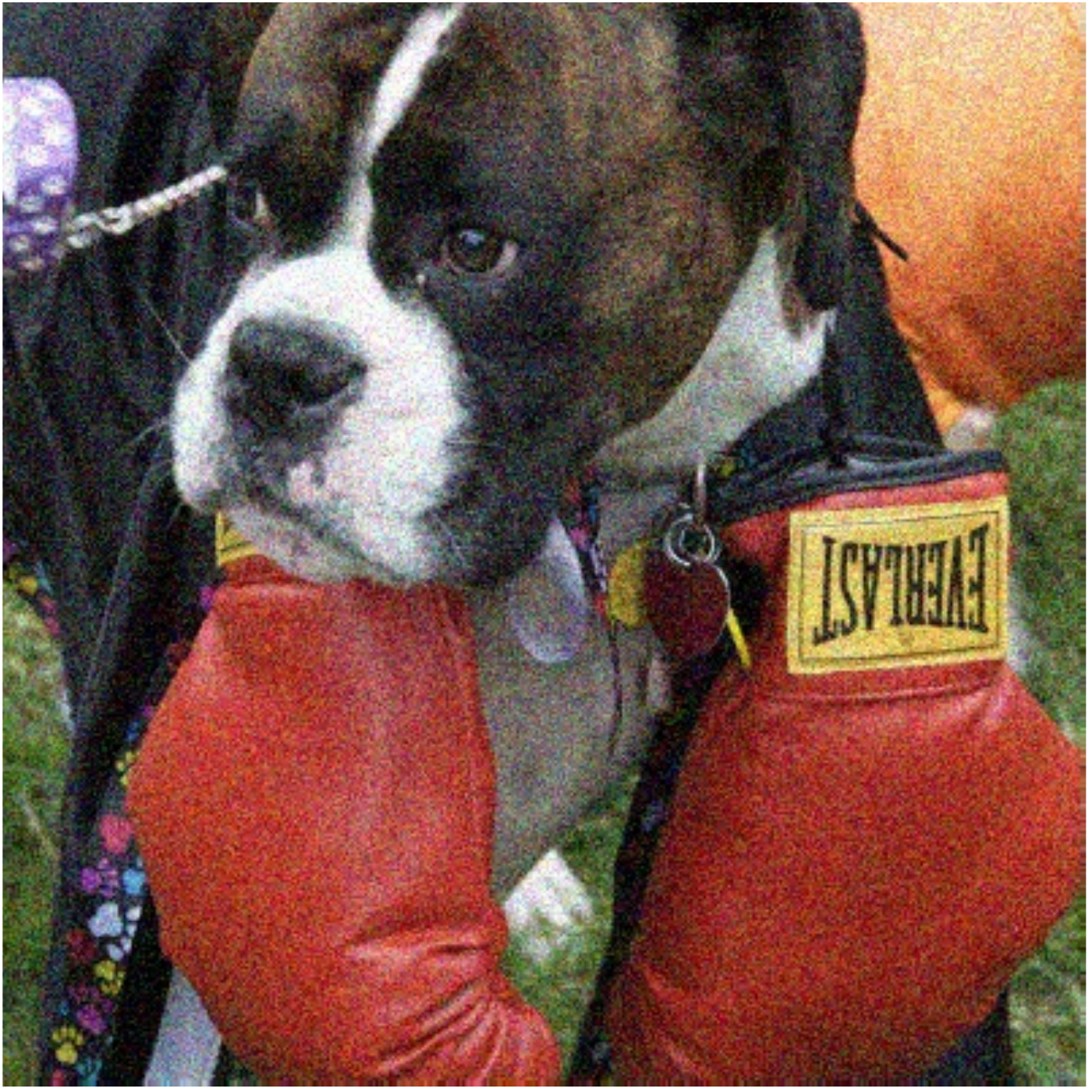}&
   \includegraphics[width=.2\linewidth]{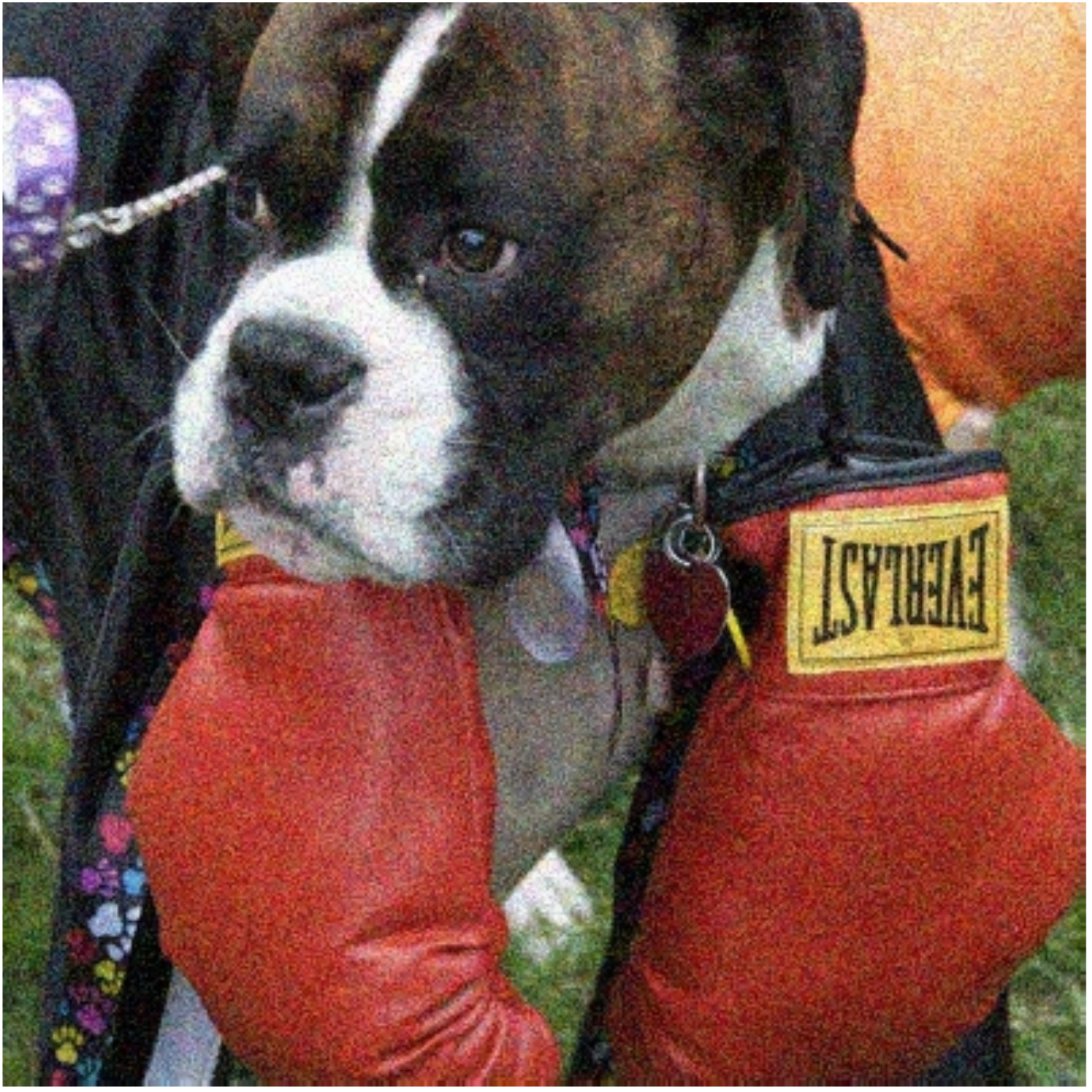}&   
   \includegraphics[width=.2\linewidth]{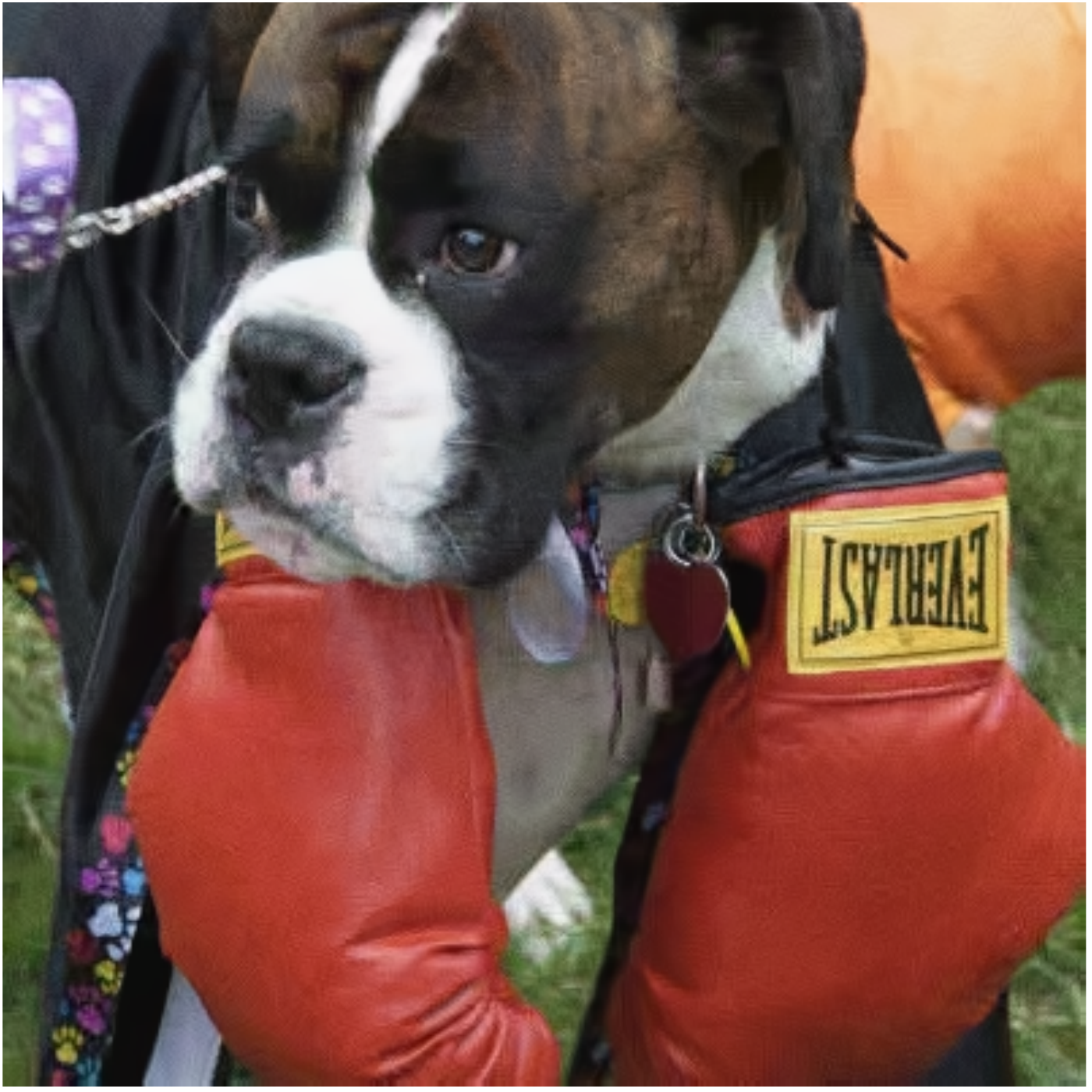}&
   \includegraphics[width=.2\linewidth]{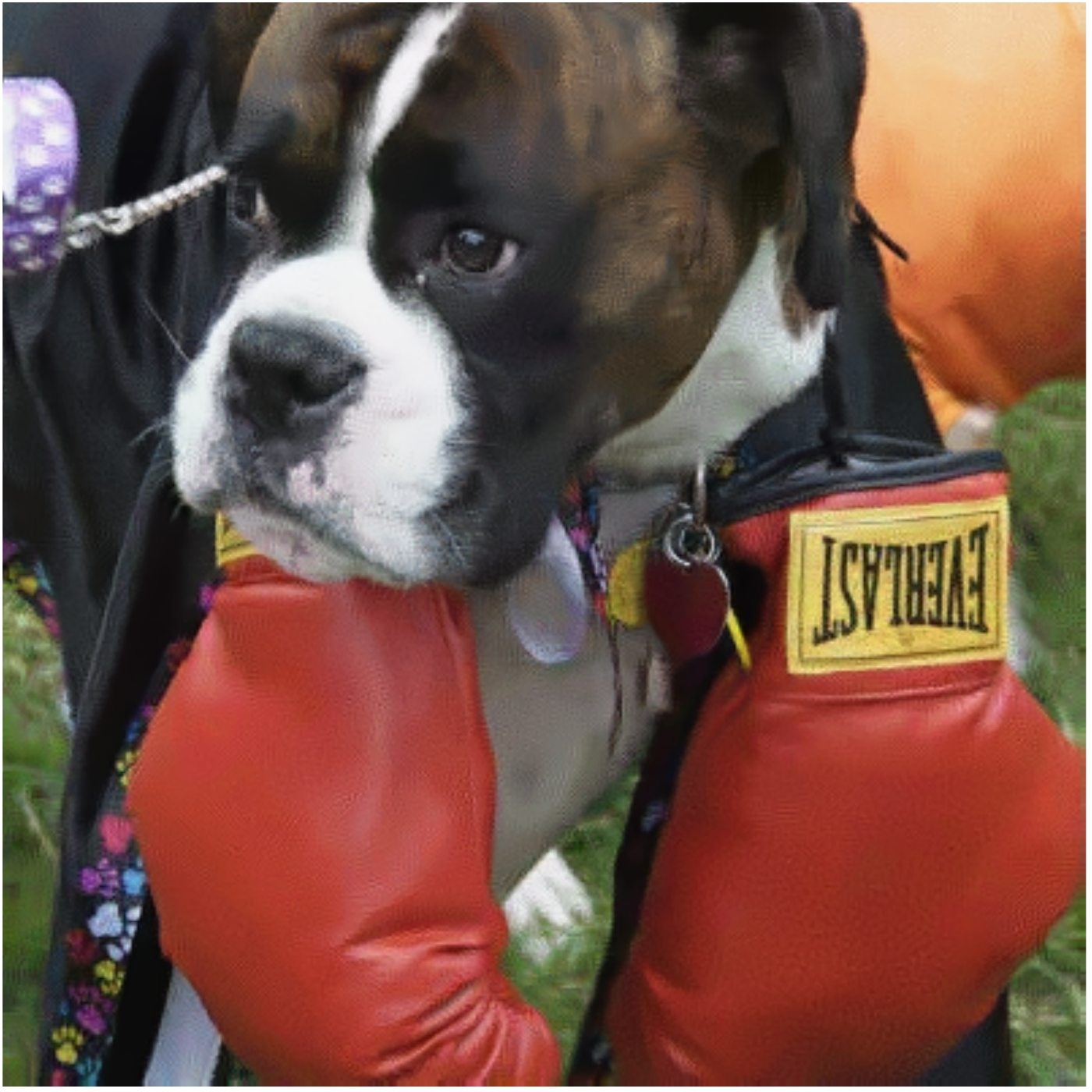}&
   \includegraphics[width=.2\linewidth]{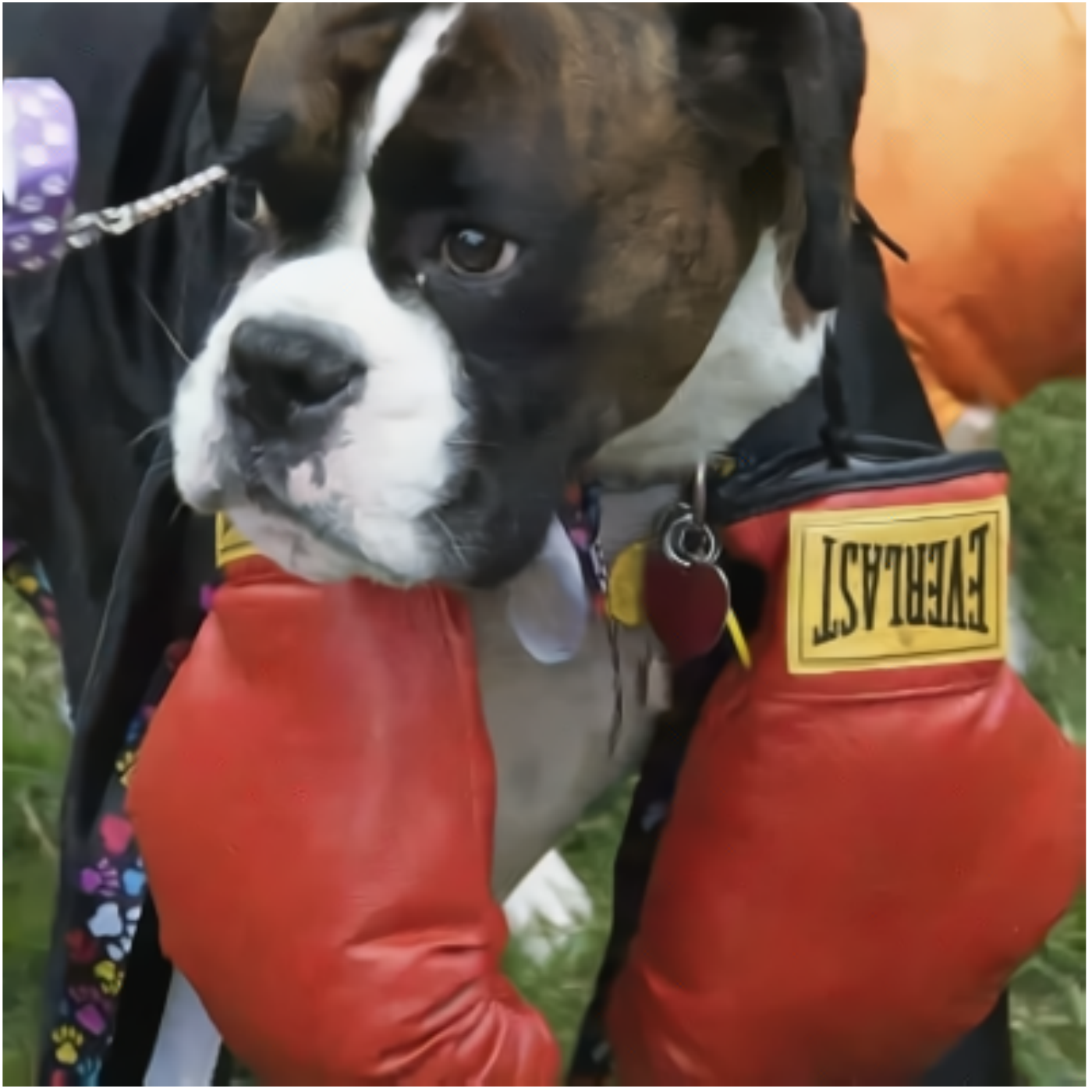}\vspace{-0.15cm}\\
   {\footnotesize (a) Noisy image ($\sigma = 22$)} & \footnotesize{(b) CDnCNN~\cite{Zhang2017} }& \footnotesize{(c) CBM3D~\cite{Dabov2007}} & \footnotesize{(d) Noise Clinic~\cite{Lebrun2015}} & \footnotesize{(e) $\operatorname{CUNLNet}_5$} \\   
\end{tabular}
\vspace{0.02cm}
   \caption{Real grayscale and color image denoising results. The indicated value of $\sigma$ corresponds to noise's standard deviation estimate, which we provide as additional input both to BM3D and to our proposed networks. The details in the results are better viewed magnified on a computer screen.}
   \label{fig:RealComp}
   \vspace{-0.25cm}
\end{figure*}

From these results we observe that the proposed networks perform better than all the methods except to the blind variant of the deep network (DnCNN)~\cite{Zhang2017}, which similar to ours can handle inputs distorted by different noise levels. Specifically, on average our local model ${\operatorname{\footnotesize UNet}_5}$ leads to results that are $0.25$ dB worse than DnCNN, while our non-local network ${\operatorname{\footnotesize UNLNet}_5}$ performs better than the local one but still falls behind DnCNN around $0.15$ dB on average. Nevertheless, the memory footprint of the proposed networks is about 14 times smaller than that of DnCNN (48K versus 666K parameters), which makes them ideal for deployment in mobile devices where memory storage is limited. More importantly, as we demonstrate later, our models show an excellent denoising performance under more realistic noise conditions, as opposed to the blind  DnCNN that performs poorly. In Table~\ref{tab:GrayComp} we further report the results obtained by our non-local network when the indices of similar patches are computed from the ground-truth images. In this case we observe that ${\operatorname{\footnotesize UNLNet}_5}^\mathrm{orc}$ outperforms DnCNN and leads to an average increase of $0.25$ dB. While this is not a practical configuration, these results highlight the fact that a better grouping approach, which is out of the scope of the current work, can lead to further improvements in the restoration quality without any need to re-train the network. Representative grayscale denoising results that demonstrate visually the restoration quality of the proposed models are shown in Fig.~\ref{fig:GrayComp}.
\vspace{-0.5cm}\paragraph{Color denosing}
Similar to the grayscale case we trained two different network configurations, one using a local operator and one using a non-local operator. The only difference between the color denoising network architecture and the grayscale one is that the convolution layers used in the color-denoising networks consist of 74 filters of support $5\times 5\times 3$. The rest of the network parameters and the training setup remains the same as above. In Table~\ref{tab:ColorComp} we report results for several noise levels and our comparisons involve only methods that are specifically designed to handle color images. We refrain from reporting results by methods that are applied on each image channel independently, since such results are not competitive enough.

From Table~\ref{tab:ColorComp} we observe that our color networks outperform CBM3D~\cite{Dabov2007}, which has been the state-of-the-art method for almost a decade, by 0.35 dB for the local model and 0.5 dB for the non-local model and they are very competitive to CDnCNN~\cite{Zhang2017}, which is the current state-of-the-art method. Specifically, the proposed non-local network on average matches the performance of CDnCNN while it is considerably more shallow with 7 times less parameters (93K versus 668K). Another important advantage of the proposed networks, as it will be demonstrated next, is that similarly to the grayscale case they perform very well when the noise distorting the input is not AWGN, as opposed to blind DnCNN that cannot handle successfully such cases. For a visual inspection of the restoration performance of the proposed color models we refer to Fig.~\ref{fig:ColorComp}.

\vspace{-.55cm}\paragraph{Results on real images}
To demonstrate the practical significance of the proposed network architecture, we further report representative results on images obtained from~\cite{Lebrun2015}, which are distorted by real noise and whose distribution and noise level are unknown. Since ground-truth images are not available, the evaluation of the different methods is only possible by visual comparisons. From Fig.~\ref{fig:RealComp} we observe that as opposed to the rest of the methods, blind DnCNN has a hard time in removing the noise. We note that the noise-specific variant of DnCNN does not have the same drawback, but in this case the advantage of using a single network for different noise levels is lost. Regarding the performance of the proposed networks, they lead to visually pleasing results with most of the noise being  removed and without introducing any spurious artifacts, as those present in the rest of the methods under comparison. More results on real images can be found in the supplementary material.

\vspace{-.2cm}\section{Conclusions and Future Work}\label{sec:Conclusions}\vspace{-0.2cm}
In this work we proposed a novel network architecture for grayscale and color image denoising. The design of the resulting image models has been inspired by local and non-local variational methods and a constrained optimization formulation of the problem, which allows us to train our networks for a wide range of noise levels using a single set of parameters. While the architecture of the proposed networks is considerably more shallow than current state-of-the-art deep CNN-based approaches, the resulting models lead to very competitive results for AWGN distortions while they also appear to be very robust when the noise degrading the input deviates from the Gaussian assumption.

Based on the reported results using oracle grouping, a promising future research direction that has the potential to lead to further improvements in the restoration quality is to investigate different block-matching approaches for finding the similar patches used in the non-local variant of the proposed network. Another direction that we plan to explore is the use of the proposed networks as sub-solvers in restoration methods that deal with more general inverse imaging problems such as deblurring, inpainting, demosaicking, \etc.

\section{Appendix: Additional results on real images}
\begin{figure*}[!t]
\centering
\begin{tabular}{@{} c @{ } c @{ } c @{ } }
\includegraphics[width=.33\linewidth]{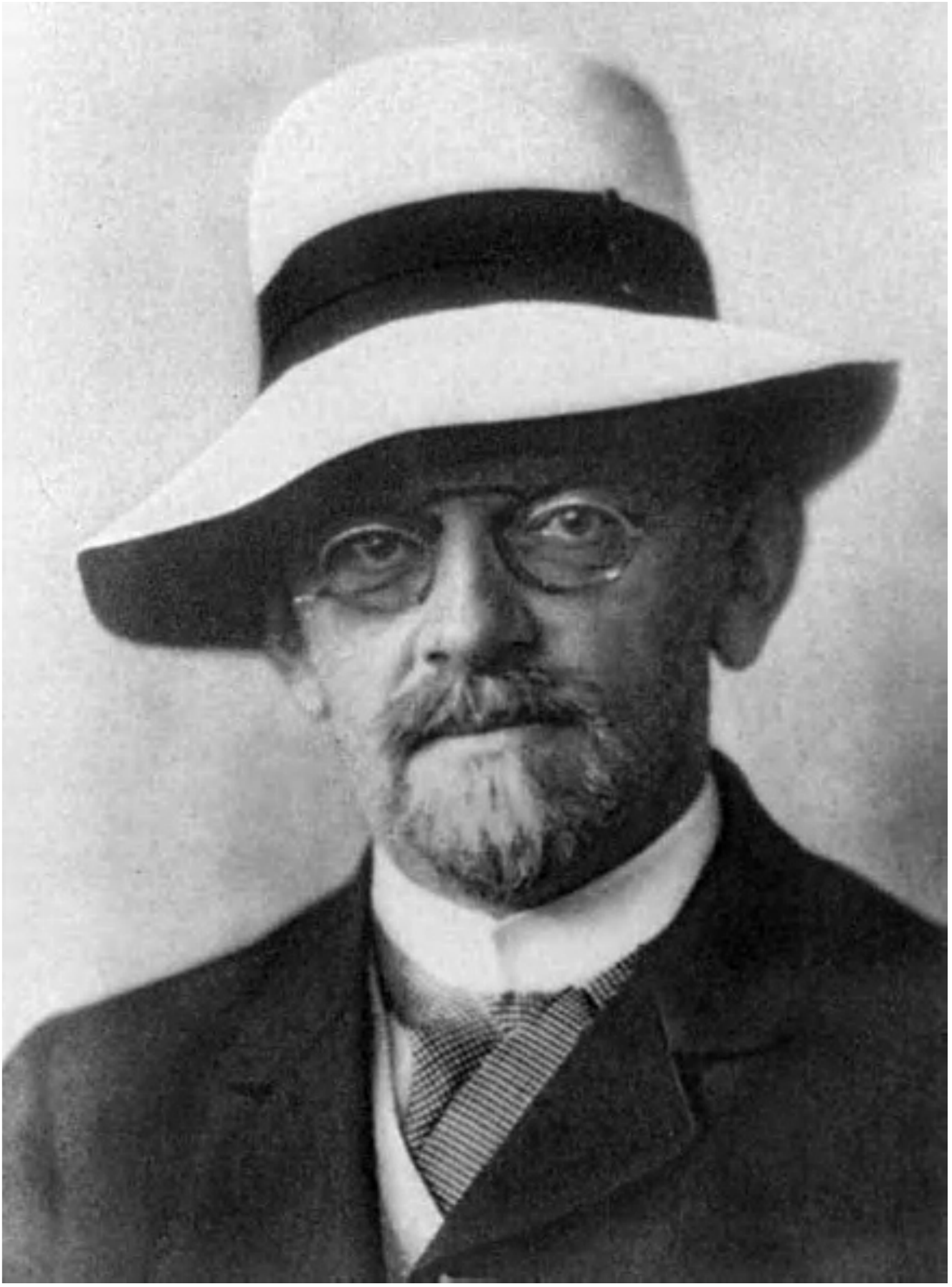}\label{fig:subfig}&
\includegraphics[width=.33\linewidth]{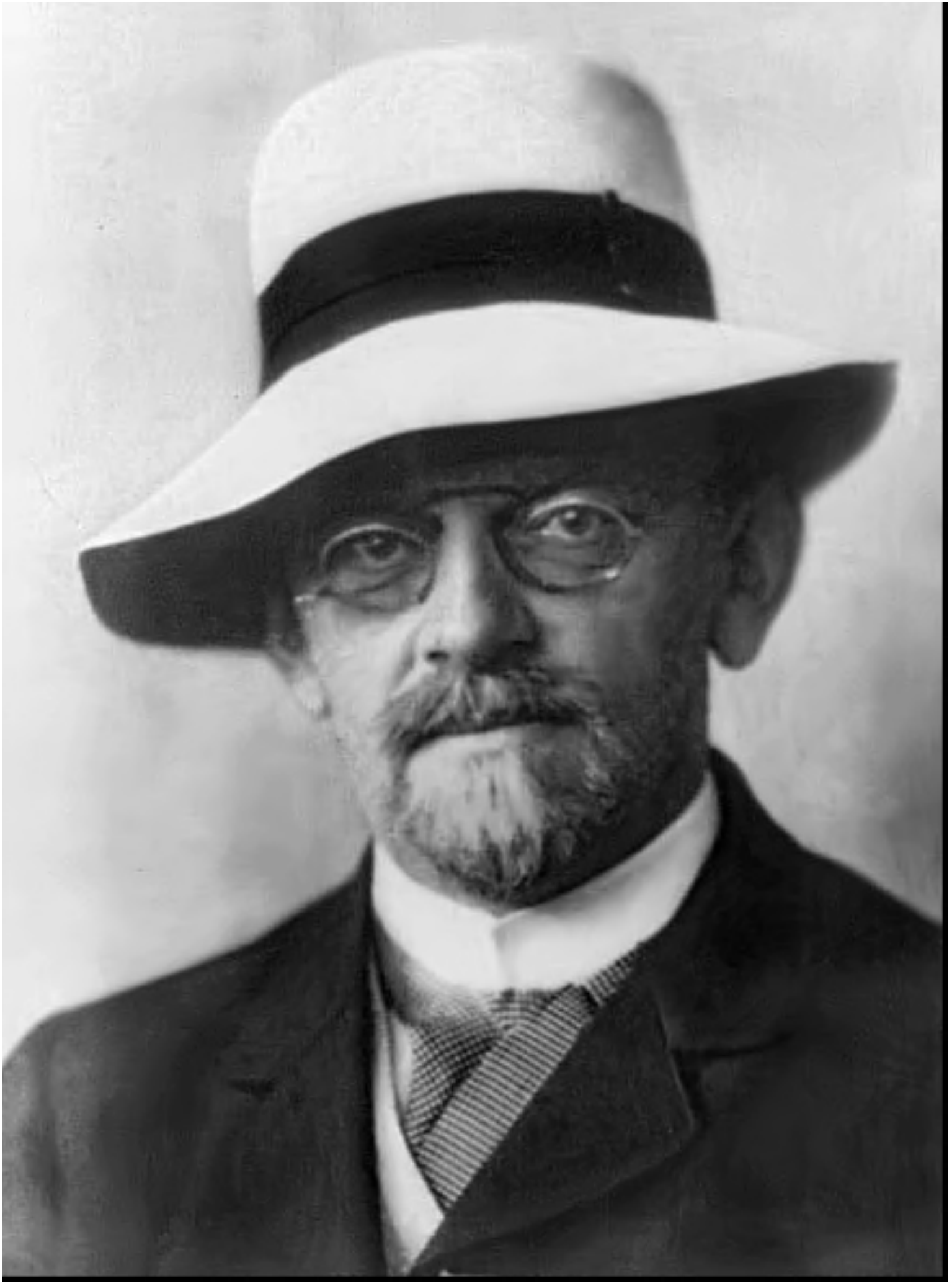}&
\includegraphics[width=.33\linewidth]{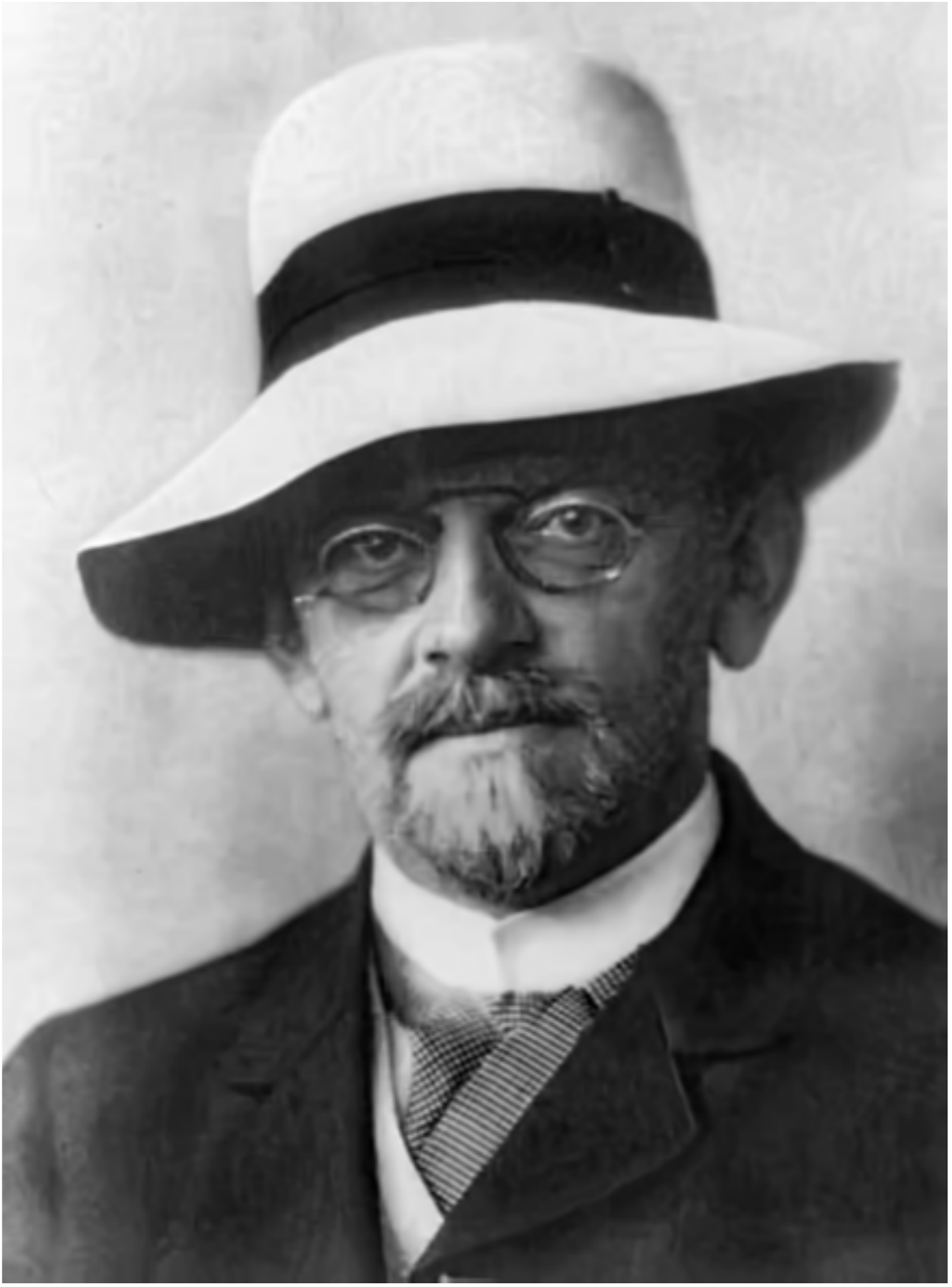}\\
(a) Noisy image ($\sigma = 8$) & (b) Noise Clinic~\cite{Lebrun2015} & (c) BM3D~\cite{Dabov2007}\\
\includegraphics[width=.33\linewidth]{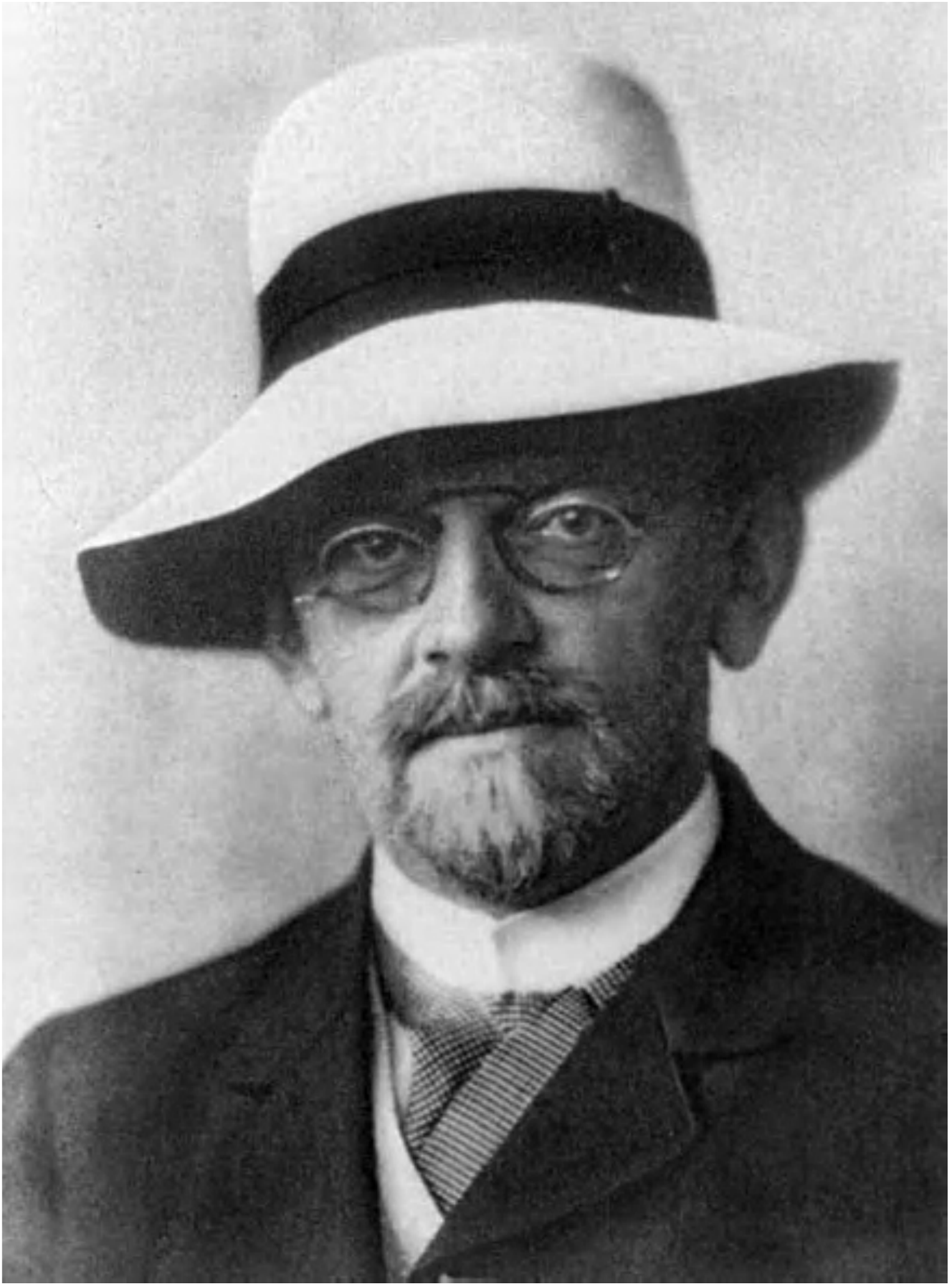}&
\includegraphics[width=.33\linewidth]{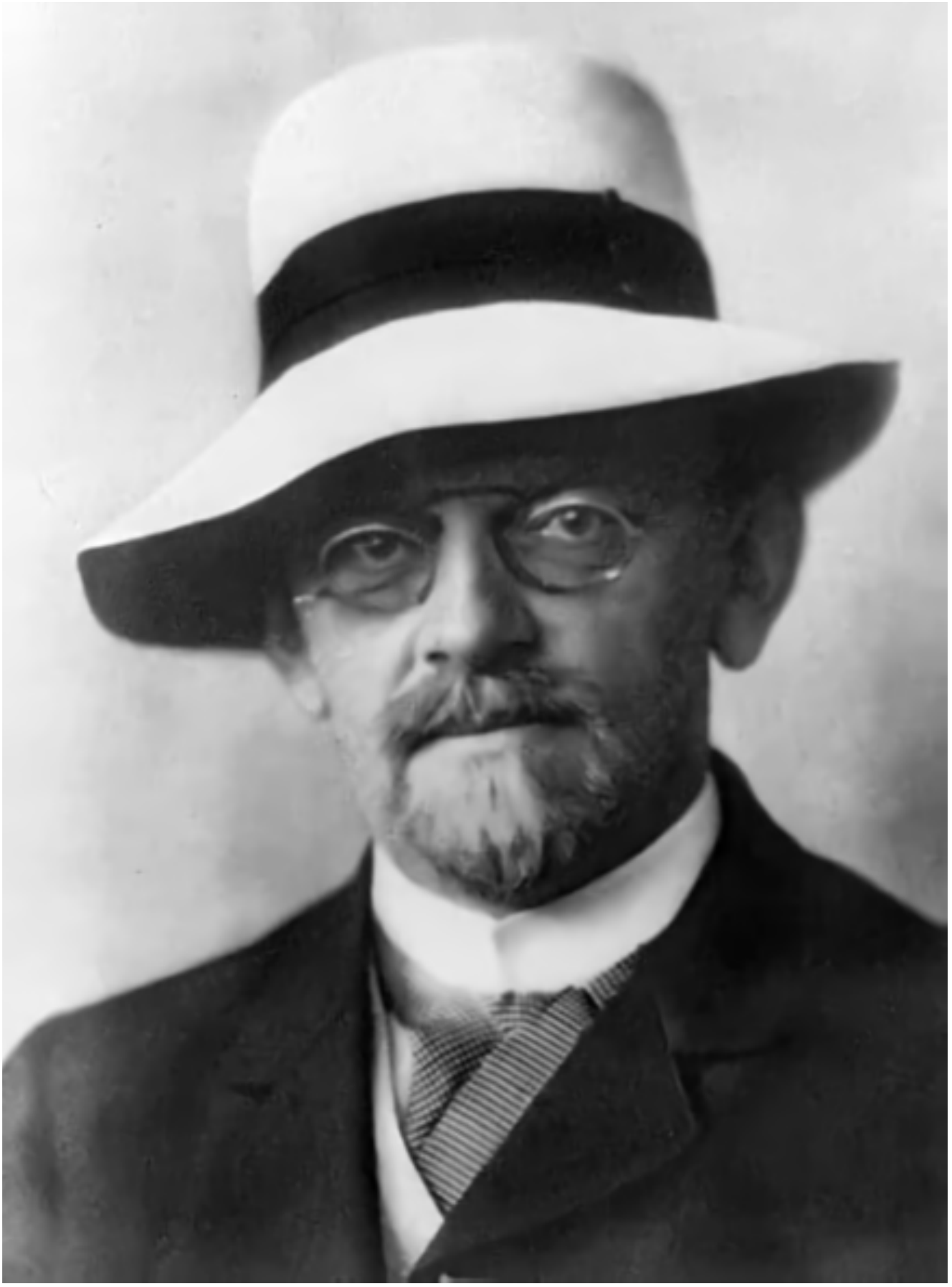}&
\includegraphics[width=.33\linewidth]{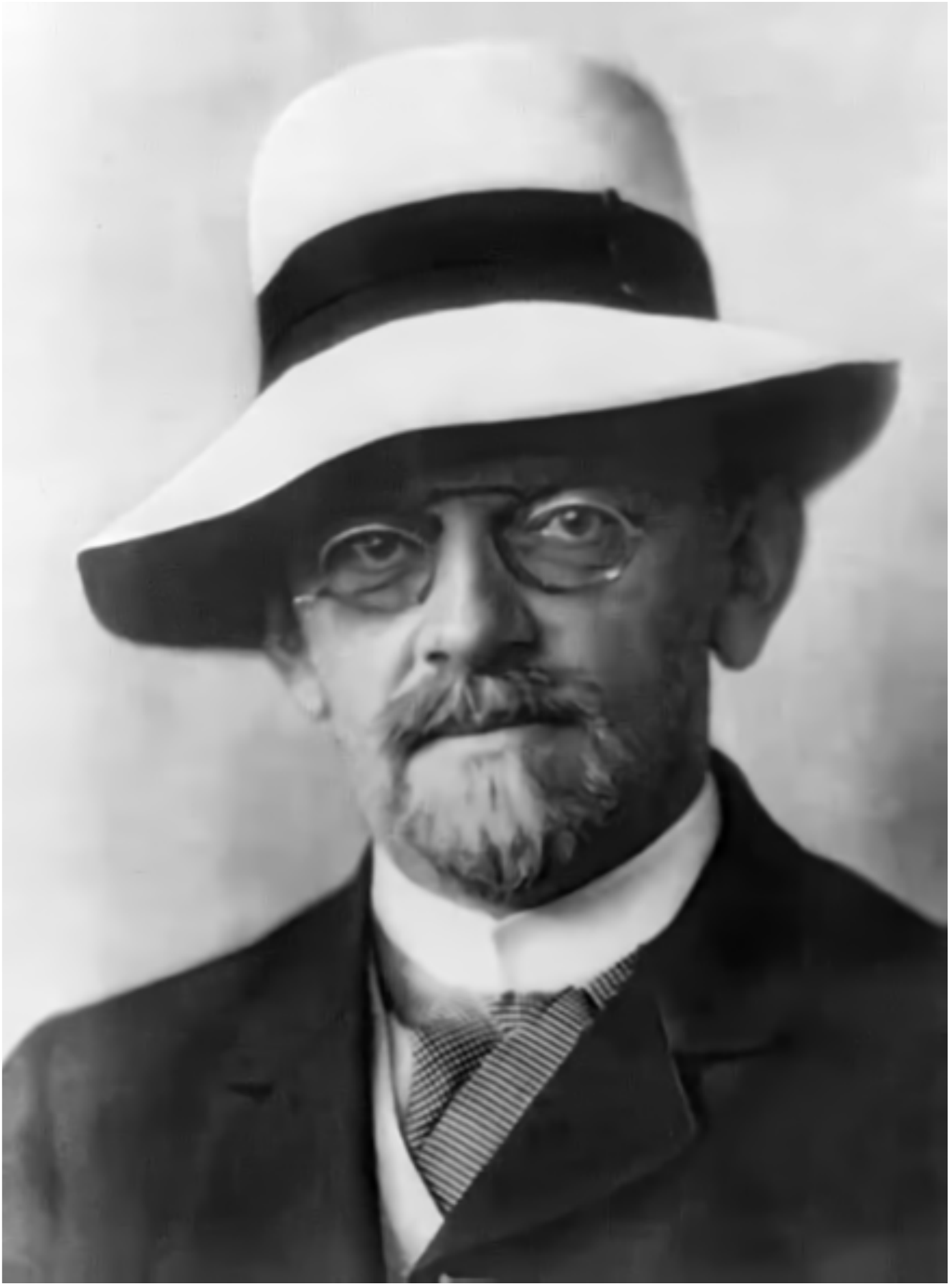}\\
(d) DnCNN~\cite{Zhang2017} & (e) $\operatorname{UDNet}_5$ & (f) $\operatorname{UNLDNet}_5$\\
\end{tabular}
    \caption{Real grayscale image denoising. \textbf{Images are best viewed magnified on a computer screen.}}
   \label{fig:RealGrayComp1}
\end{figure*}

\begin{figure*}[!t]
\centering
\begin{tabular}{@{} c @{ } c @{ } c @{ } }
\includegraphics[width=.33\linewidth]{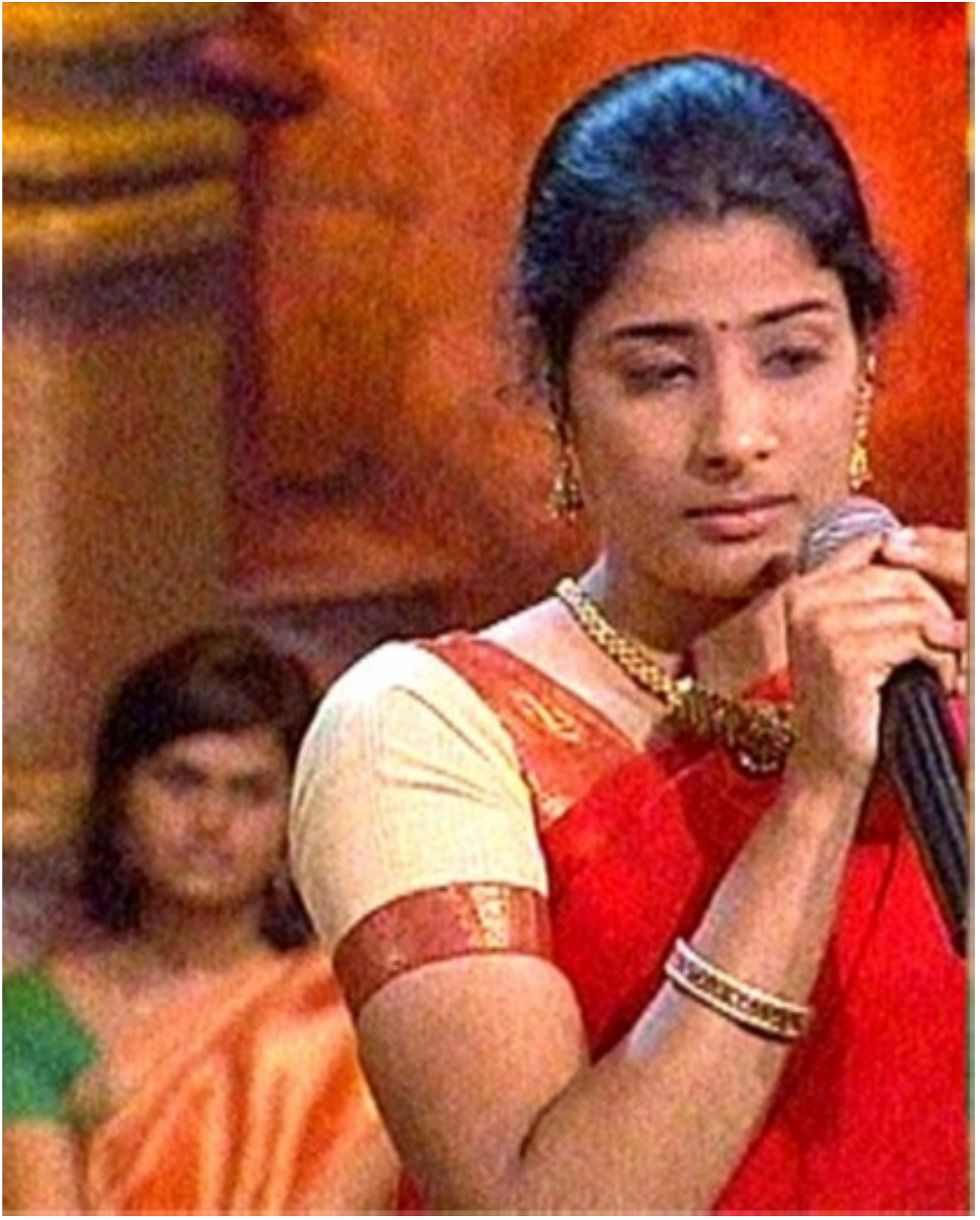}&
\includegraphics[width=.33\linewidth]{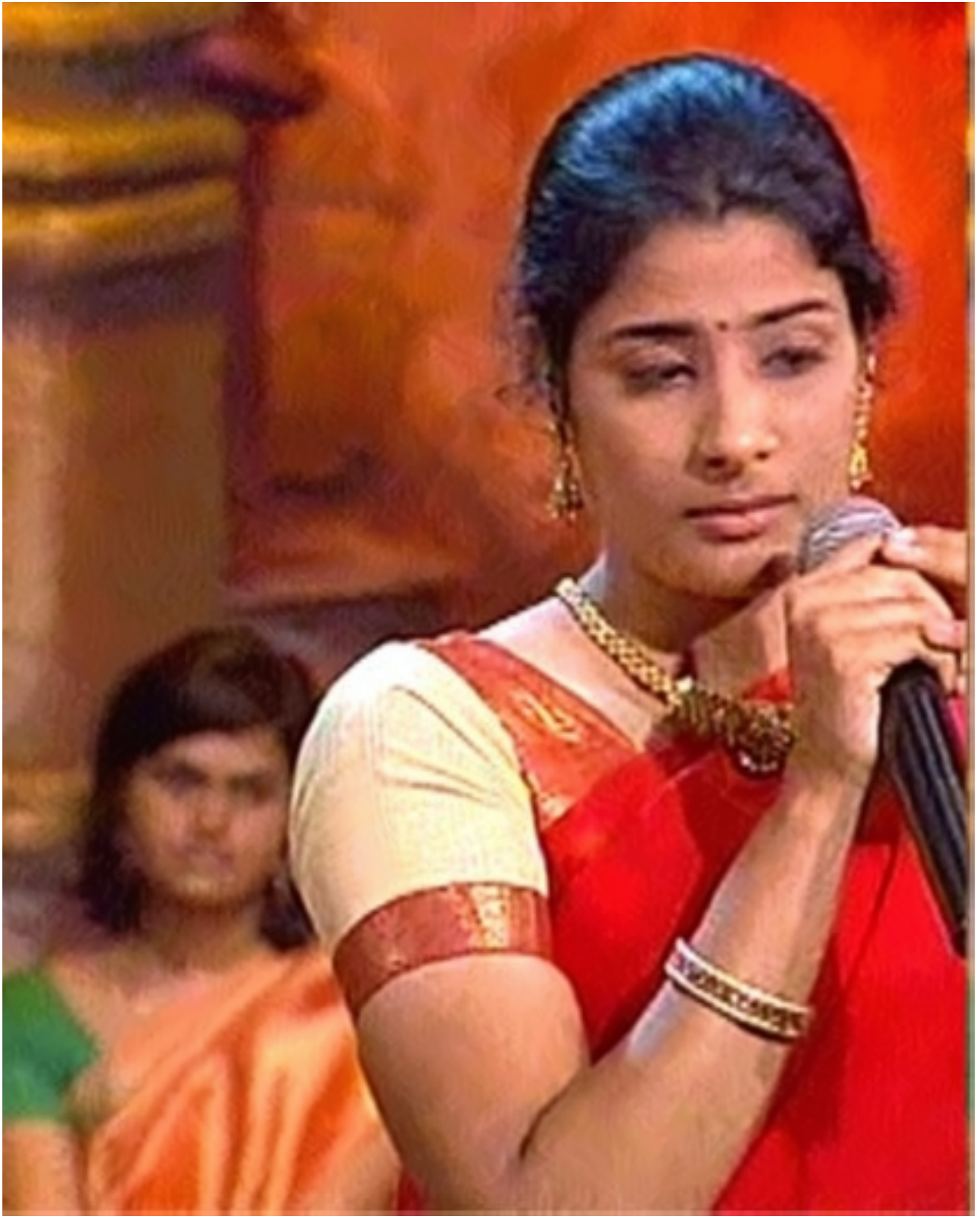}&
\includegraphics[width=.33\linewidth]{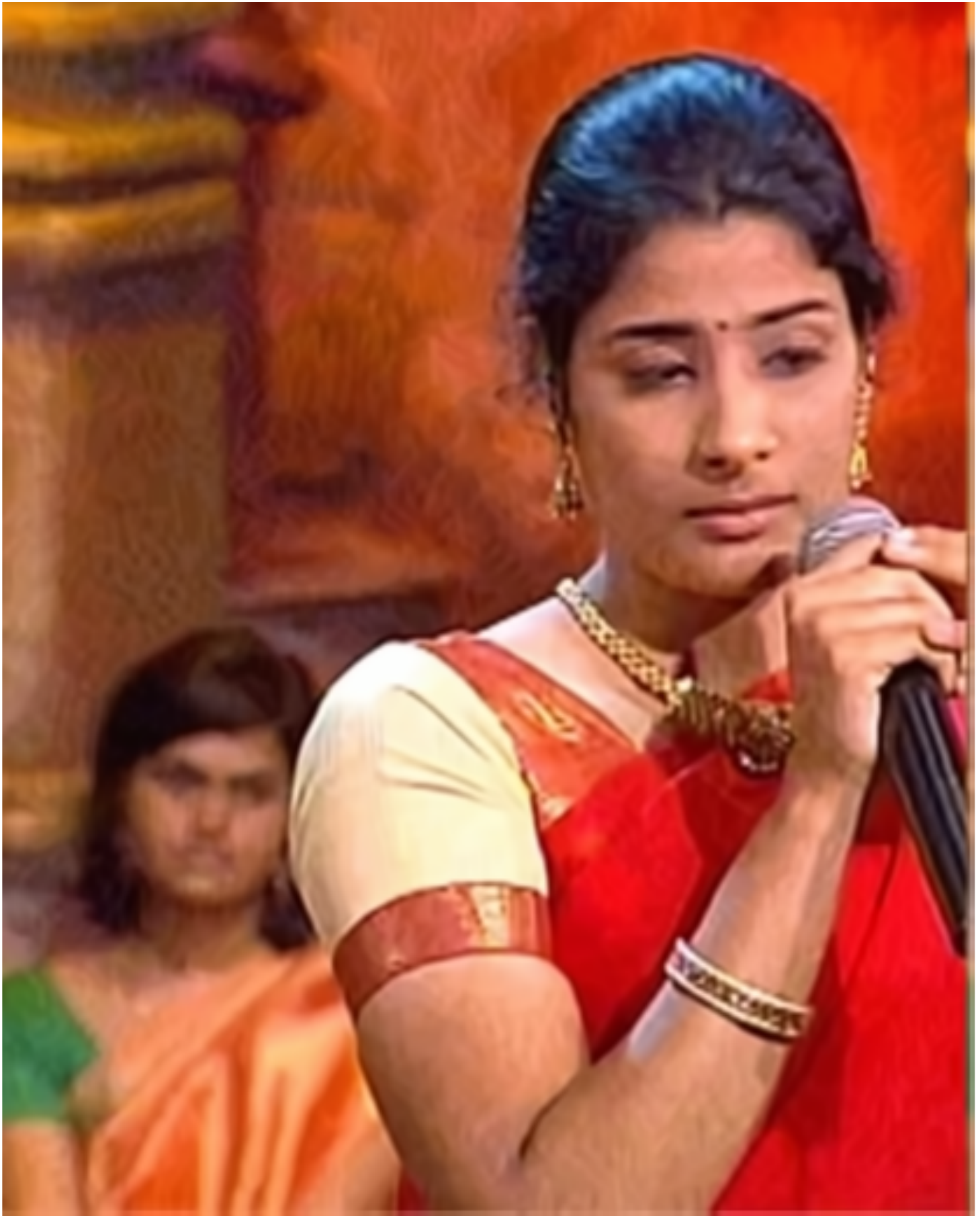}\\
(a) Noisy image ($\sigma = 25$) & (b) Noise Clinic~\cite{Lebrun2015} & (c) CBM3D~\cite{Dabov2007}\\
\includegraphics[width=.33\linewidth]{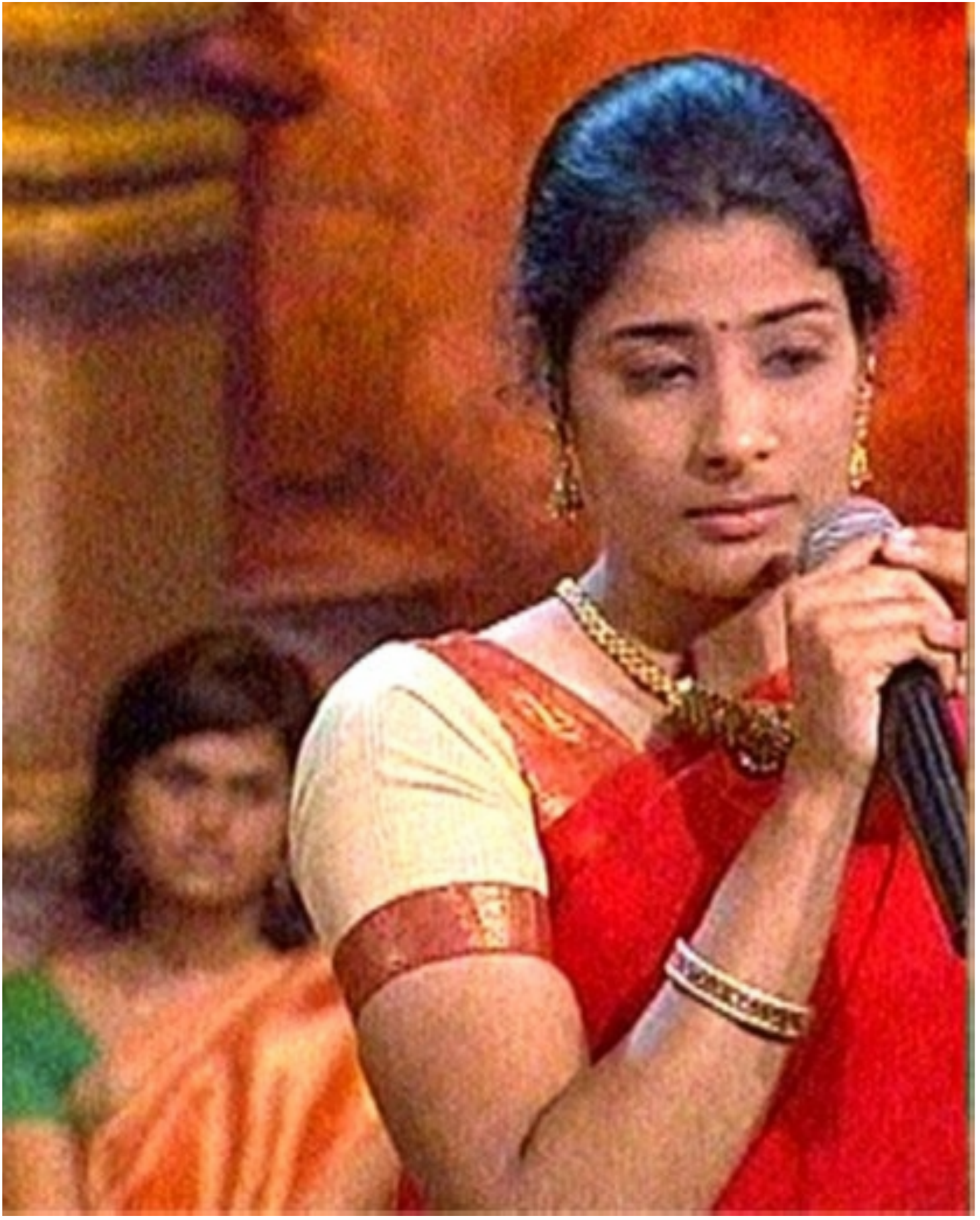}&
\includegraphics[width=.33\linewidth]{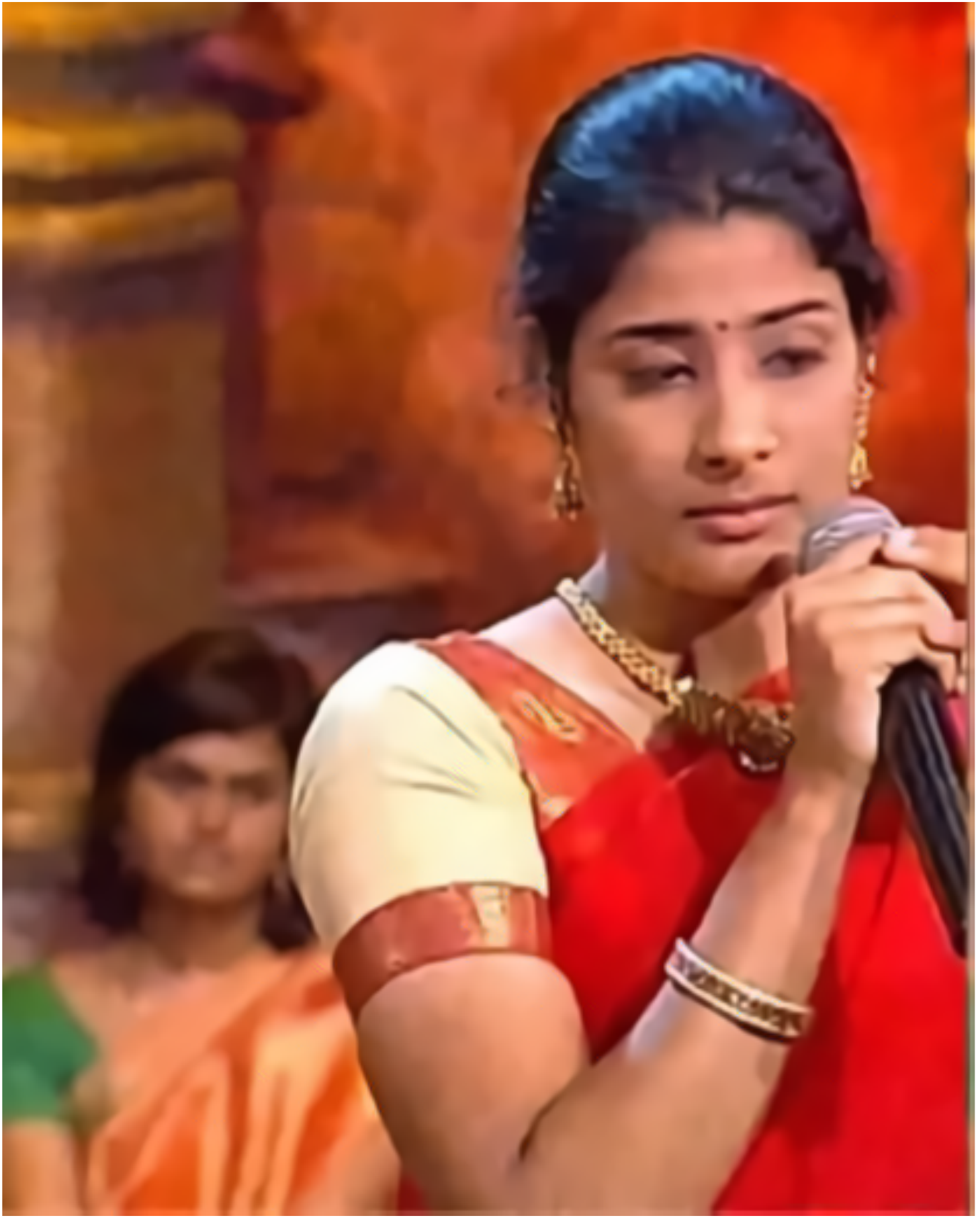}&
\includegraphics[width=.33\linewidth]{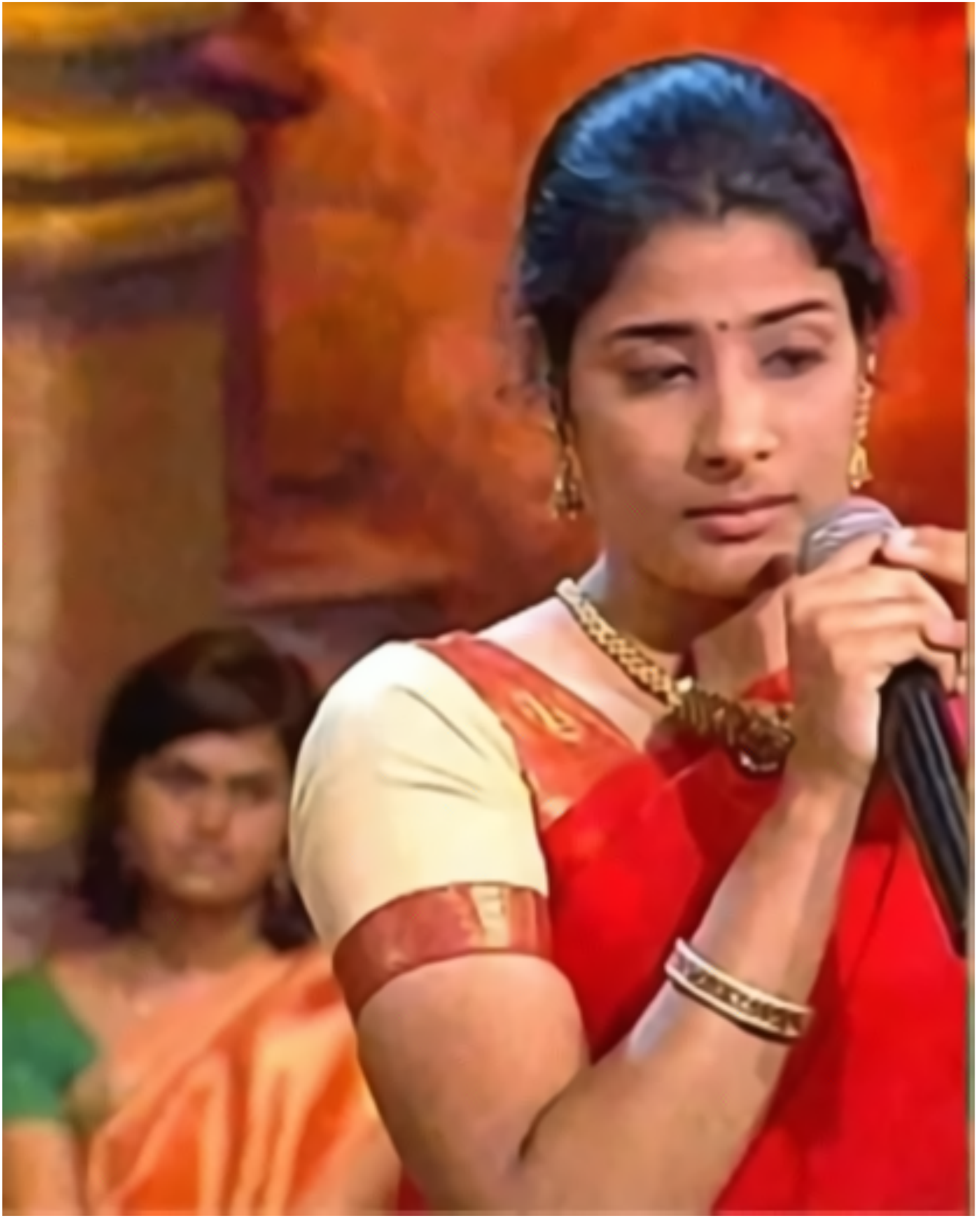}\\
(d) CDnCNN~\cite{Zhang2017} & (e) $\operatorname{CUDNet}_5$ & (f) $\operatorname{CUNLDNet}_5$\\
\end{tabular}
   \caption{Real color image denoising. \textbf{Images are best viewed magnified on a computer screen.}}
   \label{fig:RealColorComp1}
\end{figure*}

In Figs.~\ref{fig:RealGrayComp1}-\ref{fig:RealColorComp3} we provide additional grayscale and color image denoising results on images that have been distorted by real noise,  whose level and distribution are unknown. Further, these images are quantized and their values are in the range [0, 255]. All the images are publicly available and were obtained from~\cite{Lebrun2015}, except to  Fig.~\ref{fig:RealGrayComp1} which is available from \url{https://en.wikipedia.org/wiki/David_Hilbert}. In our reported results we compare 5 different methods that are all applicable both to grayscale and color images. In particular, we consider the method proposed in~\cite{Lebrun2015}, which the authors refer to as ``noise clinic" and it was developed so that it can be adapted to any signal dependent colored noise,  the BM3D algorithm~\cite{Dabov2007}, which has been the state-of-the-art Gaussian denoising method for almost a decade and still leads to very competitive results, DnCNN~\cite{Zhang2017}, which is a deep learning method that achieves the current state-of-the-art performance in Gaussian denoising, and the two variants (local and non-local) of our proposed denoising network. Since ground-truth images do not exist, we cannot provide any quantitative comparisons and the evaluation of the different methods is only possible by a visual comparison of their restoration results. It is also worth mentioning that all the methods under comparison but the ``noise clinic" have been originally designed to deal with additive white Gaussian noise (AWGN). Therefore, the main goal of our comparisons is to assess how robust each method is when the noise deviates significantly from the assumed noise model. Finally, we note that the noise clinic method and the blind variant of the DnCNN network are equipped with an internal mechanism to estimate the noise level. On the other hand, the BM3D algorithm, the noise-specific variant of DnCNN (DnCNN-S) and our proposed networks apart from the noisy input, they accept a second input argument which corresponds to the standard deviation of the noise, $\sigma$. For these four methods, in our comparisons we have chosen empirically the value of $\sigma$ (we indicate this value in the caption of each image) that led to the best restoration results.\vspace{-0.3cm}

\begin{figure*}[!t]
\centering
\begin{tabular}{@{} c @{ } c @{ } c @{ } }
\includegraphics[width=.33\linewidth]{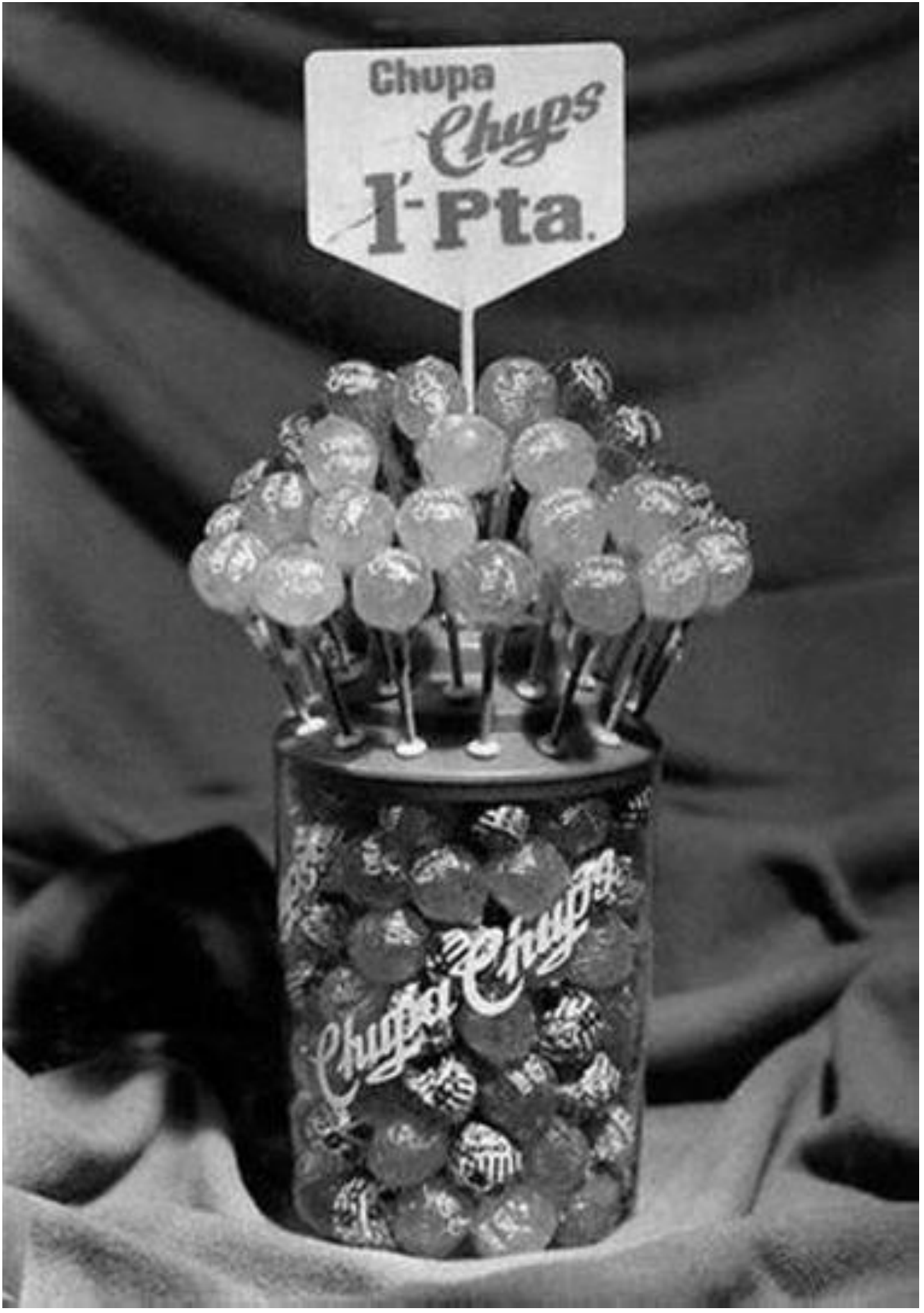}&
\includegraphics[width=.33\linewidth]{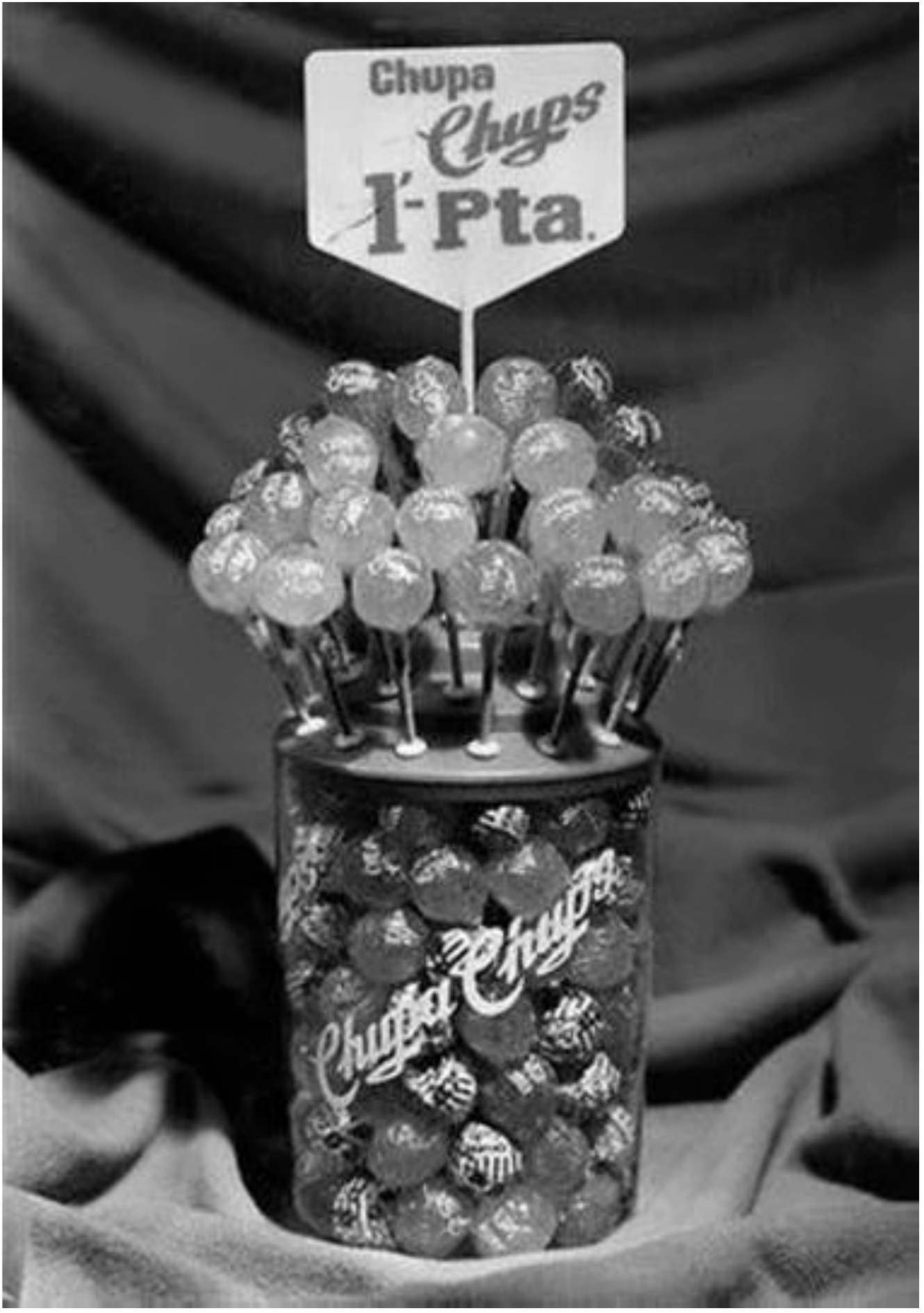}&
\includegraphics[width=.33\linewidth]{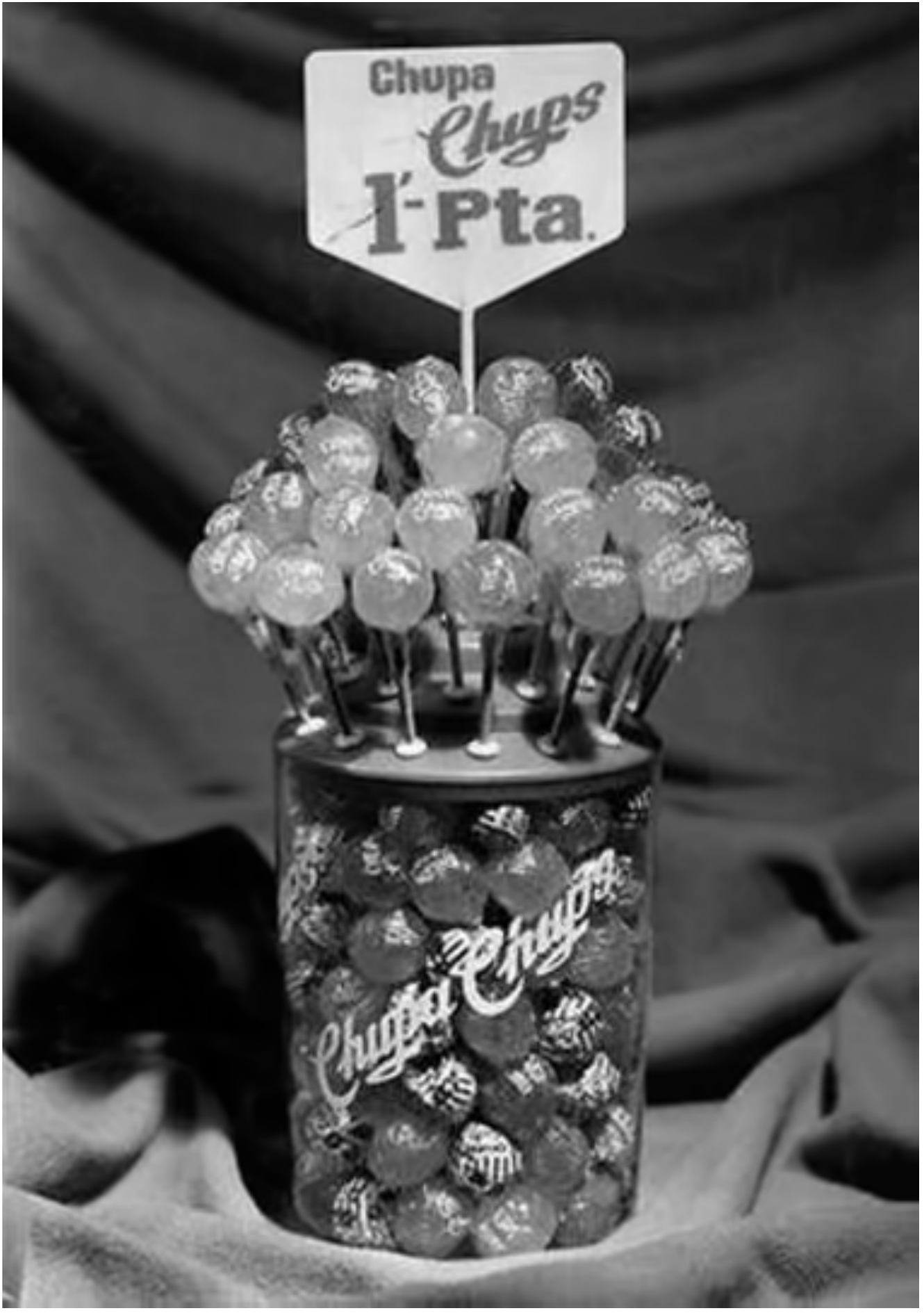}\\
(a) Noisy image ($\sigma = 5$) & (b) Noise Clinic~\cite{Lebrun2015} & (c) BM3D~\cite{Dabov2007}\\
\includegraphics[width=.33\linewidth]{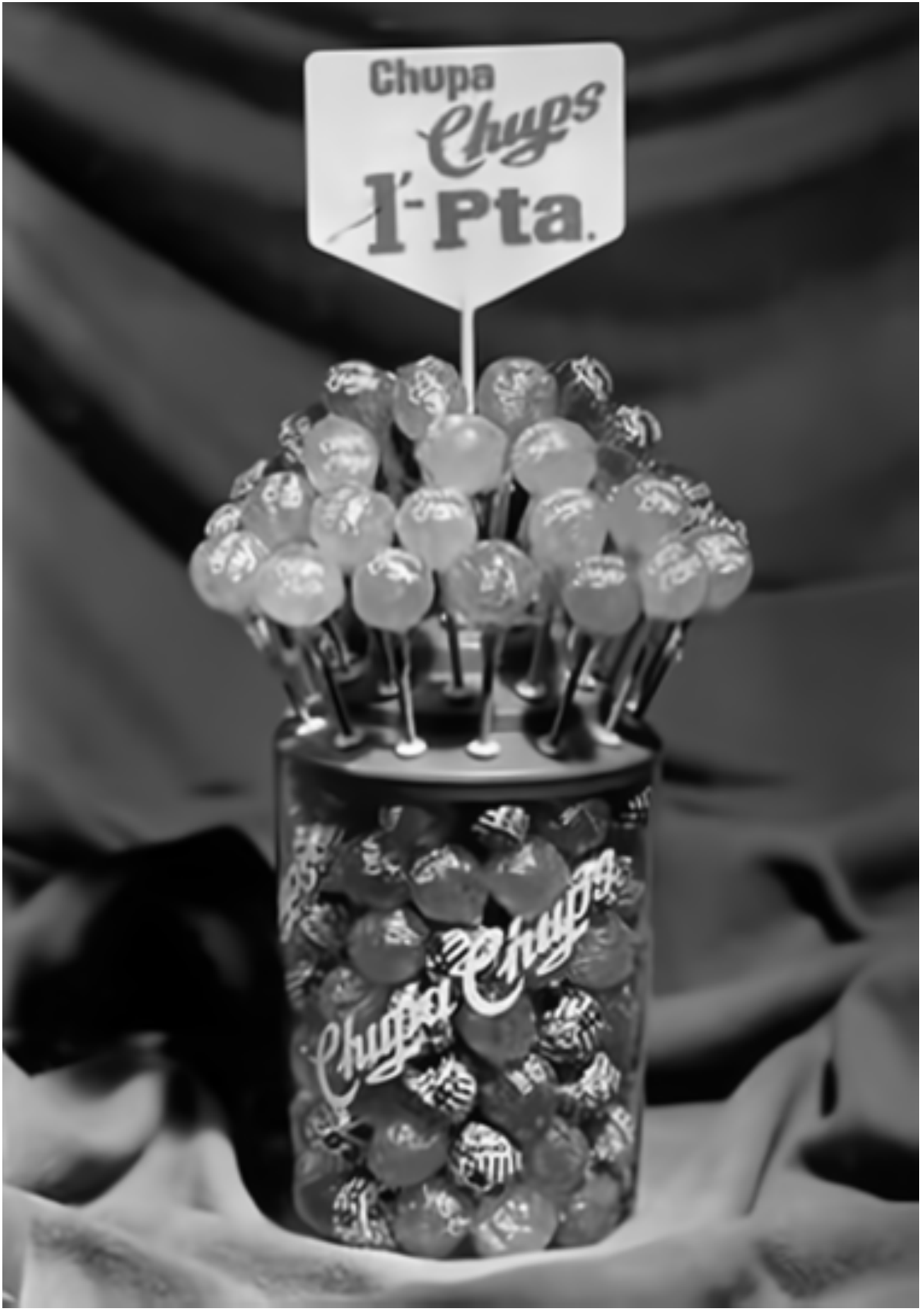}&
\includegraphics[width=.33\linewidth]{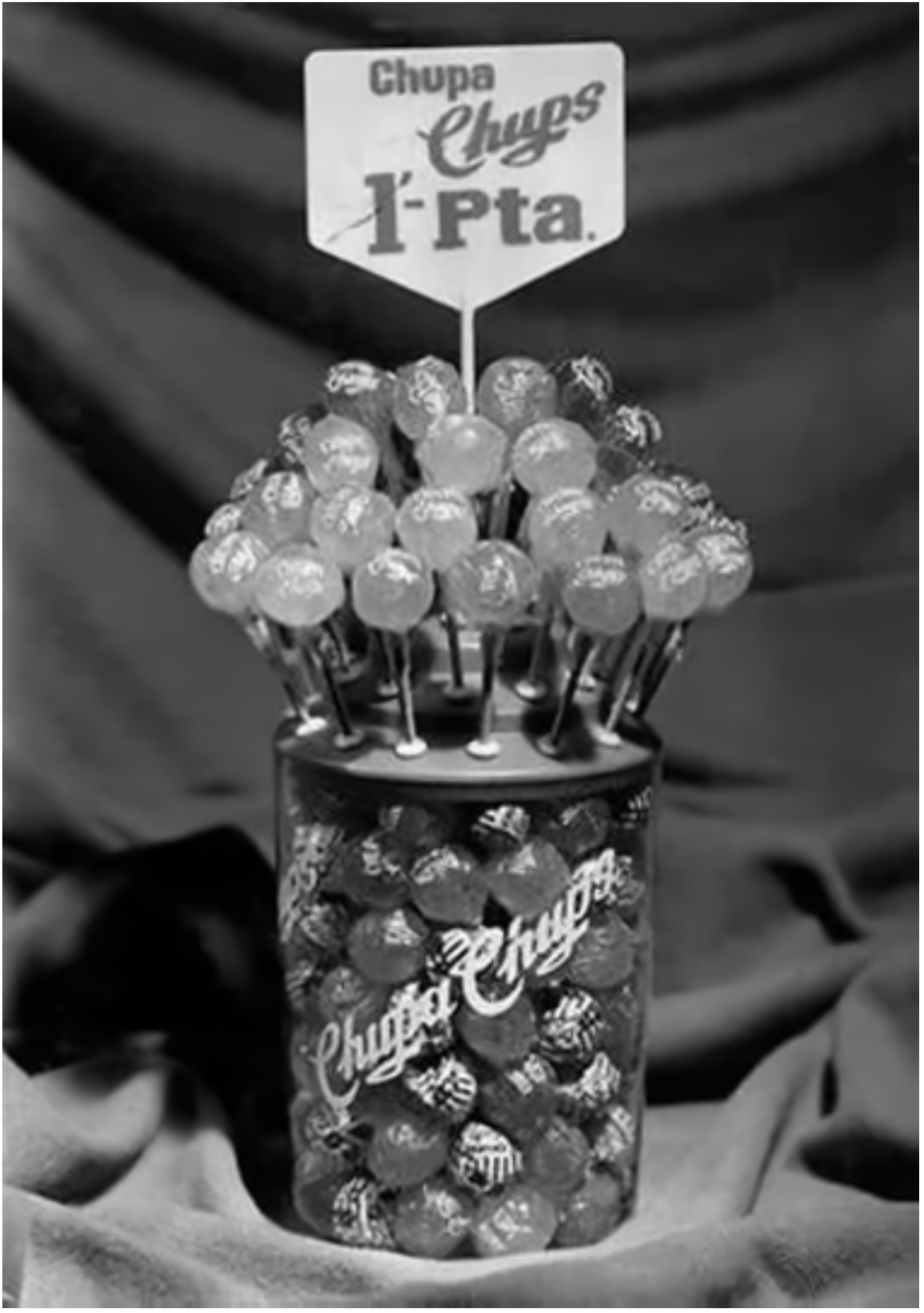}&
\includegraphics[width=.33\linewidth]{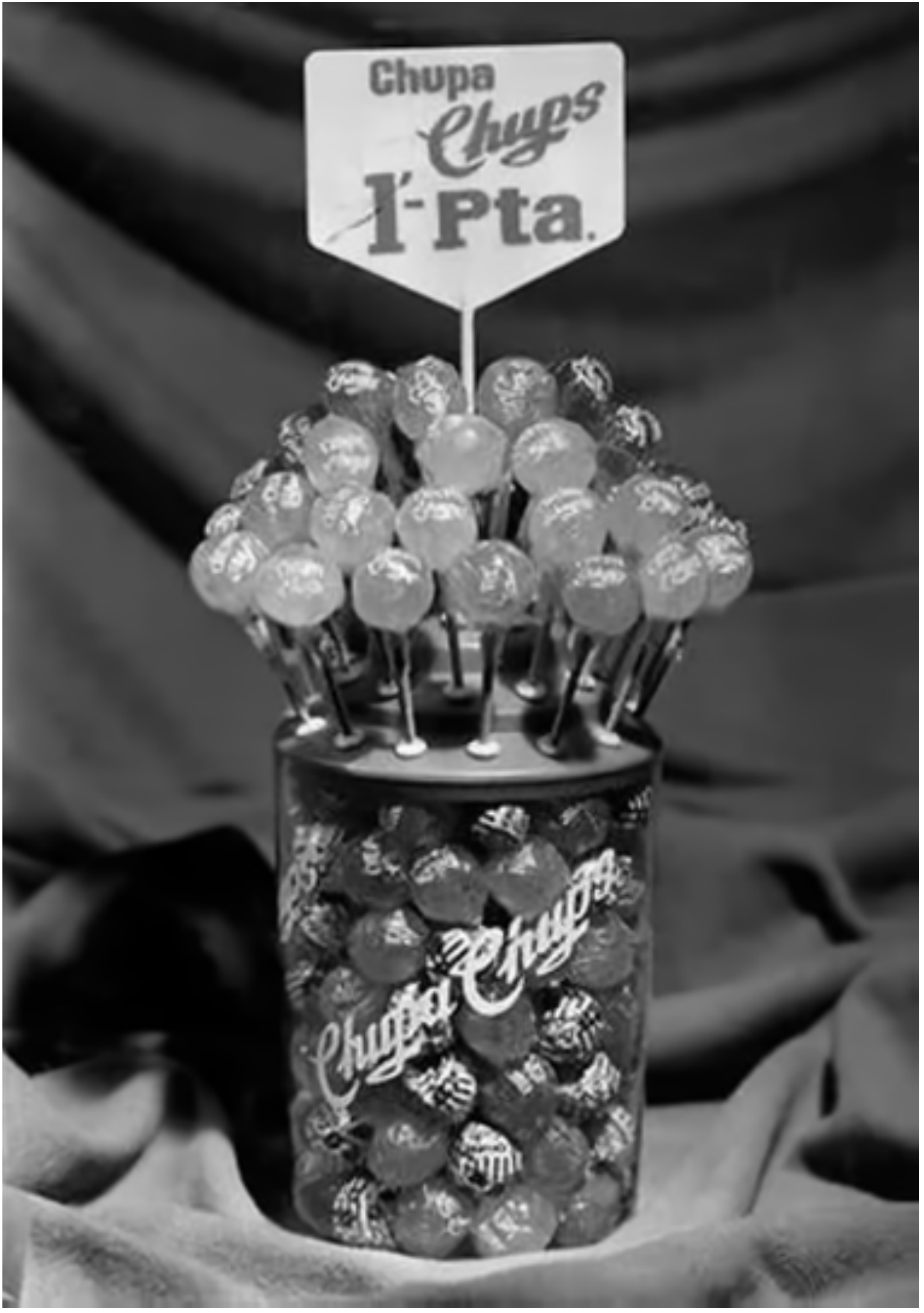}\\
(d) DnCNN-S~\cite{Zhang2017} & (e) $\operatorname{UDNet}_5$ & (f) $\operatorname{UNLDNet}_5$\\
\end{tabular}
   \caption{Real grayscale image denoising. \textbf{Images are best viewed magnified on a computer screen.}}
   \label{fig:RealColorComp2}
\end{figure*}

\begin{figure*}[!t]
\centering
\begin{tabular}{@{} c @{ } c @{ } c @{ } }
\includegraphics[width=.33\linewidth]{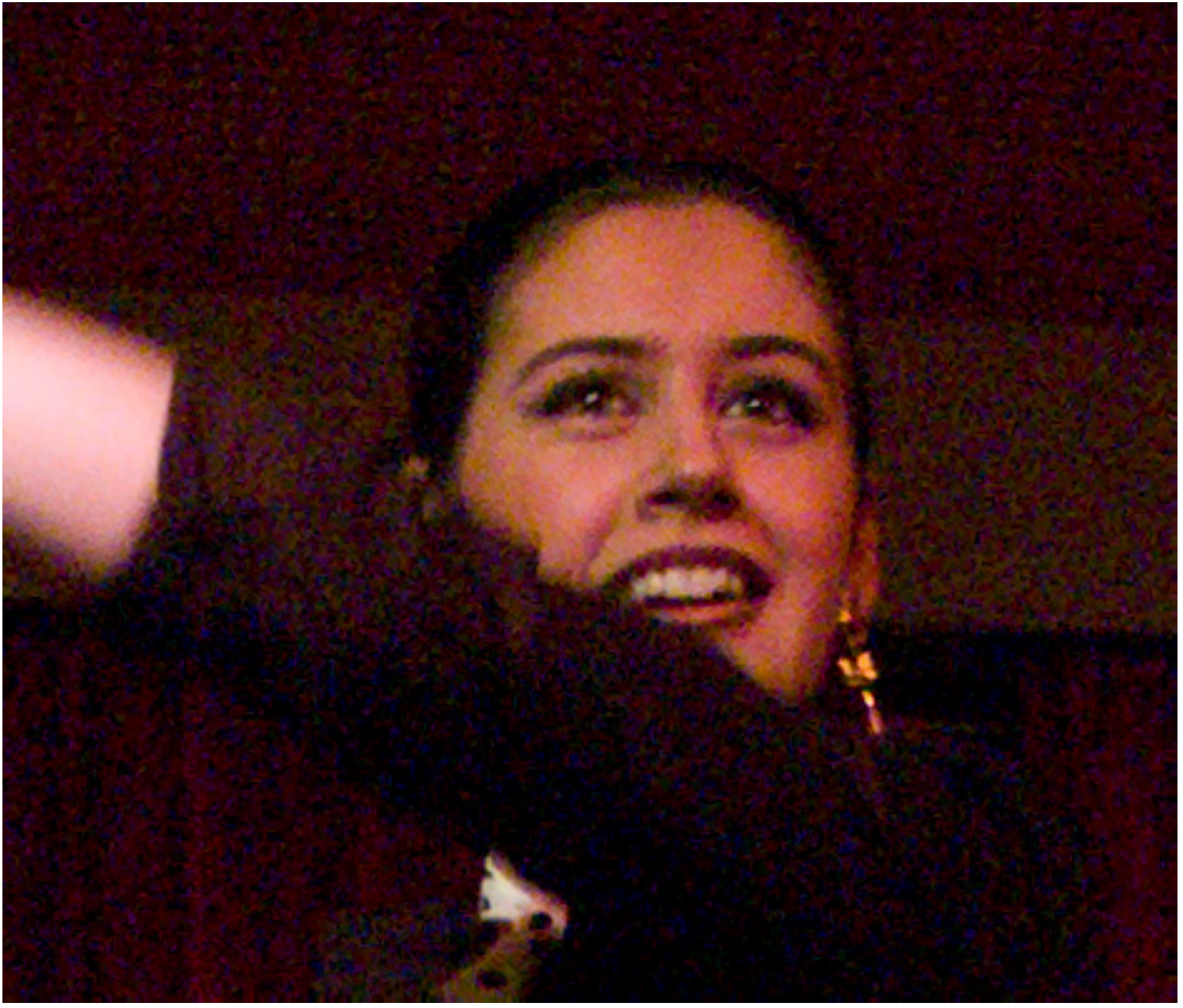}&
\includegraphics[width=.33\linewidth]{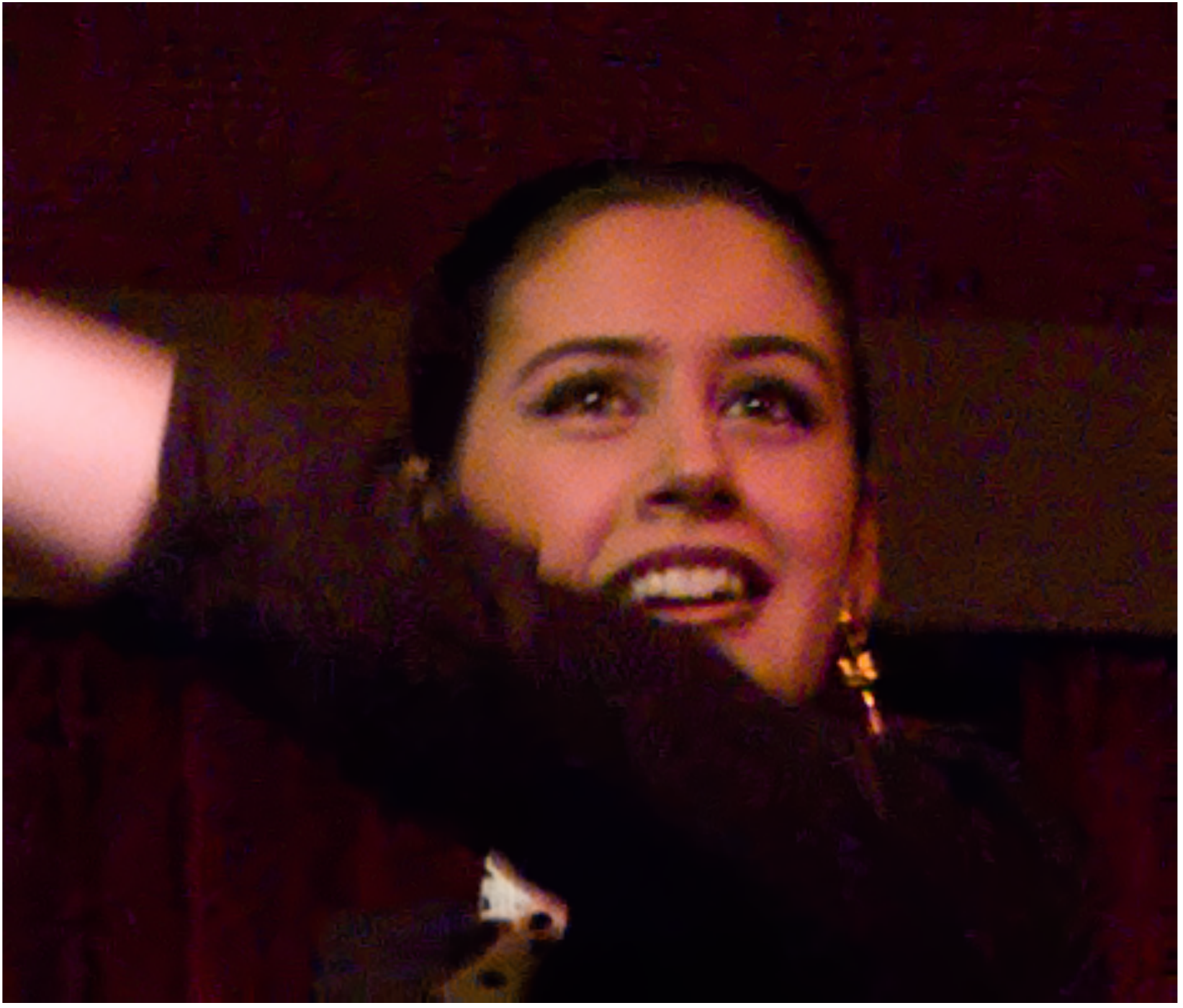}&
\includegraphics[width=.33\linewidth]{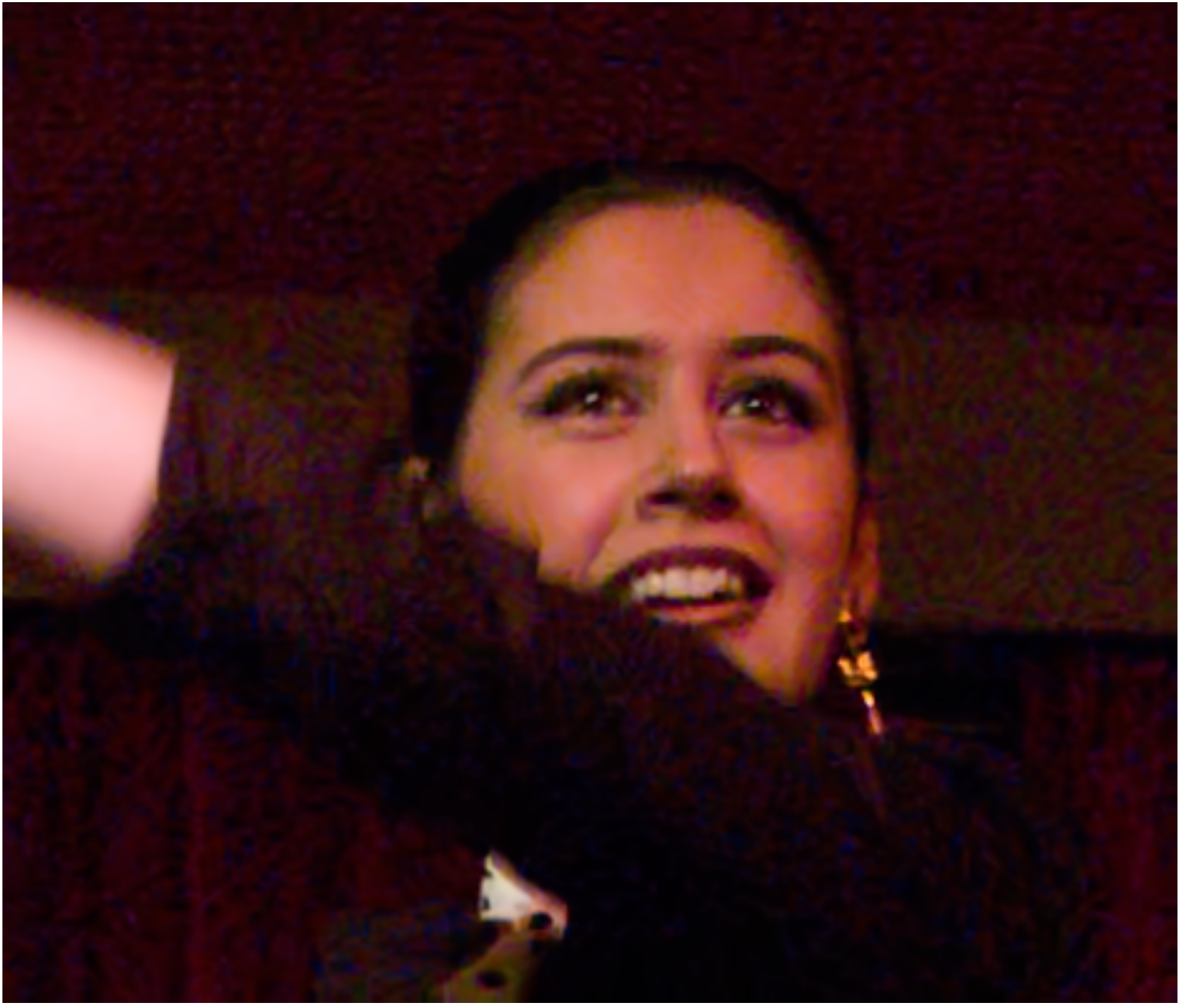}\\
(a) Noisy image ($\sigma = 15$) & (b) Noise Clinic~\cite{Lebrun2015} & (c) CBM3D~\cite{Dabov2007}\\
\includegraphics[width=.33\linewidth]{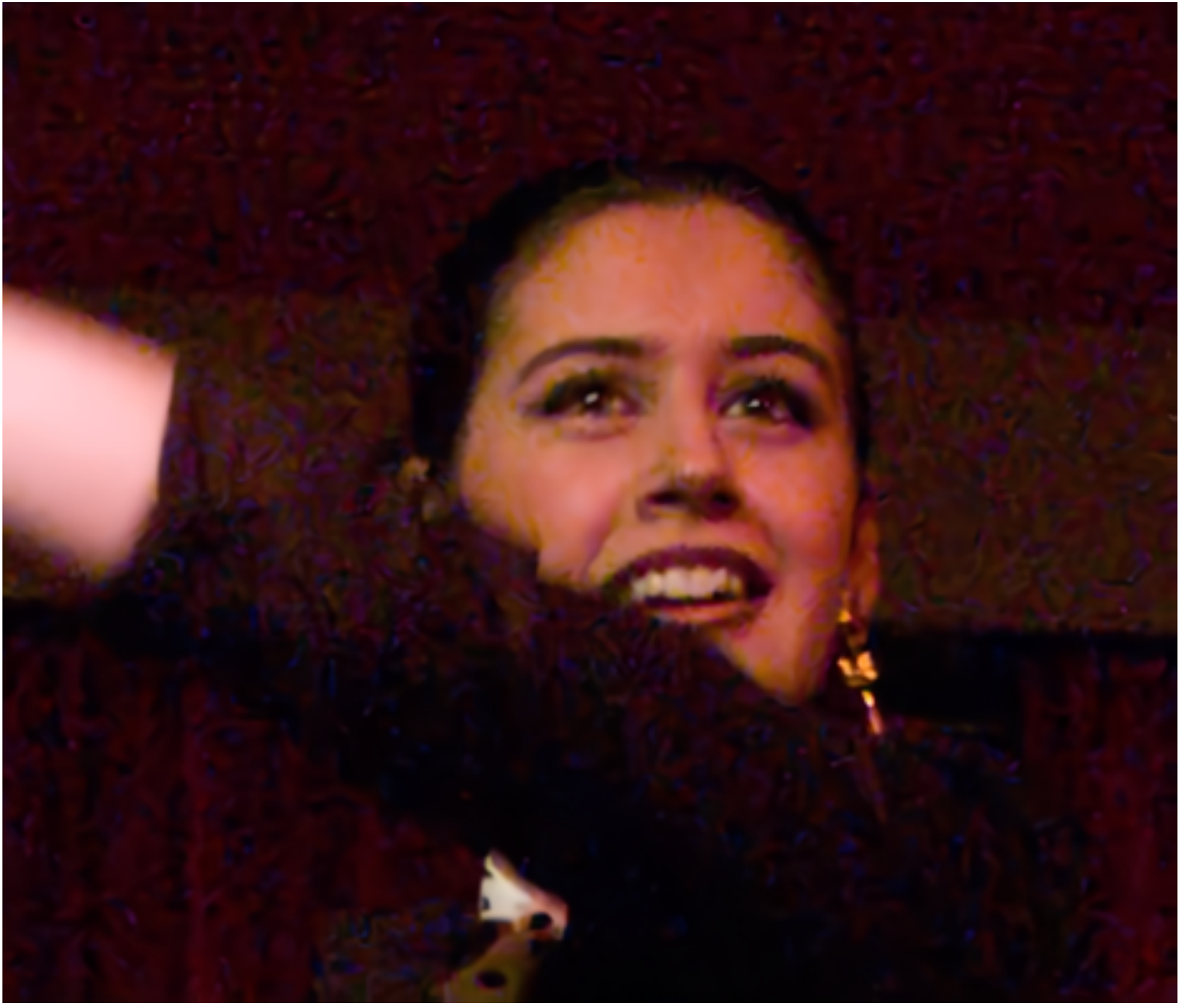}&
\includegraphics[width=.33\linewidth]{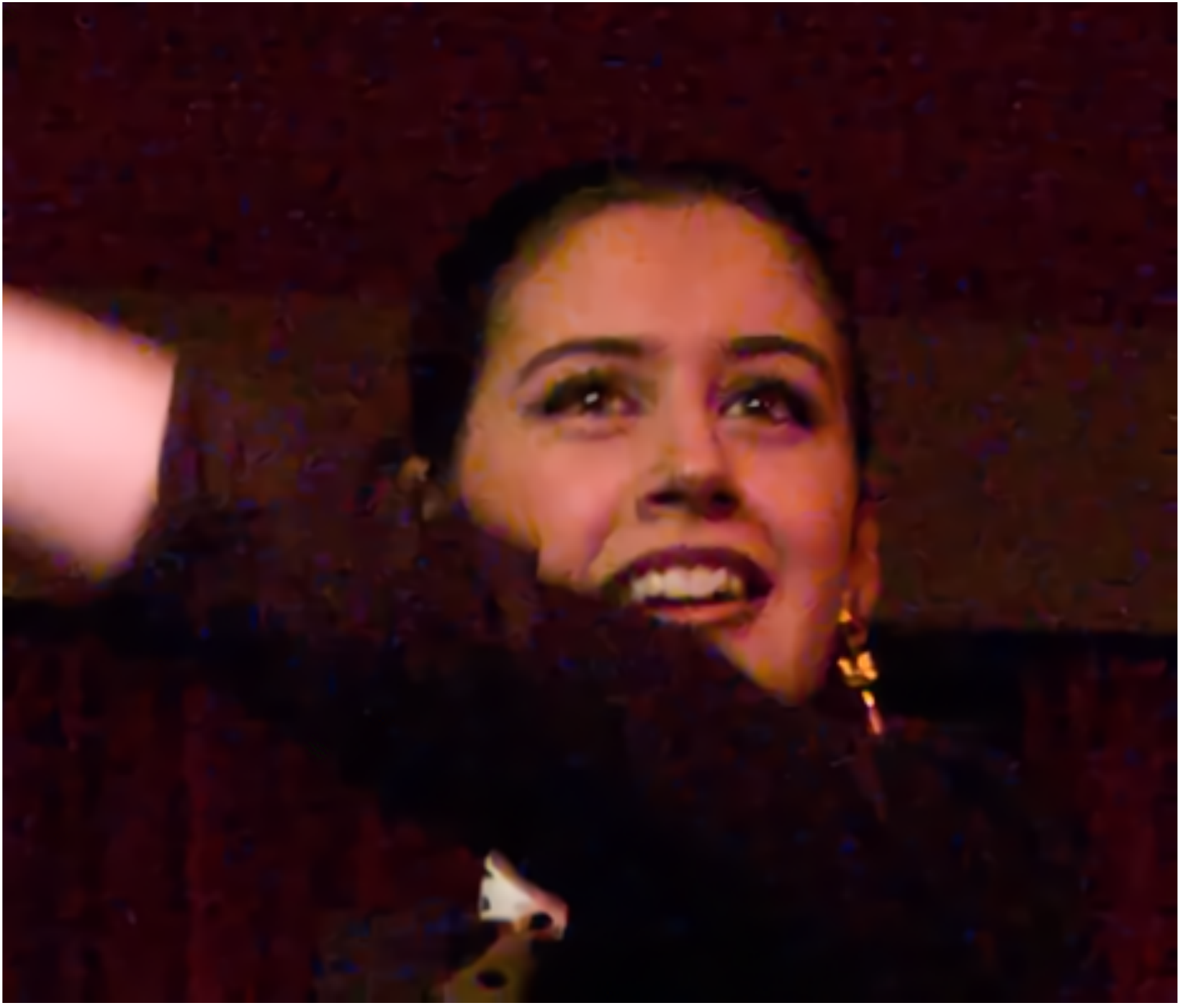}&
\includegraphics[width=.33\linewidth]{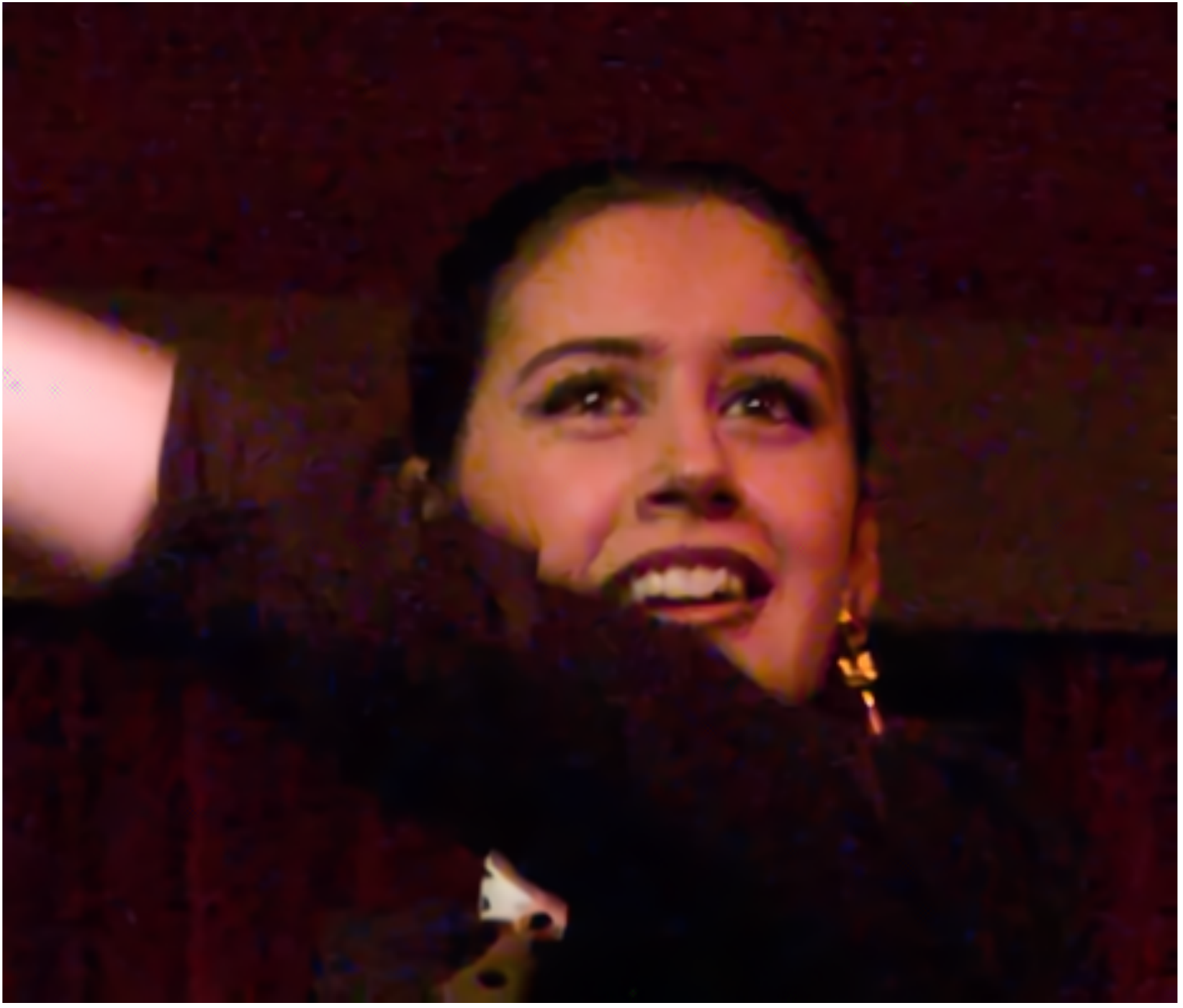}\\
(d) CDnCNN-S~\cite{Zhang2017} & (e) $\operatorname{CUDNet}_5$ & (f) $\operatorname{CUNLDNet}_5$\\
\end{tabular}
   \caption{Real color image denoising. \textbf{Images are best viewed magnified on a computer screen.}}
   \label{fig:RealColorComp2}
\end{figure*}

\newcommand{\cfbox}[1]{%
    \colorlet{currentcolor}{.}%
    {\color{#1}%
    \fbox{\color{currentcolor}}}%
}

\begin{figure*}[!h]
\centering
\begin{tabular}{@{} c @{ } c @{ } c @{ } }
 \begin{overpic}[width=.33\linewidth]{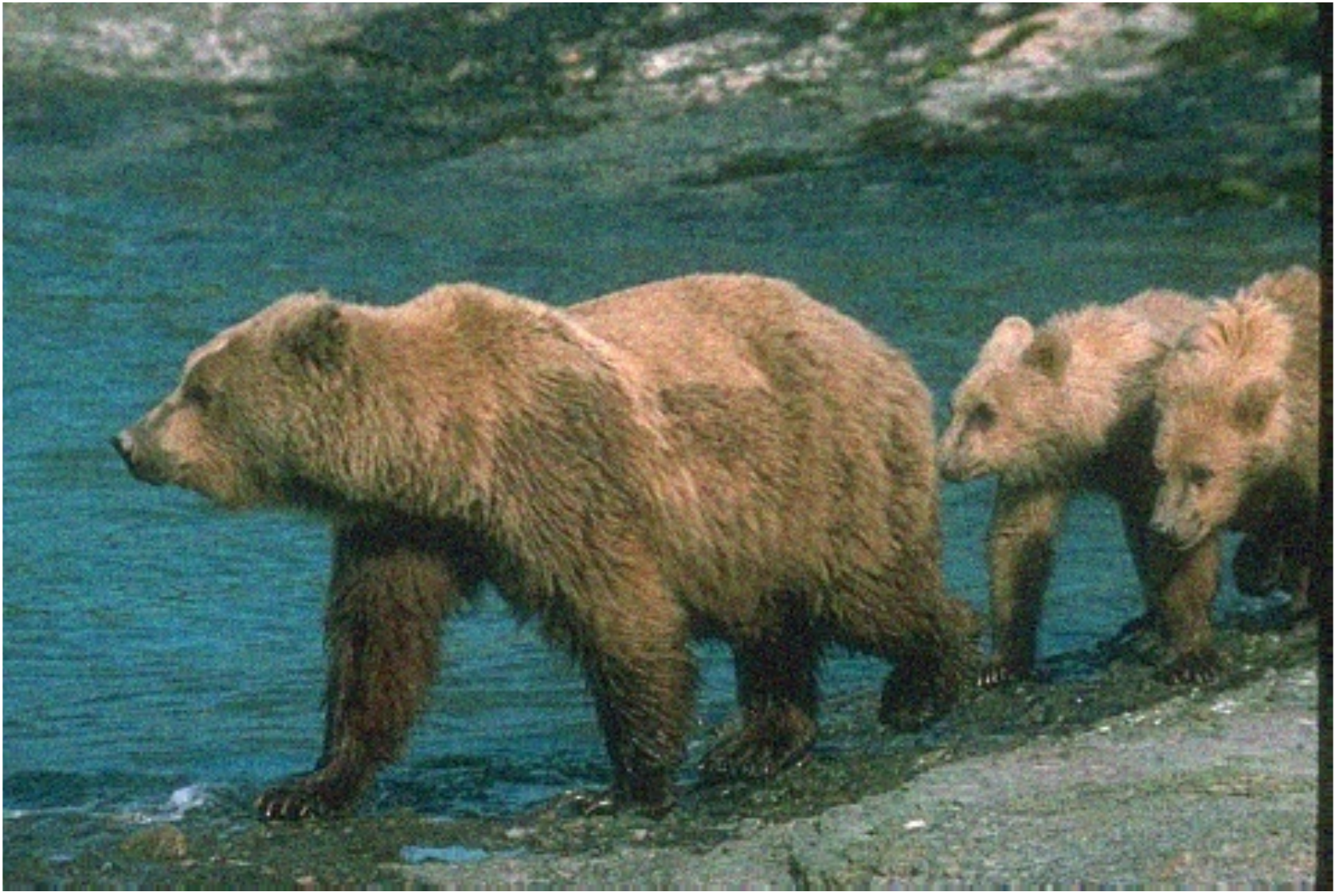}
  \put(67.5,35.5){\setlength{\fboxsep}{15pt}\setlength{\fboxrule}{0.5pt}\cfbox{red}}
  \end{overpic}&
\includegraphics[width=.33\linewidth]{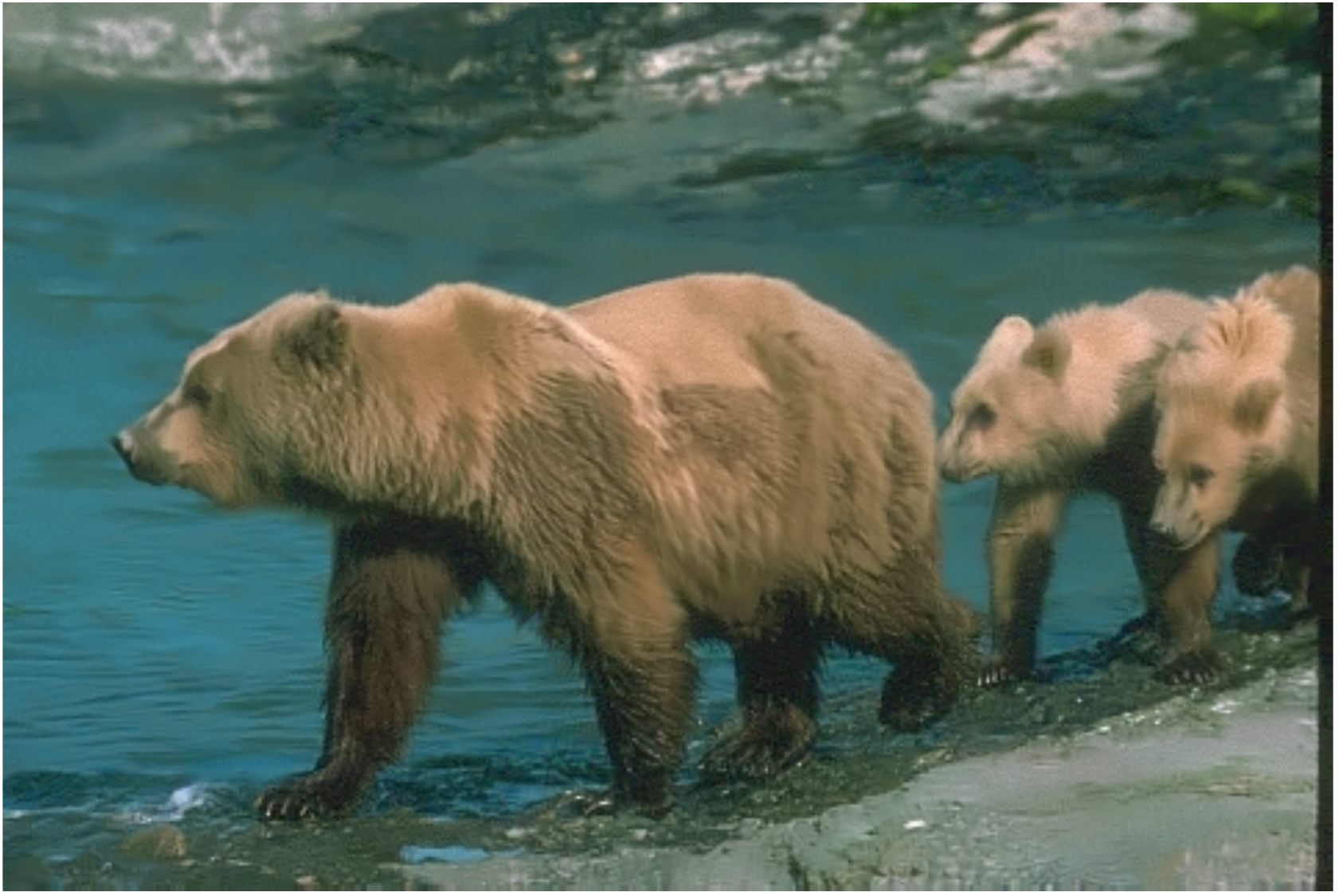}&
\includegraphics[width=.33\linewidth]{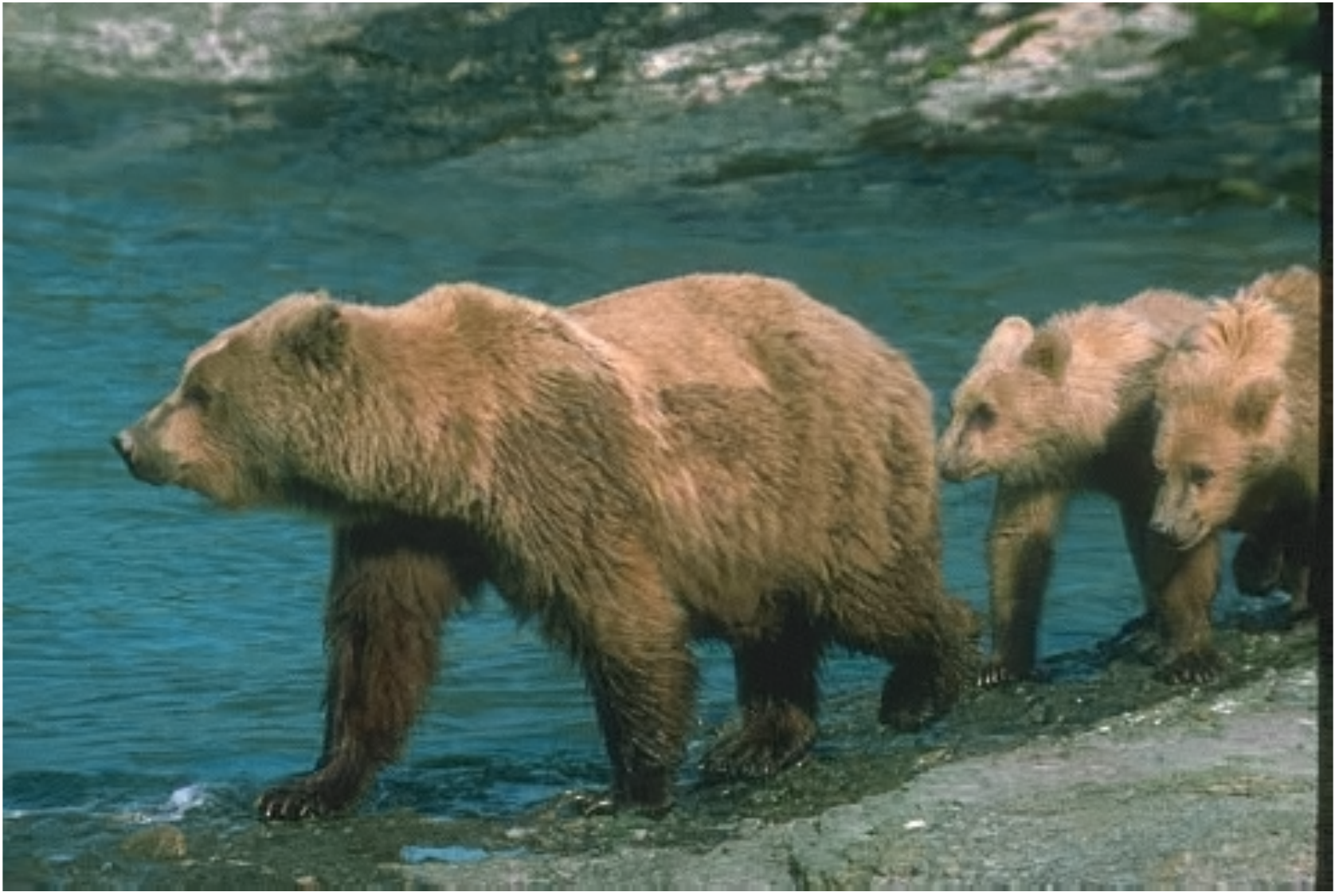}\\
(a) Noisy image ($\sigma = 12$) & (b) Noise Clinic~\cite{Lebrun2015} & (c) CBM3D~\cite{Dabov2007}\\
\includegraphics[width=.33\linewidth]{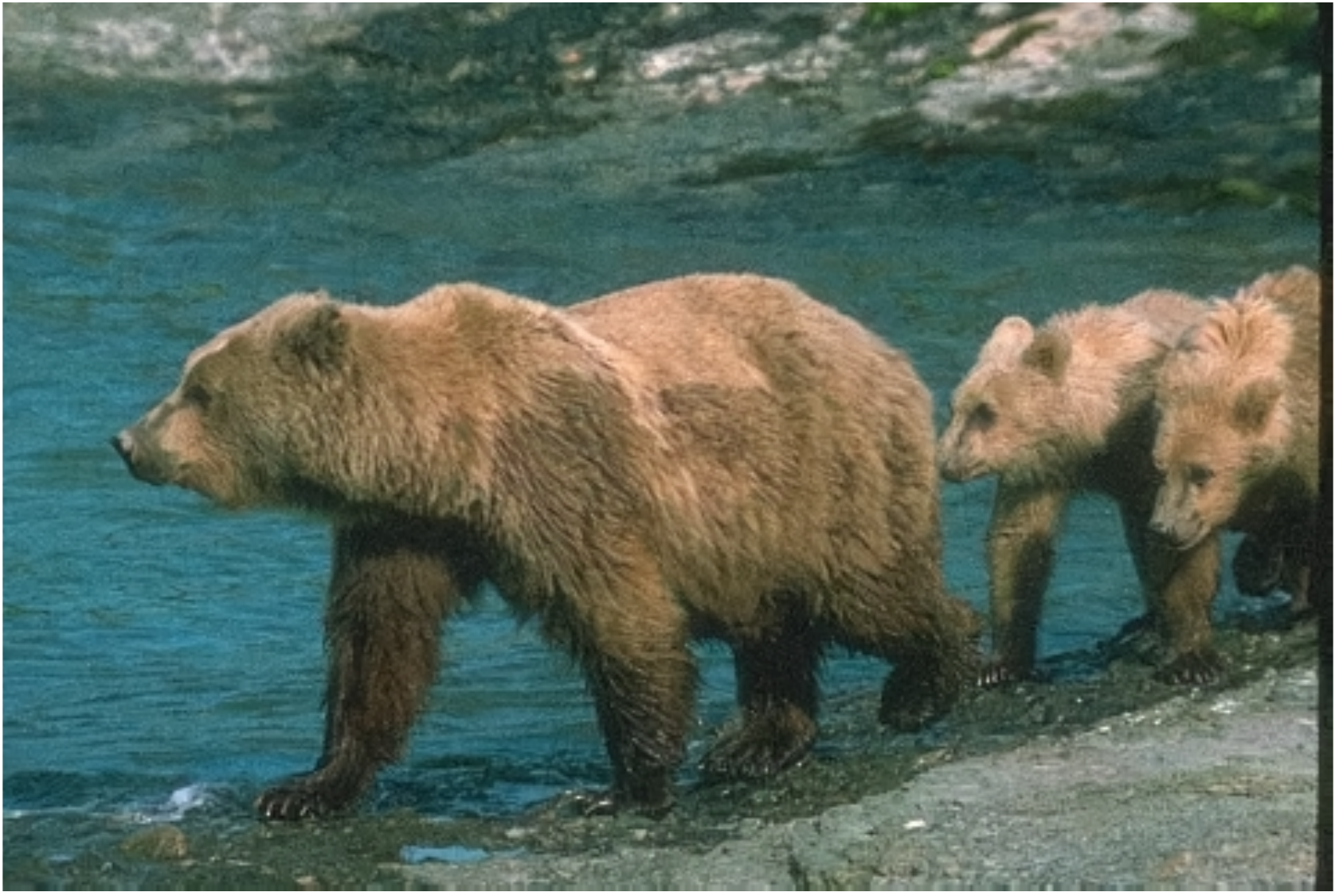}&
\includegraphics[width=.33\linewidth]{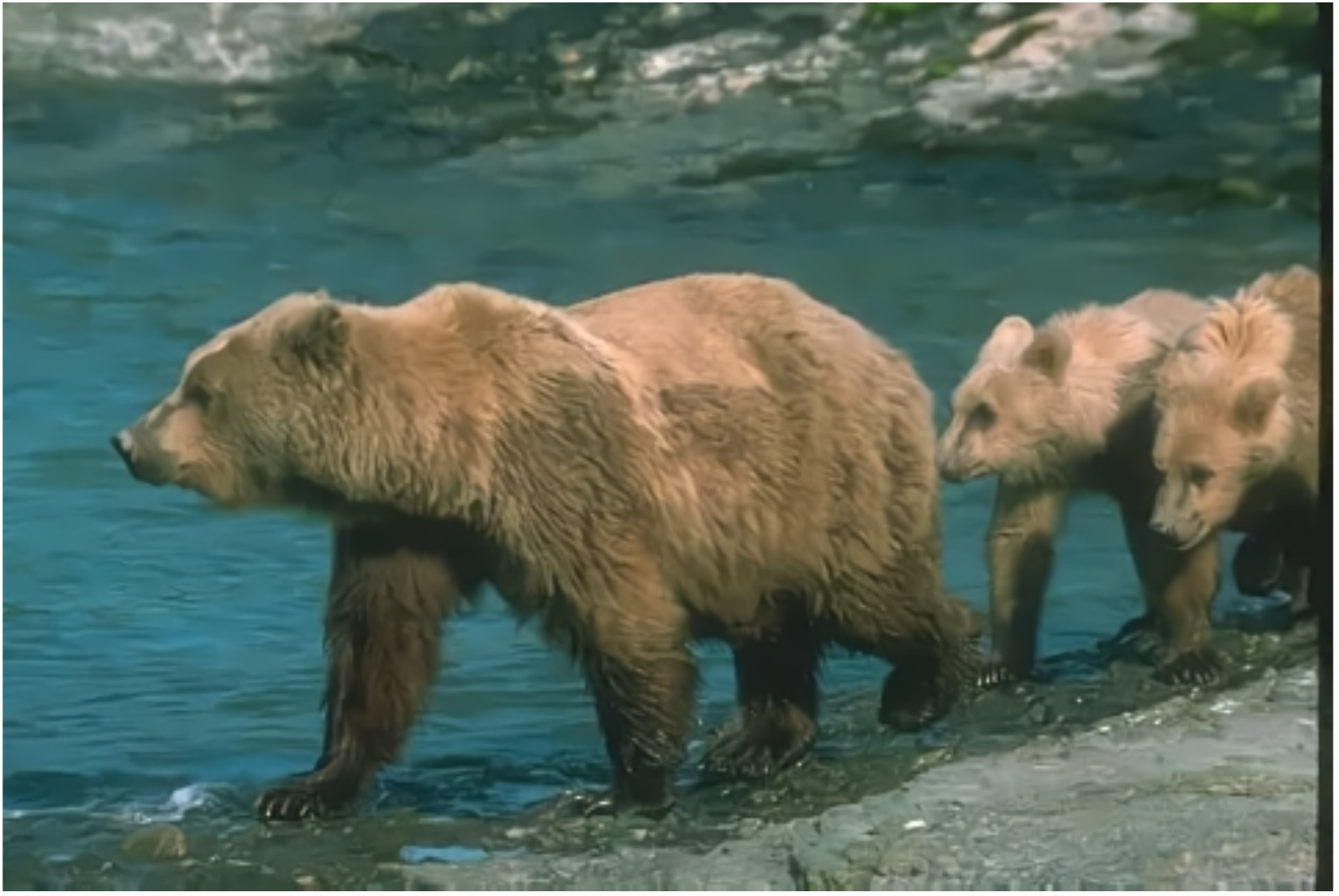}&
\includegraphics[width=.33\linewidth]{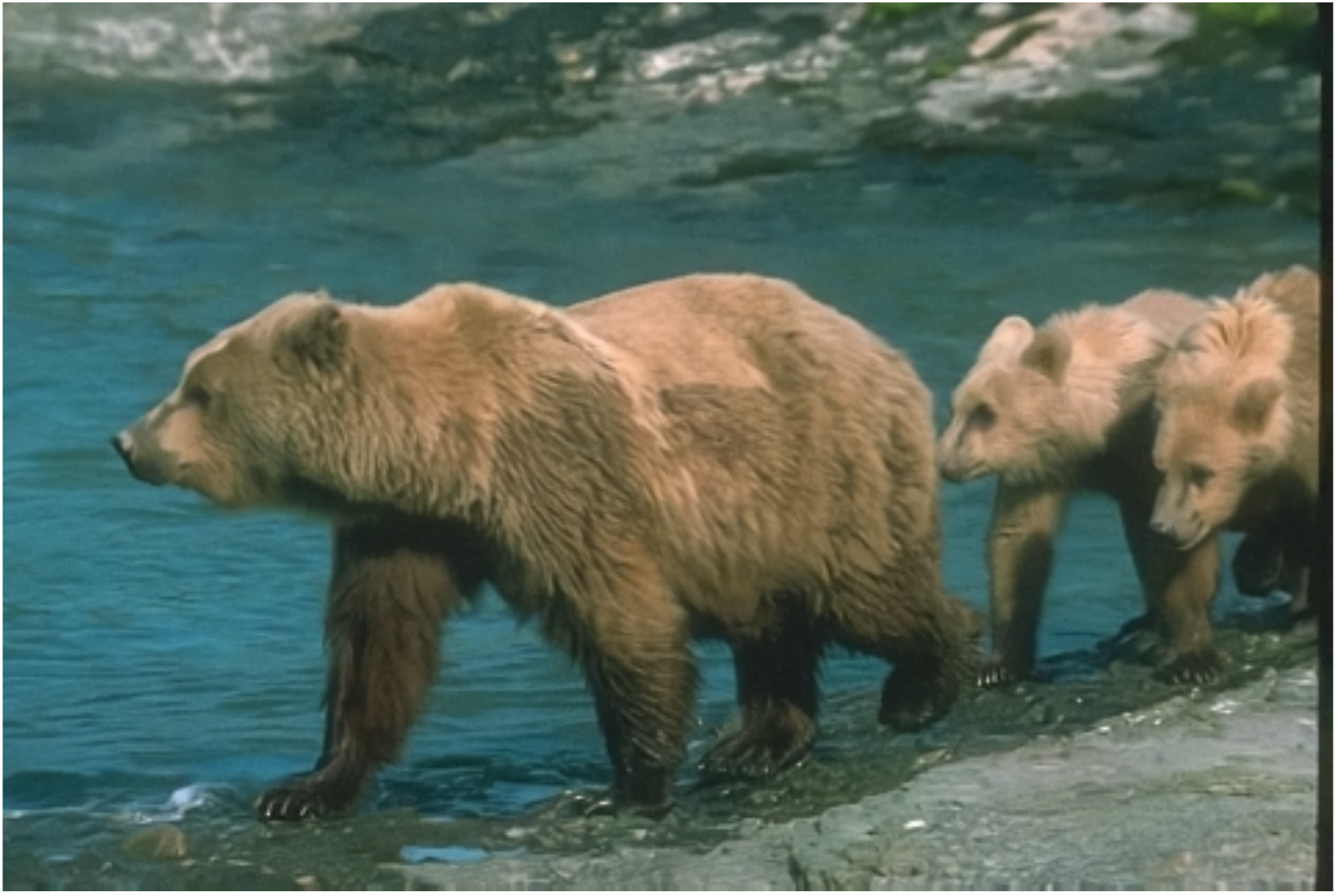}\\
(d) CDnCNN-S~\cite{Zhang2017} & (e) $\operatorname{CUDNet}_5$ & (f) $\operatorname{CUNLDNet}_5$\\
\end{tabular}
   \caption{Real color image denoising. Some important differences in the restoration quality between the methods under comparison can be spotted in the highlighted image region. \textbf{Images are best viewed magnified on a computer screen.}}
   \label{fig:RealColorComp3}
\end{figure*}

{\small
\bibliographystyle{ieee}
\bibliography{references}
}

\end{document}